%% file: main.tex
\documentclass[a4paper, twoside, leqno]{article}
\usepackage{PRIMEarxiv} 
\input{Header_setting}
\usepackage[dvipsnames]{xcolor}
\AddToHook{cmd/appendix/before}{%
	\crefalias{section}{appendix}%
	\crefalias{subsection}{appendix}%
}

\begin{document}
\title{AMORE: \textbf{A}daptive \textbf{M}ulti-Output \textbf{O}perato\textbf{r} N\textbf{e}twork for Stiff Chemical Kinetics}

\author[1]{Kamaljyoti Nath} 
\author[1]{Additi Pandey}
\author[2]{Bryan T. Susi} 
\author[3]{Hessam Babaee} 
\author[1]{George Em Karniadakis}
\affil[1]{Division of Applied Mathematics, Brown University, Providence, RI, USA}
\affil[2]{Applied Research Associates, Inc., Raleigh, NC, USA}
\affil[3]{Department of Mechanical Engineering and Materials Science, University of Pittsburgh, Pittsburgh, PA, USA}

\fancypagestyle{empty}{\fancyhf{}\renewcommand{\headrulewidth}{0pt}\fancyfoot[R]{\today}\fancyfoot[L]{Preprint}}
\maketitle
\begin{abstract}
Time integration of stiff systems is a primary source of computational cost in combustion, hypersonics, and other reactive transport systems. This stiffness can introduce time scales significantly smaller than those associated with other physical processes, requiring extremely small time steps in explicit schemes or computationally intensive implicit methods. Consequently, strategies to alleviate challenges posed by stiffness are important. While neural operators (DeepONets) can act as surrogates for stiff kinetics, a reliable operator learning strategy is required to appropriately account for differences in the error between output variables and samples. Here, we develop AMORE, Adaptive Multi-Output Operator Network, a framework comprising an operator capable of predicting multiple outputs and adaptive loss functions ensuring reliable operator learning. The operator predicts all thermochemical states from given initial conditions. We propose two adaptive loss functions within the framework, considering each state variable's and sample's error to penalize the loss function. We designed the trunk to automatically satisfy the Partition of Unity. To enforce unity mass-fraction constraint exactly, we propose an invertible analytical map that transforms the $n$-dimensional species mass-fraction vector into an ($n-1$)-dimensional space, where DeepONet training is performed. We extend the proposed adaptive loss functions to trunk and branch training in two-step training of DeepONet with multiple outputs. Furthermore, we also implemented another unity mass fraction constraint exactly using a softmax function on the predicted mass fraction. We demonstrate the efficacy and applicability of our models through two examples: the syngas (12 states) and GRI-Mech 3.0 (24 active states out of 54). The proposed DeepONet will be a backbone for future CFD studies to accelerate turbulent combustion simulations. AMORE is a general framework, and here, in addition to DeepONet, we also demonstrate it for FNO.
\end{abstract}
\let\thefootnote\relax\footnotetext{\textit{Corresponding author:} G. E. Karniadakis (george\_karniadakis@brown.edu) \\
\textit{E-mail addressees:} kamaljyoti\_nath@brown.edu (K. Nath), additi\_pandey@brown.edu (A. Pandey), bsusi@ara.com (B. T. Susi), h.babaee@pitt.edu (H. Babaee), george\_karniadakis@brown.edu (G. E. Karniadakis)}
\section{Introduction}
\label{Section:Introduction}
Chemical kinetics plays a key role spanning various disciplines of science and engineering, extending beyond combustion, hypersonics, environmental engineering, chemical engineering, and biomedicine. In chemical kinetics, the ordinary differential equations (ODEs) representing the net production rate of the species involved in the reaction are stiff in nature because of the presence of widely varying reaction rates. This issue of stiffness becomes particularly challenging in turbulent combustion simulations, where chemical kinetics must be solved concurrently with the balance of mass, momentum, and energy conservation. In reactive flow simulations, the shortest time scales introduced by chemical reactions can be orders of magnitude smaller than those associated with key physical processes, such as convection and diffusion, collectively known as hydrodynamic time scales. This disparity becomes particularly challenging in high-fidelity turbulent combustion simulations, such as large eddy simulations (LES) or direct numerical simulations (DNS). In these simulations, the spatial resolution requirements are driven by the need to resolve the smallest turbulent length scales and/or flame thickness, resulting in extremely large computational grids. Under these conditions, even for a moderate number of chemical species, fully implicit time integration becomes computationally infeasible. Consequently, semi-implicit or operator-splitting schemes are often employed, wherein non-reactive terms are integrated explicitly, while the stiff chemical source terms are treated implicitly. However, implicit integration of chemical kinetics is itself highly expensive for direct solvers; the computational cost scales as $\mathcal{O}(n^3)$, where $n$ is the number of species. Given the large number of grid points required to resolve turbulent and flame structures, the overall computational cost remains a major bottleneck in high-fidelity turbulent combustion simulations.
\par Consequently, alleviating the computational cost associated with chemical stiffness has rightfully merited substantial attention in the turbulent combustion literature, resulting in the development of numerous approaches that span a spectrum of trade-offs between \emph{generalizability}, \emph{accuracy}, and \emph{computational cost}. A large subset of these methods is rooted in the principle of timescale separation, wherein fast species, subspaces, or manifolds are distinguished from their slower counterparts. Some well-known techniques to address stiffness in reactive systems include the quasi-steady state approximation (QSSA), partial equilibrium (PE), intrinsic low-dimensional manifold (ILDM) \citep{MP92}, reaction–diffusion manifold (REDIM) \citep{Bykov_2007aa}, computational singular perturbation (CSP) \citep{LG94}, slow invariant manifold (SIM) and approximate slow invariant manifold (ASIM). Among these, QSSA and PE are the most computationally efficient techniques. However, they require stringent equilibrium assumptions and expert-driven identification of fast species and reactions, which limits their generalizability and can compromise accuracy. The other techniques such as ILDM, REDIM, CSP, SIM, and ASIM, offer greater generalizability and accuracy by identifying slow and fast subspaces or manifolds; however, they require Jacobian evaluations, manifold construction, or numerical optimization, making them considerably more complex and computationally demanding than QSSA and PE. Recently, an on-the-fly reduced-order modeling approach was introduced to mitigate chemical stiffness \citep{JLBC25}. Using a matrix cross-approximation algorithm, this method solves the species transport equations on a low-rank matrix manifold. It demonstrates strong generalizability and high accuracy, and reduces computational cost for the cases studied by at least an order-of-magnitude.
\par With the advancement of machine learning methods and techniques, there has been a noticeable increase in the use of artificial neural networks in place of classical numerical methods to solve a system of stiff differential equations due to their computational efficiency when dealing with large and complex systems and their accuracy in approximations. Moreover, the flexibility offered by neural networks, in terms of the model architecture, hyperparameters, and use of varied activation and loss functions, makes them adaptable for modeling complex systems. To this end, artificial neural network-based models \cite{Sharma_2020_Deep}, parallel ResNets \cite{Brown_2021_RESDON}, physics-informed neural networks (PINNs) \citep{Raissi_2019_PINN} based methodologies using the extreme theory of functional connections (X-TFC) \citep{Schiassi_2021_Extreme} in \cite{Florio_2022_SKPINNXTFC}\cite{Frankel_2024_Hybrid}, have been employed to model stiff chemical kinetics problems. Neural ODEs (NODEs) have also been applied to stiff systems \citep{Kim_2021_Stiff, Kumar_2025_NODE}. In addition to this, Stiff PINNs by \citeauthor{Ji_2021_Stiff} \cite{Ji_2021_Stiff} attempted to solve stiff ODEs using PINNs via a QSSA. The authors also highlighted that, for a complex chemical system, manually deriving the QSSA formula cannot only be time-consuming but also requires a thorough understanding of the process. Moreover, mild stiffness may still persist in the reduced system. In general, although PINN-based methods can predict the dynamics of stiff chemical systems, they need to be trained for any change in the given conditions, such as the initial conditions.

\par There has also been a surge in learning dynamical systems from given data \citep{Ayed_2019_Dynamics, Kostic_2022_Koopman}, in particular, using operator learning methods \citep{Lu_2021_DeepONet, Li_2020_FNO, Tripura_2023_WNO, Kovachki_2023_Neural, centofanti_2024_HHM, pellegrini_2025_HDIModel}. Neural operator surrogates involve \emph{offline} training costs, but their \emph{online} (inference) costs are much smaller than implicit solvers. The advantage of using a neural operator is that once trained offline, it can be used for fast online predictions over a new input dataset.  \citeauthor{Venturi_2022_SVDPF} \cite{Venturi_2022_SVDPF} proposed flexDeepONet, a DeepONet \citep{Lu_2021_DeepONet} based architecture, and employed it to analyze combustion chemistry in an isobaric reactor, where each thermodynamic state variable has its own branch and trunk network. The architecture prepends an additional pre-transformation network that uses branch input to transform the trunk input before they enter the trunk network. This allows the discovery of a moving frame of reference, allowing multiple scenarios to be compressed into a lower number of modes. \citeauthor{Goswami_2024_Stiff_Kinetics} \cite{Goswami_2024_Stiff_Kinetics} proposed different variants of DeepONet (e.g., DeepONet, PoU-DeepONet, and autoencoder-based DeepONet) for different stiff chemical kinetics problems such as ROBER, POLLU, pure kinetic syngas, and turbulent flame syngas problems. The authors considered different DeepONets for different species for ROBER and POLLU problems, with only one input at a time. Furthermore, the authors did not consider the dynamics and stiffness of the problem. While in the syngas problems, the DeepONet can predict multiple output variables, the training and predictions are considered at times $t+250\Delta t, t+500\Delta t, t+750\Delta t$ and $t+1000\Delta t$, with $\Delta t = 10^{-8}$ s. With such a high time stepping, there is a high possibility that the model does not take the stiffness (i.e., the peaks in the dynamics) into consideration. Furthermore, the authors have also confirmed that the auto-regressive prediction error reported in the paper for the syngas pure kinetics problem is not predicted using an auto-regressive methodology. In the present study, we have shown that our vanilla DeepONet and DeepOKAN can achieve better accuracy. Moreover, with the proposed adaptive loss functions and two-step training, we obtained better and more reliable accuracy. Moreover, for the auto-regressive prediction, our results show that the 90 percentile of the samples has less than 5\% error when predicted using an auto-regressive manner for the complete time series.
\par \citeauthor{Kumar_2024_Combustion} \cite{Kumar_2024_Combustion} introduced React-DeepONet, where a subset of thermochemical scalars is evolved and others are reconstructed using ANNs. The model takes as input this subset, called representative scalars $\phi$, at time $t$ and a time increment $\delta t$ and predicts them at $t+\delta t$. This model includes a pre-transformation network to transform the trunk input and a parameter input sub-network to encode parametric variations. In addition to the initial training, a refined training paradigm is adopted. A latent space React-DeepONet is also introduced for dimensionality reduction of representative scalars using an autoencoder to yield latent variables on which React-DeepONet is implemented. \citeauthor{Weng_2025_ExFNO} \cite{Weng_2025_ExFNO} implemented a Fourier neural operator (FNO) \citep{Li_2020_FNO} based extended FNO to model the stiff chemical kinetics of species via the Box-Cox transformation \citep{BoxCox_1964_BCtransf}. They introduce additional constraints, such as the conservation of mass and elements in the loss function, along with a balanced data loss function (based on focal loss). The balanced loss function updates for every data sample, taking constant values for weights depending on the variation of mass fractions between adjacent time steps. State space sequence models such as Mamba \citep{Gu_2023_mamba} have emerged as efficient models for operator learning \citep{Hu_2024_state}, and more recently,  their application to stiff kinetics has been explored in \cite{Pandey_2025_kineticmamba}.
\par For efficient integration with a CFD solver, we require the neural operator to accurately predict the kinetics, ensuring that the stiffness is taken into account. Moreover, for computational efficiency and scalability, it is desirable that a single neural operator should be able to predict all thermo-chemical state variables, taking into account the difference in error between the state variables. To this end, we propose a framework called \textbf{A}daptive \textbf{M}ulti-output \textbf{O}perato\textbf{r} N\textbf{e}twork (AMORE) by considering a Deep Operator Network, popularly known as DeepONet, and its variants to accurately predict the dynamics of the stiff chemical system and adaptive loss functions to ensure reliable operator learning. One of the major advantages of using neural operators (e.g., DeepONet) that are a functional approximation model over a fully connected neural network (FCNN) that are function approximation model is that it can predict the dynamics of the output variables from the initial conditions, unlike one-step prediction in the case of an FCNN. DeepONet is a neural operator, proposed by \citeauthor{Lu_2021_DeepONet} \cite{Lu_2021_DeepONet}, wherein the two one-hidden-layered networks of \citeauthor{Chen_1995_universal_operator} \cite{Chen_1995_universal_operator} are replaced by two deep neural networks. These two networks are known as the branch and the trunk network of DeepONet, which aims to map an infinite-dimensional input function ($u\in \bm{U}$) to another infinite-dimensional output function ($v\in \bm{V}$) in the Banach space. The input function is taken by the branch network, which captures its characteristics, while the spatio-temporal coordinates where the output needs to be predicted are given as input into the trunk network. The output of the DeepONet is given by a dot product of the output of the branch and trunk networks. We can assume the trunk output as the universal basis function and the branch output as the coefficient for these basis functions for a particular input function. DeepONet has been successfully applied in modeling many complex problems \citep{Pengzhan_2022_MIONet, Lu_2022_Comp_deep_fno, Liu_2022_Causality, Zhu_2023_fourier, Kiyani_2024_Crack, Rivera-Casillas_2025_SWE, Nath_2025_DeepONet_FWI}, demonstrating its potential to act as a surrogate model to diverse fields of problems. One of the advantages of DeepONet is that it can be applied to simulated data, experimental data, or both \citep{Higgins_2021}, and it can predict a continuous output. Moreover, there is no restriction on the type of network considered in either branch or trunk networks. With the recent proposal of Kolmogorov Arnold networks (KANs) \citep{Liu_2024_KAN}, researchers have considered different variants of KAN as a choice of network architecture for the trunk and branch network \cite{Abueidda_2025_DeepOKAN, Shukla_2024_DeepOKAN}. \citeauthor{Shukla_2024_DeepOKAN} \cite{Shukla_2024_DeepOKAN}, has performed a comprehensive and FAIR comparison between MLP and KAN for various problem types, including DeepONet and DeepOKAN. Recently \citeauthor{Sanghyun_2024_two_step} \cite{Sanghyun_2024_two_step} proposed a novel two-step training strategy for DeepONet, where the trunk and the branch are trained in a sequential manner rather than being trained together. The method involves QR decomposition of the trunk output, making the basis function orthogonal and more expressive. 
\par The present study aims to accurately model the dynamics of stiff chemical kinetics via the AMORE framework. Through this framework, we propose two different types of DeepONet architectures, using ResNet and KAN as choices of network architecture for the branch and trunk networks. To improve the accuracy of the prediction, we propose two adaptive loss functions within the framework. Furthermore, we consider the two-step training of DeepONet. In the next section, \Cref{Subsection:Main contribution}, we discuss the main contribution of the present study. 
\subsection{Main contribution}
\label{Subsection:Main contribution}
In this work, we propose a neural operator framework, specifically DeepONets, to accurately predict the dynamics of stiff systems, particularly of stiff chemical systems described by the system of stiff ODEs. We call this framework adaptive multi-output operator network (AMORE) for stiff chemical kinetics. In \Cref{Figure:AMORE}, we have illustrated the AMORE framework, which consists of an operator network that is capable of predicting multiple outputs and adaptive loss functions. Since stiff chemical systems have multiple state variables (e.g., species' mass fractions and temperature), it is desired, and therefore, we propose, to consider a single DeepONet to predict all the state variables instead of considering separate DeepONets for each state variable. This configuration will help in the scalability of the proposed method for chemical systems involving a large number of state variables. We have trained our proposed DeepONet with the AMORE scheme. This framework will serve as a cornerstone for future studies on integrating neural operators with CFD solvers to improve the computational efficiency of the coupled models. We summarize the main contributions through our proposed AMORE framework as follows:
\begin{figure}[H]
    \centering 
    \includegraphics[width=0.8\textwidth]{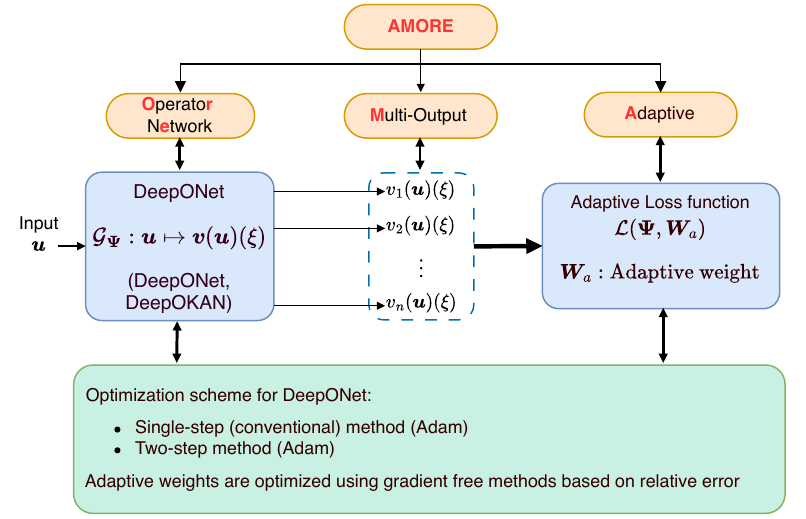}
    \caption{\textbf{AMORE:} A schematic diagram of the proposed \textbf{A}daptive \textbf{M}ulti-output \textbf{O}perato\textbf{r} N\textbf{e}tworks (AMORE) framework. It consists of an operator network that is capable of predicting multiple outputs and adaptive loss functions. In the present study, we consider Deep Operator Network (DeepONet) with two different architectures as the choice of the operator network, the DeepONet and DeepOKAN. We propose two adaptive loss functions that are updated using a gradient-free method based on the relative error between the state variables and samples.}
    \label{Figure:AMORE}
\end{figure}
\begin{itemize}[leftmargin=*, label=$\blacksquare$]
    \item \textbf{DeepONet architecture}: We propose two DeepONet architectures to predict the dynamics of the multiple state variables of stiff chemical systems from given initial conditions of these variables. These architectures consist of a single trunk network and a branch network. The first proposed architecture consists of ResNet as both the branch and trunk networks, while the second proposed architecture consists of a ResNet for the branch network and a KAN as the trunk network. We shall refer to them as DeepONet and DeepOKAN, respectively.
    \begin{itemize}[label=$\blacklozenge$, leftmargin=5mm]
        \item \textbf{Partition of Unity}: We design our trunk network such that it automatically satisfies the partition of unity (PoU) constraint in the trunk network. 
        \item \textbf{Conservation of mass}: We know that the total mass of the reactants and the products at any instant is the same. Hence, we consider a loss function due to conservation of mass (CoM) in terms of mass fraction along with the data loss. The sum of the mass fractions of all the species is one at any given time, and we consider it as an additional penalty in the loss function.
        \item \textbf{Adaptive loss functions}: To improve prediction accuracy and efficient operator learning, we propose two gradient-free adaptive loss functions. The basic idea behind these loss functions is that the state variables and the samples with higher error need to be penalized more in the loss function.
    \end{itemize}
    \item \textbf{Two-step training of DeepONet}: We consider the two-step training of DeepONet to predict the dynamics of the multiple state variables of stiff chemical systems from given initial conditions of these variables with one trunk but different basis functions for each output variable, and one branch but different coefficients for each output variable. 
    \begin{itemize}[label=$\blacklozenge$, leftmargin=5mm]
        \item \textbf{Adaptive loss functions}: In the two-step training process, we extend the two adaptive loss functions in training both the trunk and the branch networks with the appropriate modifications.
    \end{itemize}
    \item \textbf{Mass conserving DeepONet}: The loss function corresponding to the CoM constraint does not guarantee the conservation of mass in terms of the mass fraction of the species and depends on the penalty considered in the loss function. Consequently, we propose a mass-conserving neural operator that automatically satisfies the CoM constraint at each instant for all the participating species whose sum of mass fractions is one. Furthermore, we also incorporate another CoM constraint using the softmax activation function on the species' mass fractions so that the summation of these mass fractions is always one.
    \item \textbf{AMORE implementation in FNO}: The proposed AMORE framework is a general framework. It can be used with other neural operators as well. To demonstrate this, we have implemented the AMORE framework with FNO.
\end{itemize}
\vspace{0.25cm}
Together, these contributions result in a remarkable improvement in prediction accuracy, surpassing the current state of the art. As discussed above, this work introduces several contributions in different forms. Conceptually, we propose a framework for accurately predicting multiple state variables from a single neural operator (e.g., DeepONet/DeepOKAN or FNO). At the algorithmic level, we develop two gradient-free adaptive loss functions that update based on the relative error between state variables/samples for DeepONet. We extend these adaptive weights to the two-step training of DeepONet. We also proposed a mass-conserving map that automatically satisfies the conservation of mass fraction of unity, which may be categorized as conceptual and implementation novelty. From an implementation perspective, we consider different architectures and incorporate a loss function enforcing conservation of mass. We design our trunk network to satisfy the PoU automatically. We also implement another CoM constraint by applying a softmax function to the predicted mass fraction to automatically ensure the CoM is unity.
\section{Problem statement}
\label{Subsection:Problem statement}
For demonstration purposes, we consider adiabatic, constant-pressure zero-dimensional reactions of ideal gases described by the following system of nonlinear ordinary differential equations:
\begin{subequations}\label{Eq:ODE general}
\begin{align}
    \dfrac{d y_k}{d t} &= f_{y_k}(\bm y, T)= \frac{W_k}{\rho}\sum_{j=1}^{n_r} \nu _{kj} \mathcal Q_j, \quad \bm y(0) = \bm y_0,\label{Eq:Temperature_const_prssure} \\
    \dfrac{d T}{d t}   &= f_{T}(\bm y, T) = -\frac{1}{c_p}\sum_{k=1}^{n} h_k f_{y_k}, \quad T(0)= T_0, \label{Eq:Temperature_const_volume} 
\end{align} 
\end{subequations}
where $\rho$ is the gas mixture density, $n_r$ is the number of reactions, $c_p$ is the heat specific capacity of the gas mixture, $W_k$ and $h_k$ are the molecular weight and enthalpy of species $k$, respectively. Moreover, $\nu_{kj}$ is the net molar stoichiometric coefficient of species $k$ and reaction $j$, and $\mathcal Q_j$ is the progress rate of reaction $j$. 
\par Through the present study, our objective is to develop a comprehensive framework of neural operators that efficiently and accurately models a stiff ODE system governing a chemical-kinetics reaction mechanism (\Cref{Eq:ODE general}). Once the neural operator is trained offline, it can be used as a surrogate for the time integration of \Cref{Eq:ODE general}. The neural operator maps the initial condition of the stiff ODEs (\Cref{Eq:ODE general}) to the dynamics of the state variables, i.e., temperature and the species' mass fractions. Thus, the operator learning can be mathematically written as,
\begin{equation}
    \mathcal{G}_{\bm{\Psi}}:\bm{Y}_0 \mapsto \bm{Y}\left(\bm{Y}_0\right)(t),
\end{equation}
where $\bm{Y}_0 = [T(0), y_1(0), y_2(0),\cdots,y_n(0)]^\top$ is the initial condition corresponding to the temperature and the species' mass fractions. The output is the transient evolution of temperature and the species' mass fractions.
\par
\begin{figure}[H]
    \centering
    \includegraphics[width=1\textwidth, trim={0.325cm, 0cm, 0.325cm, 0cm}, clip]{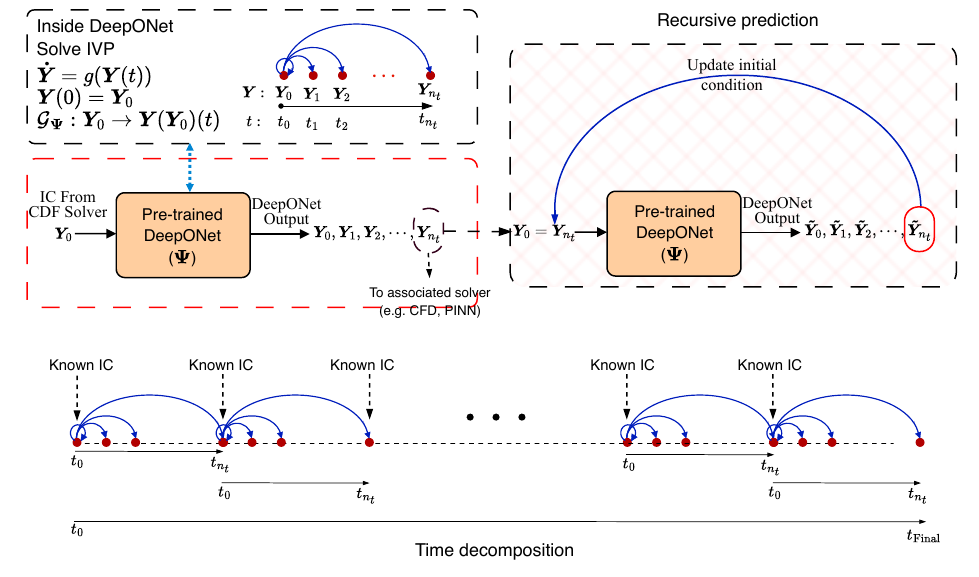}
    \caption{\textbf{DeepONet for stiff chemical kinetics:} Schematic of the DeepONet prediction scheme considered in the present study. In the left middle, within the red dashed box, we have shown the prediction of the state variable using the pre-trained DeepONet parametrized with $\bm{\Psi}$. The input to the DeepONet is the initial condition of the state variables, $\bm{Y}_0$. The output is the predicted dynamics of the state variables, $\bm{Y}_0, \bm{Y}_1, \bm{Y}_2,\cdots\bm{Y}_{n_t}$. At the top right, we have shown the DeepONet learning process. It solves the initial value problem of the stiff system (\Cref{Eq:ODE general} by mapping the initial condition to the dynamics of the state variables. Generally, the input for the DeepONet, the initial conditions, comes from the other associated solver, and the output of the DeepONet at a particular time instant goes to the other associated solver as input. Since we have not considered the DeepONet-CFD/PINN integration in the present study, we consider the time decomposition of the dynamic equation as shown at the bottom of the figure. In the time decomposition conditions, it is assumed that the initial conditions of the ODEs are known after every $t=t_{n_t}=n_t\Delta t$, which ideally comes from the associated solver. Thus, we will design our DeepONet to predict the dynamics considering time decomposition, assuming the initial conditions are known at every $t=t_{n_t} = n_t\Delta t$. Furthermore, when the DeepONet output is not sufficiently progressed in time to be input to the associated solver, we propose to consider a recursive prediction strategy as shown on the right within the black dashed box with the hatch. The output of the DeepONet at time instant $t=t_{n_t}=n_t\Delta t$ is considered as the input to the pretrained DeepONet, and the process continues till sufficient progress in time is achieved for the chemical kinetics to be input to the associated solver.}
    \label{Figure:Problem statement}
\end{figure}

 In \Cref{Figure:Problem statement}, we have shown a schematic of the prediction using DeepONet for the problem we have considered. We have considered a system with computational time domain $[t_0,\: t_\text{Final}]$ with the time step for the chemical kinetics $\Delta t$. We assume the time step for the other associated physical process (e.g., CFD or a PINN model for mass, momentum, and energy conservation) is $\Delta \tau$ and $\Delta t \ll \Delta \tau$. The input to the DeepONet is the initial condition ($\bm{Y}_0$) of the state variables, and it predicts the dynamics of the state variables, $\bm{Y}(t)$ from $t=t_0 = 0$ to $t=t_{n_t}=n_t\Delta t$. The predicted state variable at $t=\Delta \tau = n_t\Delta t$ can be considered as input to the associated solver for the other physical system. We would like to mention that in the present study, we have not considered the integration with the CFD solver. However, we have discussed it as an example for illustrative purposes of our approach. Using the CFD application as an example, after the transport of the fluid is calculated by the CFD solver, the instantaneous states are considered as the initial conditions for the chemical kinetics. Typically, this would be an implicit numerical solver, possibly with additional heuristic modeling for partial combustion, but in this case becomes input to the DeepONet in lieu of the conventional numerical solution. Since in the present study, we did not consider the integration with CFD solver, we consider time decomposition of the system of ODEs given by \Cref{Eq:ODE general} where the initial conditions of each segment are assumed to be known (generally input from the other associated solver). In \Cref{Figure:Problem statement}, bottom, we have shown the time decomposition of the ODEs where, after every $t_{n_t} = n_t\Delta t$, the initial condition is known for each segment. Furthermore, in some cases, it is possible that the dynamics predicted by the DeepONet are not sufficiently advanced in time to be given as input to the other associated solver. In that case, we propose to consider a recursive prediction methodology in the neural operator where the output of the DeepONet at the last time instant of a segment is considered again as input to the DeepONet for the next segment. This process is continued until we have obtained sufficient progress in time for the chemical kinetics to be considered as input for the associated solver. 

\par To obtain the objectives discussed above, we propose neural operator, particularly DeepONet \citep{Lu_2021_DeepONet} architectures which is based on the universal approximation theorem for operators \citep{Chen_1995_universal_operator}. In the following sections, we shall discuss the proposed DeepONet architectures and variations considered to obtain satisfactory prediction accuracy. We shall demonstrate this through two computational experiments using the syngas (synthetic gas) and GRI-Mech 3.0 mechanisms.
\section{Methodology}
\label{Section:Methodology}
We will use the AMORE framework introduced in the previous section to meet our objective for accurately predicting the stiff chemical kinetics described by a system of homogeneous ODEs, i.e., \Cref{Eq:ODE general}. We will consider DeepONet as the neural operator and propose two adaptive loss functions. With $n$ species involved in the system, the total number of state variables is $j=n+1$. These state variables consist of the temperature of the system and the mass fractions of $n$ species. We will design our neural operator to predict the dynamics of $j=n+1$ state variables from the initial condition of these $j=n+1$ state variables. In the subsequent sections, we shall discuss the operators considered in this study, i.e., the DeepONet and its variants, including different loss functions formulated to enhance the accuracy of the operator learning. We will propose two adaptive loss functions. We will also discuss two training paradigms of DeepONet: the standard training, where both trunk and branch are trained together, and the two-step training, where the trunk and branch networks are trained separately in a sequential manner. We will also propose a mass-conserving operator that automatically satisfies the CoM constraint. AMORE is a general framework and can be used with other neural operators as well. We will also implement AMORE with Fourier Neural Operator (FNO) \citep{Li_2020_FNO} as a case study. In \Cref{Table:Different methods}, we have tabulated various methods considered in this study and their respective sections where these methods are discussed in detail.
\begin{table}[H]
\centering
\caption{Different methods considered in the present study and the corresponding sections where the methods are discussed in detail.}
\label{Table:Different methods}
\renewcommand{\arraystretch}{1.25}
\begin{tabular}{C{2.45cm}|L{10.cm}|C{1.15cm}}
\hline
& \multicolumn{1}{c|}{Description} & Section \\ \hline
 & Deep Operator Network including DeepONet and DeepOKAN & \S \ref{Subsection:DeepONet} \\ \cline{2-3}
 & Partition of Unity (PoU) and Conservation of mass (CoM) & \S \ref{Subsection:POU and CoM} \\ \cline{2-3}
\multirow{-3}{*}{\begin{tabular}{C{2cm}}
     DeepONet (Single-step training)
\end{tabular}} & Adaptive loss functions & \S \ref{Subsection:Adaptive loss function} \\ \hline
Two-step DeepONet & Two-step training of DeepONet/DeepOKAN along with adaptive loss functions for trunk and branch training & \S \ref{Subsection:Two-step training} \\ \hline
& Mass conserving neural operator & \S \ref{Subsection:Mass conserving DeepONet} \\ \hline
& Recursive/Autoregressive prediction & \S \ref{Subsection:Recursive prediction} \\ \hline 
FNO & AMORE implementation in FNO & \S \ref{Section:AMORE:FNO:Syngas:FNO}  \\ \hline
ResNet & Discussion on ResNet model considered in the trunk and branch network & \S \ref{Appendix:Subsection:ResNet} \\ \hline
KAN & Discussion on KAN model considered in the trunk network & \S \ref{Appendix:Subsection:KAN} \\ \hline
\end{tabular}
\end{table} 
\subsection{Deep Operator Network (DeepONet)}
\label{Subsection:DeepONet}
\begin{figure}[H]
    \centering
    \includegraphics[width=1\textwidth]{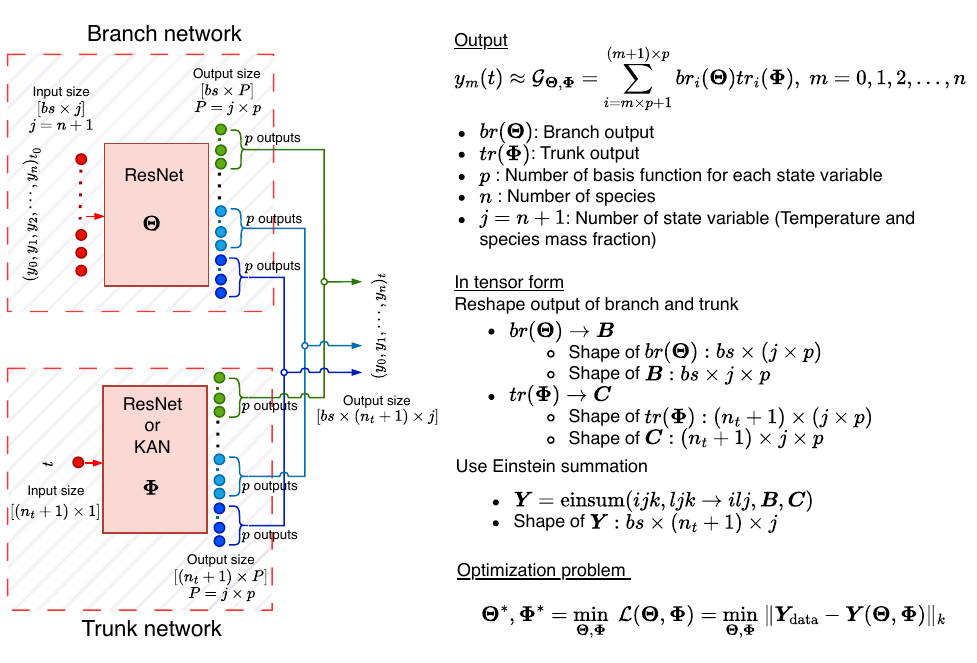}
    \caption{\textbf{Schematic diagram of the proposed DeepONet} architecture showing branch and trunk networks and multiple outputs. We have shown the branch network in the top left in the red dashed box. The inputs to the branch are the initial conditions of the state variables $\bm{Y}_0\in \mathbb{R}^{bs\times j}$ and the outputs of the branch are $br(\bm{\Theta})\in \mathbb{R}^{bs\times P}$. Where $bs$ and $j$ are the number of samples and the number of state variables, respectively, and $P$ is the number of output neurons. The input to the trunk is time $t\in\mathbb{R}^{(n_t+1)\times 1}$ and the outputs of the trunk are $tr(\bm{\Phi})\in \mathbb{R}^{(n_t+1)\times P}$. Where $(n_t+1)$ is the number of time points where the output needs to be predicted. In the present study, we have considered ResNet (with trainable parameters $\bm{\Theta}$) as a choice of network architecture for the branch network and ResNet or KAN (with trainable parameters $\bm{\Phi}$) as a choice of network architecture for the trunk network. A brief discussion on the ResNet and KAN architectures is included in \Cref{Appedix:Network Architecture}.  The branch and trunk network each has $P = j\times p$ neurons in the output layers, where $j$ is the number of state variables and $p$ is the number of basis functions and coefficients considered to approximate each variable. The output $\bm{Y}\in\mathbb{R}^{bs\times (n_t+1)\times j}$ is given by the dot product between the output of the branch and trunk networks, and is shown on the right side in standard form and tensor form. We also implemented the partition of unity in the trunk network, and the implementation is discussed in \Cref{Subsubsection:PoU}. Furthermore, we also consider a loss due to the CoM along with data loss, which is discussed in \Cref{Subsubsection:CoM}. As discussed in the AMORE framework, we propose two adaptive loss functions, and these are discussed in \Cref{Subsection:Adaptive loss function}.}
    \label{Figure:Schematic DeepONet}
\end{figure}

To learn the dynamics of the chemical kinetics governed by the system of stiff ODEs given by \Cref{Eq:ODE general}, we designed a DeepONet \citep{Lu_2021_DeepONet} with a single trunk network and a single branch network. The DeepONet predicts the dynamics of the multiple state variables of the stiff ODEs, i.e., the temperature and mass fractions of the species involved in the reaction. Thus, the output of the DeepONet is $\bm{Y}(t) = [y_0(t), y_1(t), y_2(t), \dots y_n(t)]^\top$, where $y_0$ is the temperature ($T$), and $y_1$, $y_2$, $\dots$, $y_n$ are the mass fractions of the $n$ number of species participating in the chemical process. The branch takes the initial conditions, i.e., initial temperature, and the initial mass fractions of the species as input. Thus, the branch input is $\bm{Y}_0 = [y_0, y_1, y_2, \dots y_n]_0^\top$, while the trunk takes the time $t$ as the input where the outputs need to be predicted. Thus, the operator learning can be written as 
\begin{equation}
    \bm{Y}(t)\approx \mathcal{G}_{\bm{\Psi}}: \bm{Y}_0 \mapsto \bm{Y}(\bm{Y}_0)(t),
\end{equation}
where $\bm{\Psi}$ are the parameters of the DeepONet. A dot product between the output of the branch and the trunk networks gives the output of the DeepONet. As discussed earlier, we will refer to the temperature and species' mass fractions together as state variables. Since we are predicting $j=n+1$ state variables from a single DeepONet, we divide the output neurons of the branch and trunk networks into $j=n+1$ equal parts, each part corresponding to one state variable. Thus, the approximation for the dynamics of the state variables using DeepONet can be written as 
\begin{equation}
    y_{m}(t) \approx \mathcal{G}_{\bm{\Theta}, \bm{\Phi}} =  {\displaystyle{\sum_{i=m\times p+1}^{(m+1)\times p}}}br_i(\bm{\Theta})tr(\bm{\Phi})_i,\;\;\;\;\; m=0,1,2,\dots,n
    \label{Eq:DeepONet normal}
\end{equation}
where $br(\bm{\Theta}) \in \mathbb{R}^{bs\times (n+1)p}$ and $tr(\bm{\Phi})\in \mathbb{R}^{(n_t+1)\times (n+1)p}$ are the output of the branch and trunk with $P=(n+1)p=jp$ output neurons, i.e., we consider $p$ individual basis functions and coefficients for each state variable. $\bm{\Theta}$ and $\bm{\Phi}$ are the parameters of the branch and trunk network, respectively, i.e., $\bm{\Psi} = \{\bm{\Theta}, \bm{\Phi}\}$. We can write \Cref{Eq:DeepONet normal} in tensor notation by reshaping the output of branch $br(\bm{\Theta})$ to $\bm{B}$ and trunk $tr(\bm{\Phi})$ to $\bm{C}$ and using Einstein summation as 
\begin{equation}
\begin{split}
    \bm{Y} \approx \mathcal{G}_{\bm{\Theta}, \bm{\Phi}} =\; &\text{einsum}(ijk,ljk\rightarrow ilj, \bm{B}, \bm{C}), \\
    & \bm{Y}\in\mathbb{R}^{bs\times (n_t+1)\times (n+1)},\;\;\; \bm{B}\in \mathbb{R}^{bs\times (n+1)\times p}, \;\;\; \bm{C}\in \mathbb{R}^{(n_t+1)\times (n+1)\times p}
\end{split}
\label{Eq:DeepONet einsum}
\end{equation}
where in the Einstein summation, the $i$ indicates the number of batches in the branch (sample size), $j$ indicates the number of state variables, $k$ indicates the number of basis functions for each state variable, and $l$ indicates the number of time points where the output has to be evaluated (trunk input). Furthermore, $j\times k$ is the number of output neurons in the branch and trunk networks. For convenience, we have not indicated the parameters $\bm{\Theta}$ and $\bm{\Phi}$ in $\bm{B}$ and $\bm{C}$. Furthermore, $\bm{Y}$ is a 3D tensor of size $bs\times (n_t+1)\times(n+1)$, and we consider that the batch (samples) vary from $bs \sim 1, 2, \cdots bs$, the time varies from $n_1+1\sim 0, \Delta t, 2\Delta t,\cdots, n_t\Delta t $, and the state variables from $n+1\sim 0, 1, 2, \cdots j-1$ (i.e., from $n+1\sim 0, 1, 2, \cdots n$). As discussed earlier, the DeepONet architecture does not restrict the type of neural network architecture considered in the branch and trunk networks. In the present study, we consider two types of DeepONet architectures. In the first case, we consider ResNet (with MLP) in both the branch and trunk networks, which we refer to as DeepONet, in short, DON. In the second case, we consider ResNet in the branch while we consider KAN in the trunk network, which we call DeepOKAN, in short, DOK. We will use the term DeepONet as a general term to indicate both DON and DOK. A brief discussion on ResNet and KAN architecture is included in \Cref{Appedix:Network Architecture}. A schematic representation of the proposed DeepONet architecture is shown in \Cref{Figure:Schematic DeepONet} with input and output data structures.
\par The optimal parameters of the DeepONet ($\bm{\Psi}=\{\bm{\Theta}, \bm{\Phi}\}$) can be obtained by minimizing a loss function,
\begin{equation}
    \bm{\Theta}^*, \bm{\Phi} ^* = \min_{\bm{\Theta}, \bm{\Phi}}\;\mathcal{L}(\bm{\Theta}, \bm{\Phi}) = \min_{\bm{\Theta}, \bm{\Phi}}\;\lVert \bm{Y}_{\text{data}}-\bm{Y}(\bm{\Theta}, \bm{\Phi}) \rVert_{k},
    \label{Eq:Optimize DON single step}
\end{equation}
where $\bm{Y}_{\text{data}}$ is the labeled data and the $\bm{Y}(\bm{\Theta}, \bm{\Phi})$ is the DeepONet prediction, $\lVert \cdot \rVert_k$ is a chosen norm to measure the discrepancy between the two. For convenience, from this point forward, we will denote the true value as $\bm{Y}$ and the predicted value as $\widehat{\bm{Y}}$ (with a hat). In the present study, for the optimization process, we consider the Adam \cite{Kingma_2014adam} optimizer in a TensorFlow version 2 environment with 64-bit data precision. We consider the data loss as the mean square of discrepancies between the predicted and true values,
\begin{equation}
    \mathcal{L}_{\text{data}}(\bm{\Theta}, \bm{\Phi})
    = \dfrac{1}{j\times bs\times (n_t+1)}\sum_{a=0}^{j-1}\sum_{b=1}^{bs}\sum_{c=0}^{n_t}\left(Y_{bca} - \widehat{Y}_{bca}\right)^2, \;\;\;\;\;\; 
    \begin{array}{l}
    j\rightarrow \text{No. of states}, \\
    n_t+1\rightarrow \text{No. of time points} \\
    bs\rightarrow \text{No. of samples} \\
    \end{array}
\label{Eq:Loss:Data_loss}
\end{equation}
We also like to mention that we consider the subscripts for time point and species starts from zero ($0$) to match the previous nomenclature used.
\subsection{Partition of Unity and conservation of mass}
\label{Subsection:POU and CoM}
\subsubsection{Partition of Unity (PoU)}
\label{Subsubsection:PoU}
Partition of unity (PoU) \citep{Babuvska_1997_PoU} allows us to stitch locally well-defined approximation spaces to give them a conforming global definition of desired regularity. PoU is defined such that the basis functions are bounded $[0,1]$ and the sum of the basis functions at any point is one, i.e., for the trunk output for each state variable, 
\begin{subequations}
\begin{align}
    \label{Eq:PoU def 1}
    0 \le tr_i(t)\le 1, & \;\;\;\; \forall \;t\in [0, n_t\Delta t], \;\;\;\; \forall \; i\in [1, p] \\
    \label{Eq:PoU def 2}
    \sum_i^p tr_i(t) = 1, & \;\;\;\;\forall\; t\in [0, n_t\Delta t]
\end{align}
\end{subequations}
\citeauthor{Lee_2021_PoU} \cite{Lee_2021_PoU}, \citeauthor{N_trask_2022_PoU} \cite{N_trask_2022_PoU} consider PoU in a neural network by considering a softmax function in the last hidden layer of a neural network. \citeauthor{Goswami_2024_Stiff_Kinetics} \cite{Goswami_2024_Stiff_Kinetics} considered PoU in DeepONet in the trunk network by incorporating a PoU-based additional loss function to satisfy the second equation of PoU, i.e., \Cref{Eq:PoU def 2} in the loss function as an additional penalty term. However, this does not ensure the bound ($[0,1]$) of the basis function. Furthermore, this also does not ensure the summation of the basis functions to be one, as it depends on the penalty term considered in the loss function for the PoU-based loss. \citeauthor{Sharma_2025_Ensemble} \cite{Sharma_2025_Ensemble} proposed an ensemble DeepONet with stacked trunk networks forming an ensemble trunk, and a PoU-mixture-of-experts (PoU-MoE) trunk architecture, wherein the domain is partitioned into overlapping spherical patches with a separate trunk on each of them, blended to form a single trunk network. The authors also noted that the PoU-MoE trunk incurs a higher computational cost due to the forward pass consisting of serial looping over the patches of the domain.

\par In the present study, we design our trunk network to automatically satisfy the PoU in the trunk network for each state variable, which are the basis functions in the approximation of the output. We consider the softmax function to enforce PoU in the trunk network. The output of the trunk network is passed through a softmax function, which ensures that both the basis functions are bounded in $(0,1)$, ($0 < tr_i(t)< 1, \forall \;t\in [0, n_t\Delta t], \forall \; i\in [1, p]$ ) and the sum of the basis functions as one ($\sum_i^p tr_i(t) = 1, \;\;\forall\; t\in [0, n_t\Delta t]$). Since we have divided the output of the trunk into as many numbers as the number of state variables, we have enforced the PoU for each state variable. This is implemented by considering the softmax function to $\bm{C}$ before \Cref{Eq:DeepONet einsum} along the basis dimension. Thus,
\begin{equation}
    \bm{C}_{ljk} = \text{softmax}(\bm{C}_{ljk}, \;\; \text{axis}=k).
    \label{Eq:Partition of Unity}
\end{equation}
where $l$ is the number of time points where the output has to be evaluated (trunk input), $j$ is the number of state variables, and $k$ is the number of basis functions for each state variable.
\subsubsection{Conservation of mass}
\label{Subsubsection:CoM}
Under the assumption of a zero-dimensional reactor made in \Cref{Subsection:Problem statement}, we know that the chemical reactions are balanced, i.e., the total mass of the reacting agents and products at any time is the same, and the sum of mass fractions is always unity. Thus, we consider an additional loss function corresponding to the CoM defined as,
\begin{equation}
        \mathcal{L}_{\text{CoM}}(\bm{\Theta}, \bm{\Phi}) = \dfrac{1}{bs\times (n_t+1)}\sum_{b=1}^{bs}\sum_{c=0}^{n_t}\left(\sum_{a=1}^{j-1}\widehat{Y}_{bca} - 1.0\right)^2, \;\;\;\;\;\; \begin{array}{l}
j\rightarrow \text{No. of states variable}, \\
n_t+1\rightarrow \text{No. of time points} \\
bs\rightarrow \text{No. of samples} \\
\end{array}
\label{Eq:Loss conservation of mass}
\end{equation}
We would like to draw attention to the fact that the summation over the state variables starts from $a=1$, as the state variable corresponding to $a=0$ is temperature. Secondly, if any normalization scheme is considered, then the summation in \Cref{Eq:Loss conservation of mass} needs to be calculated in the physical domain. The total loss function is
\begin{equation}
    \mathcal{L}(\bm{\Theta}, \bm{\Phi}) = \mathcal{L}_{\text{data}}(\bm{\Theta}, \bm{\Phi}) + \mathcal{W}_{\text{CoM}}\mathcal{L}_{\text{CoM}}(\bm{\Theta}, \bm{\Phi})
    \label{Eq:Combined loss data+CoM}
\end{equation}
where, $\mathcal{L}_{\text{data}}(\bm{\Theta}, \bm{\Phi})$ is the data loss discussed in \Cref{Eq:Loss:Data_loss} or later for adaptive data loss functions, and $\mathcal{L}_{\text{CoM}}(\bm{\Theta}, \bm{\Phi})$ is the loss due to CoM and defined in \Cref{Eq:Loss conservation of mass}, $\mathcal{W}_{\text{CoM}}$ is a relative weight term which regulates the contribution of the CoM loss in the loss function ($\mathcal{L}(\bm{\Theta}, \bm{\Phi})$). We will discuss the weight, $\mathcal{W}_{\text{CoM}}$, in detail in the next section.

\subsection{Adaptive loss function}
\label{Subsection:Adaptive loss function}
Most deep learning algorithms involve the minimization of a cost or a loss function, characterized by the task to which they are employed. In many cases, it is observed that the error is not uniform over different data points. For instance, solving a forward problem using a physics-informed network (PINN) \citep{Raissi_2019_PINN} involves residual, initial/boundary condition losses. The errors at different collocation points are different. To address this issue \citeauthor{Mcclenny_2023_self_adaptive} \cite{Mcclenny_2023_self_adaptive} proposed self-adaptive weights (SAW) in PINNs for each collocation, initial/boundary points. The idea of the self-adaptive weights is that the points where the loss is greater should be weighted more. The weights are updated using a min-max problem with a gradient-based optimizer. While the loss is minimized with respect to network parameters, it is maximized with respect to the self-adaptive weights. \citeauthor{Anagnostopoulos_2024_RBA} \cite{Anagnostopoulos_2024_RBA} proposed residual-based attention (RBA), where instead of updating the adaptive weights using a gradient-based optimizer, the authors propose a gradient-free update scheme which is based on the residual from the previous epoch. The idea of self-adaptive weights in the loss functions has been extended to the DeepONet \citep{Goswami_2023_Physics,Kontolati_2023} to enhance the robustness of these models. In \citep{Goswami_2023_Physics}, each adaptive weight is associated with the corresponding evaluation point. Furthermore, in both cases (SAW, RBA), additional hyperparameters are needed, such as learning rate and decay parameters. 
\par In the present study, the proposed DeepONet architecture predicts multiple state variables; consequently, the errors for each of these state variables often differ, sometimes in the order of magnitude. Thus, considering a single global data loss function defined in the previous section does not do justice to all the state variables. This results in an uneven distribution of error across samples of different state variables, as also observed in numerical examples later. This can lead to poor model predictions. Thus, we need to design loss functions to account for such unevenness in distribution by giving more weight to specific state variables and their corresponding samples with higher error values. This will lead to a balanced loss function. Based on this, we propose two gradient-free adaptive weighting schemes in the loss function such that state variables and the samples with the higher error values are penalized more. This will enhance the prediction accuracy due to a balanced loss function. The proposed adaptive loss functions do not require any additional hyperparameters, such as an additional learning rate.
\subsubsection{Adaptive loss function: Type-A}
\label{Subsubsection:Adaptive loss:Type-A}
The first adaptive loss function we propose is based on the idea that the state variables with higher error values should have more weight in the loss function, so that they are penalized more heavily. Thus, we define the data loss function as \Cref{Eq:Adaptive loss:Type-A}, with adaptive weights $\mathcal{W}_a$, 
\begin{equation}
    \mathcal{L}_{\text{data}}(\bm{\Theta}, \bm{\Phi}) = \sum_{a=0}^{j-1}\mathcal{W}_{a}\left[\dfrac{1}{bs\times (n_t+1)}\sum_{b=1}^{bs}\sum_{c=0}^{n_t}\left(Y_{bca} - \widehat{Y}_{bca}\right)^2\right], \;\;\;\;\;\; \begin{array}{l}
    j\rightarrow \text{No. of states variable}, \\
    n_t+1\rightarrow \text{No. of time points} \\
    bs\rightarrow \text{No. of samples} \\
    \end{array}
\label{Eq:Adaptive loss:Type-A}
\end{equation}
such that $\sum_{a=0}^{j-1}\mathcal{W}_{a} = \mathcal{R}_1$, and $\mathcal{W}_a\ge 0$ for every epoch. The subscript in $\mathcal{W}_a$ are the weights associated with the state variable '$a$', where $a = 0,1,\dots, j-1$. We proposed a gradient-free updating scheme with no additional hyper-parameter requirement to update the weights. The weights are updated based on the relative error between the state variables in the previous epoch, such that 
\begin{equation}
    \mathcal{W}_a = \dfrac{X_a}{\displaystyle\sum_{d=0}^{j-1}X_d} \times \mathcal{R}_1
    \label{Eq:Adaptive loss:Type-A W_a}
\end{equation}
We consider $X_a$ (and $X_d$) as the mean (calculated over samples) of the relative $L_2$ error (calculated over time points) in the physical space for each state variable, i.e., we calculate the relative $L_2$ error in the time direction $n_t+1$ and then take the mean in the batch direction $bs$ for each state variable. This way, we calculate $j$ means of the relative $L_2$ error corresponding to $j$ state variables. We initialize the weights with an equal value of one, i.e., $\mathcal{W}_a=1$. Thus, $\sum_{a=0}^{j-1}\mathcal{W}_{a} = \mathcal{R}_1 = j $, which is the number of state variables. We also note that the limiting value of $\mathcal{W}_a$ can be zero only when $X_a$ is zero, i.e., the relative $L_2$ error is zero. We also assume that all $X_a$ are not zero at the same epoch so that the denominator of \Cref{Eq:Adaptive loss:Type-A W_a} is non-zero. Furthermore, the sum of all the weights $\mathcal{W}_a$ at any epoch is always $\mathcal{R}_1$ as shown in \Cref{Eq:Adaptive loss:Type-A W_a sum} (in \Cref{Appendix:SectionCalculation for adaptive loss function}).
\subsubsection{Adaptive loss function: Type-B}
\label{Subsubsection:Adaptive loss:Type-B}
We propose a second variant of the adaptive loss function based on the idea that, along with the state variables, the samples with higher error values should be given more weight, hence penalized more heavily, in the loss function. Thus, we defined the data loss function as \Cref{Eq:Adaptive loss:Type-B}, with an adaptive weight $\mathcal{W}_{ba}$, 
\begin{equation}
    \mathcal{L}_{\text{data}}(\bm{\Theta}, \bm{\Phi}) = \sum_{a=0}^{j-1}\sum_{b=1}^{bs}\mathcal{W}_{ba}\left[\dfrac{1}{ n_t+1}\sum_{c=0}^{n_t}\left(Y_{bca} - \widehat{Y}_{bca}\right)^2\right], \;\;\;\;\;\; \begin{array}{l}
    j\rightarrow \text{No. of states variable}, \\
    n_t+1\rightarrow \text{No. of time points} \\
    bs\rightarrow \text{No. of samples} \\
    \end{array}
\label{Eq:Adaptive loss:Type-B}
\end{equation}
such that $\sum_{a=0}^{j-1}\sum_{b=1}^{bs}\mathcal{W}_{ba} = \mathcal{R}_2$, and $\mathcal{W}_{ba}\ge0$ for every epoch. $\mathcal{W}_{ba}$ represents the weights associated with the state variable $a$ and the sample $b$. Similar to ``Type-A" adaptive loss, the weights $\mathcal{W}_{ba}$ are updated using a gradient-free algorithm. The weights are updated based on the relative error between the state variables and samples in the previous epoch, such that
\begin{equation}
    \mathcal{W}_{ba} = \dfrac{X_{ba}}{\displaystyle\sum_{d=0}^{j-1}\sum_{e=1}^{bs}X_{ed}} \times \mathcal{R}_2
    \label{Eq:Adaptive loss:Type-B W_ab}
\end{equation}
We consider $X_{ba}$ (and $X_{ed}$) as the relative $L_2$ error (calculated over time-points) in the physical space for each state variable and sample. In this way, we calculate $j\times bs$ relative $L_2$ error corresponding to the $j$ state variables and $bs$ samples. Similar to ``Type-A", we initialize the weights with an equal value of one, i.e., $\mathcal{W}_{ba} = 1$. Thus, $\sum_{a=0}^{j-1}\sum_{b=1}^{bs}\mathcal{W}_{ba}= \mathcal{R}_2 = j\times bs$. The limiting value of $\mathcal{W}_{ba}$ can be zero only when $X_{ba}=0$. We assume that all $X_{ba}$ are not zero so that the denominator of \Cref{Eq:Adaptive loss:Type-B} is non-zero. Similar to \Cref{Eq:Adaptive loss:Type-A W_a sum}, we can show that the sum of all $\mathcal{W}_{ba}$ at any epoch is always $\mathcal{R}_2$.
\par A brief discussion on the calculation of $X_a$ and $X_{ba}$ is given in \Cref{Appendix:SectionCalculation for adaptive loss function}. As discussed earlier, we also consider the two-step training of DeepONet. Appropriate modifications are needed in the adaptive loss functions to incorporate them in two-step training. These modifications are discussed in the implementation of two-step training of DeepONet in \Cref{Subsection:Two-step training}. Furthermore, as discussed in \Cref{Subsubsection:CoM}, the weight $\mathcal{W}_{\text{CoM}}$ acts as a weighting coefficient for the loss due to the CoM ($\mathcal{L}_\text{CoM}(\bm{\Theta,\Phi})$). In the cases where both data loss and loss due to mass conservation are considered, we consider $\mathcal{W}_{\text{CoM}}$ as a multiplier of the weights ($\mathcal{W}_a$ or $\mathcal{W}_{ab}$) of the data loss.
\subsection{Two-step training of DeepONet}
\label{Subsection:Two-step training}
The training method for DeepONet discussed so far considers training the parameters of the branch ($\bm{\Theta}$) and trunk ($\bm{\Phi}$) together as shown in \Cref{Eq:Optimize DON single step}. \citeauthor{Sanghyun_2024_two_step} \cite{Sanghyun_2024_two_step} proposed a novel reparameterization and training strategy for DeepONet. The method consists of re-parameterizing the existing DeepONet architecture and a two-step training of DeepONet, where the parameters of the trunk and branch are trained in a sequential manner rather than together. The parameters of the trunk are trained first, followed by training the parameters of the branch. \citeauthor{Sanghyun_2024_two_step} \cite{Sanghyun_2024_two_step} considered DeepONet with a single output variable. \citeauthor{Ahmad_2024_RiemannONets} \cite{Ahmad_2024_RiemannONets} considered DeepONet with multiple output variables. The authors considered the same basis function for all the output variables with different coefficients from a single-branch network. \citeauthor{Kiyani_2024_Crack} \cite{Kiyani_2024_Crack} also considered multiple output variables with the same basis function in a trunk in crack propagation problems. The authors considered independent branches, one for each output variable. \citeauthor{Jin_2025_Characterization} \cite{Jin_2025_Characterization} considered two-step training of DeepONet for multiple outputs with a single trunk with different basis functions. The coefficients of the basis functions are obtained from a composite branch with an independent MLP for each output variable.
\par In the present study, we consider a generalized form of two-step training of DeepONet for multiple output variables, with one trunk but different basis functions for each output variable and one branch but different coefficients for each output variable. This architecture, with different basis functions for each output variable from a single trunk and different coefficients for each output variable from a single branch, is important for the scalability of the method. Furthermore, to improve the accuracy of DeepONet using two-step training, we extend the adaptive loss functions discussed in the previous section to trunk and branch training with appropriate modifications. In the two-step training framework of DeepONets, we have not considered loss due to CoM or the PoU constraint. This is because the trunk and the branch are trained sequentially rather than together. The trunk is trained first on the labeled training data, followed by the branch, which is trained using the reconstructed output from the trunk. It is possible to consider CoM and PoU in this case; however, since we are using the QR decomposition of the trunk network and treating Q as a simplified trunk, the effective trunk network changes. During branch training, since it uses a reconstructed output, we have not considered the CoM loss.
\subsubsection{Training of trunk network}
\label{Subsubsection:Training of trunk network}
\par The first step in two-step training is the training of the parameters of the trunk ($\bm{\Phi}$) and a trainable tensor $\bm{A}$ through a minimization problem
\begin{equation}
    \bm{\Phi}^*, \bm{A} ^* = \min_{\bm{\Phi}, \bm{A}}\;\mathcal{L}(\bm{\Phi}, \bm{A}) = \min_{\bm{\Phi}, \bm{A}}\;\lVert \bm{Y}_{\text{data}}-\bm{Y}(\bm{\Phi}, \bm{A}) \rVert_{k},
    \label{Eq:2-step:trunk optimize}
\end{equation}
where $\bm{Y}_{\text{data}}$ is the labeled (DeepONet output training) data and $\bm{Y}(\bm{\Phi}, \bm{A})$ is the predicted output from the trunk multiplied by a trainable tensor $\bm{A}$. Since we predict multiple outputs from a single DeepONet, let us discuss this multiplication before proceeding. We have shown a schematic diagram for the two-step training of DeepONet in \cref{Figure:2-step DeepONet training} (in \Cref{Appendix:Subsection:Two-step training}). For the trunk training, each of the output state variables is approximated as 
\begin{equation}
\label{Eq:2-Step trunk output normal}
\begin{split}
    & y_{m}(t) \approx tr(\bm{\Phi})\left[:,\: m\times p +1:(m+1)\times p\right] \bm{A}_m, \;\;\;\;\;\;\;\;\;\; m = 0, 1, 2, \dots n, \;\;\;\;\; \\ & \hspace{3cm} tr(\bm{\Phi})[:,\: m\times p +1:(m+1)\times p] \in \mathbb{R}^{(n_t+1)\times p}, \;\;\;\;\;  \bm{A}_m\in \mathbb{R}^{p\times bs}
\end{split}
\end{equation}
where $tr(\bm{\Phi})\in \mathbb{R}^{(n_t+1)\times (n+1)p}$ is the output of the trunk with $P=(n+1)p=jp$ output neurons, i.e., we consider $p$ individual basis functions for each state variable, $\bm{A}\in \mathbb{R}^{(n+1)\times p\times bs}$ is a trainable tensor, where $bs$ is the number of samples and $j$ is the number of state variables.
\par Similar to the formulation discussed in \Cref{Subsection:DeepONet} for one step (conventional) training of DeepONet, we can rewrite the \Cref{Eq:2-Step trunk output normal} in tensor form using the Einstein summation. We reshape the output of the trunk network $tr(\bm{\Phi})$ to $\bm{C}$ and using Einstein summation as
\begin{equation}
\label{Eq:2-Step trunk output einsum}
\begin{split}
    \bm{Y} \approx \bm{Y}(\bm{\Phi}, \bm{A}) = & \;\text{einsum}(ijk, jkl\rightarrow lij, \bm{C},\bm{A}), \\ 
    & \bm{Y}\in\mathbb{R}^{bs\times (n_t+1)\times (n+1)},\;\;\; \bm{C}\in \mathbb{R}^{(n_t+1)\times(n+1)\times p}, \;\;\; \bm{A}\in \mathbb{R}^{(n+1)\times p\times bs}
\end{split}
\end{equation}
where in the Einstein summation, the $i$ is the number of time points (trunk input), $j$ is the number of state variables, $k$ is the number of basis functions for each state variable, and $l$ is the number of samples (i.e., number of time histories per batch)
\par We optimize the parameters of the trunk ($\bm{\Phi}$) and $\bm{A}$ using an appropriate loss function as shown in \Cref{Eq:2-step:trunk optimize}. We consider three different loss functions, one non-adaptive and two adaptive loss functions similar to those discussed in one-step training given by \Cref{Eq:Loss:Data_loss,Eq:Adaptive loss:Type-A,Eq:Adaptive loss:Type-B} where the $Y_{bca}$ is the labeled output data corresponding to different samples and $\widehat{Y}_{bca}$ is the prediction given by \Cref{Eq:2-Step trunk output einsum}. The weights in the adaptive loss function case were calculated as discussed in \Cref{Subsubsection:Adaptive loss:Type-A,Subsubsection:Adaptive loss:Type-B} for adaptive loss Type-A and Type-B, respectively. We used the Adam \cite{Kingma_2014adam} optimizer in a TensorFlow version 2 environment with 64-bit data precision to optimize the trunk parameters $\bm{\Phi}$ and $\bm{A}$. We also assume $p<n_t$ and $tr(\bm{\Phi}^*)$ is full-rank for each output variable \citep{Sanghyun_2024_two_step}.
\subsubsection{Training of branch network}
\label{Subsubsection:Training of branch network}
\par After the training of the trunk, the next step is the training of the branch network. The parameters of the branch network ($\bm{\Theta}$) are optimized considering a data loss function. The labeled data for the loss function is obtained from a reconstructed output from the trunk output $\bm{U}=\bm{R}^*\bm{A}^*$, where $\bm{R}^*$ is obtained from the QR decomposition of the trunk output. We also consider $\bm{A}^*$ from the least-squares approximation of the trunk output and labeled data. As we are predicting multiple outputs, let us discuss the process for calculating the labeled data ($\bm{U} = \bm{R}^*\bm{A}^*$) for the branch training. The QR decomposition of the trunk for each state variable can be written as,
\begin{equation}
    \bm{Q}^*_m\bm{R}^*_m = \text{QR}\left(tr(\bm{\Phi}^*)[:, \; m\times p+1: (m+1)\times p]\right),\;\;\;\;\; m = 0,1,2,\cdots, n
\end{equation}
and the least-square approximation of the $\bm{A}^*$ can be written as
\begin{equation}
    \bm{A}^*_m = \text{LSM}\left(tr(\bm{\Phi}^*)[:, \; m\times p+1: (m+1)\times p], \;y_m\right),\;\;\;\;\;\;\;\; m = 0, 1, 2, \cdots, n
    \label{Eq:A star LSM}
\end{equation}
where $y_m\in \mathbb{R}^{(n_t+1)\times bs}$ is the labeled data corresponding to $m$\textsuperscript{th} state variables. Note that we consider the $\bm{A}^*$ using \Cref{Eq:A star LSM} instead of the optimized $\bm{A}^*$ from \Cref{Eq:2-step:trunk optimize}, as in this case, we observed smaller error in the trunk error. The branch network is trained by minimizing a loss function 
\begin{equation}
    \mathcal{L}(\bm{\Theta}) = \rVert br(\bm{\Theta}) - \bm{U} \rVert
\end{equation}
where $br(\bm{\Theta})\in\mathbb{R}^{bs\times(n+1)p}$ and $\bm{U}$ is reconstructed data calculated from the output of the trunk network,
\begin{equation}
    \bm{U}_m = \bm{R}^*_m\bm{A}^*_m, \;\;\;\;\;\;\; m = 0, 1, 2, \cdots, n
\end{equation}
We rewrite the above in tensor notation by reshaping the branch output. The $\bm{U}\in\mathbb{R}^{(n+1)\times p\times bs}$, and $br(\bm{\Theta})\in \mathbb{R}^{bs\times (n+1)p}$. We reshape the branch output $br(\bm{\Theta})$ to $\bm{B}\in \mathbb{R}^{bs\times (n+1)\times p}$, the axis of $\bm{U}$ is changed to $\bm{U}\in\mathbb{R}^{bs\times (n+1)\times p}$.
\par The parameters ($\bm{\Theta}$) of the branch are optimized using a loss function between the branch output and $\bm{U}$,
\begin{equation}
    \bm{\Theta}^* = \min_{\bm{\Theta}}\;\mathcal{L}(\bm{\Theta}) = \min_{\bm{\Theta}}\;\lVert \bm{B}(\bm{\Theta})-\bm{U} \rVert_{k}
\end{equation}
Similar to previous optimization processes, we consider three different loss functions: the MSE loss function and two adaptive loss functions. However, in this case, marginal modifications are required in the adaptive loss functions. These are discussed in \Cref{Appendix:Subsection:Two-step training}.
\subsubsection{Prediction using trained DeepONet}
\label{Subsubsection:Prediction using trained DeepONet}
In the previous sections, we have discussed the training of the trunk and branch using the two-step training method of DeepONet, where the parameters of the trunk and branch networks are trained sequentially rather than together. The output of DeepONet, in this case, is approximated as \citep{Sanghyun_2024_two_step},
\begin{equation}
\begin{split}   
    y_m & = tr(\bm{\Phi})[:, m\times p+1:(m+1)\times p] \bm{T}_m br^{\top}(\bm{\Theta})[:, m\times p+1:(m+1)\times p], \label{Eq:Sub:Two-step-predict-1}\\
    & \hspace{5cm}\text{where}\;\;\;\; \bm{T}_m = {\bm{R}^*}^{-1}_m,\;\; m = 0, 1, 2\cdots,n 
\end{split}
\end{equation}
where $tr(\bm{\Phi})[:, m\times p+1:(m+1)\times p]\in\mathbb{R}^{(n_t+1)\times p}$ and $br(\bm{\Theta})[:, m\times p+1:(m+1)\times p]\in\mathbb{R}^{bs\times p}$. In the present study, we consider that the outputs are predicted at the same temporal points for all the samples during training and testing. Thus, we consider a simplified version of \Cref{Eq:Sub:Two-step-predict-1} as, 
\begin{subequations}
\begin{align}    
    y_m & = \bm{Q}^*_m\bm{R}^*_m\bm{T}_m br^{\top}(\bm{\Theta})[:, m\times p+1:(m+1)\times p]), \;\;\; m = 0, 1, 2, \cdots n \label{Eq:Sub:Two-step-predict-2}\\
    & = \bm{Q}^*_m br^{\top}(\bm{\Theta})[:, m\times p+1:(m+1)\times p], \;\;\;\;\;\; \text{since}\;\;\; \bm{R}^*_m\bm{T}_m = \bm{I} \label{Eq:Sub:Two-step-predict-3}
\end{align}
\end{subequations}
where $\bm{Q}^*_m\in\mathbb{R}^{(n_t+1)\times p}$ and $\bm{R}^*_m\in\mathbb{R}^{p\times p}$. We can interpret \Cref{Eq:Sub:Two-step-predict-3} as $\bm{Q}^*_m$ are the orthogonal basis functions obtained from the output dataset, and the branch gives the coefficient to these basis functions. We can rewrite \Cref{Eq:Sub:Two-step-predict-3} in a tensor form by reshaping the output of the branch and using the Einstein summation,
\begin{equation}
\label{Eq:2step DeepONet einsum}
\begin{split}
    \bm{Y} \approx \mathcal{G}_{\bm{\Theta}, \bm{\Phi}} = & \; \text{einsum}(ijk,jlk \rightarrow ilj, \bm{B}, \bm{Q}^*), \\ 
    & \bm{Y}\in\mathbb{R}^{bs\times (n_t+1)\times (n+1)}, \;\;\; \bm{B}\in \mathbb{R}^{bs\times (n+1)\times p}, \;\;\; \bm{Q}^*\in \mathbb{R}^{(n+1)\times (n_t+1)\times p}
\end{split}
\end{equation}
where in the Einstein summation, the $i$ is the number of batches in the branch (sample size), $j$ is the number of state variables, $k$ is the number of basis functions for each species, and $l$ is the number of time points where the output has to be evaluated (trunk input). We would also like to mention that in the calculation of the weights $\mathcal{W}_a$ and $\mathcal{W}_{ba}$ in the case of branch training, we use the predicted $\bm{Y}$ using \Cref{Eq:2step DeepONet einsum} for relative $L_2$ error calculation associated with each state variables and samples. In this case of weight calculation, the $\bm{Q}^*$ is constant (fixed), and the branch outputs $\bm{B}$ change with epochs.
\subsection{Mass conserving DeepONet}
\label{Subsection:Mass conserving DeepONet}
In the previous sections, we have proposed DeepONet for stiff chemical systems, including two-step training and adaptive loss functions. In \Cref{Subsubsection:CoM}, we have proposed the CoM for stiff chemical systems by incorporating an additional loss function due to the conservation of mass. However, the accuracy depends on the penalty imposed on the loss function, and the CoM is enforced only weakly. Furthermore, we did not consider the CoM in the two-step training of DeepONet. In this section, we will develop a DeepONet model that automatically satisfies the CoM. 
\begin{figure}[H]
    \centering
    \includegraphics[width=0.75\textwidth]{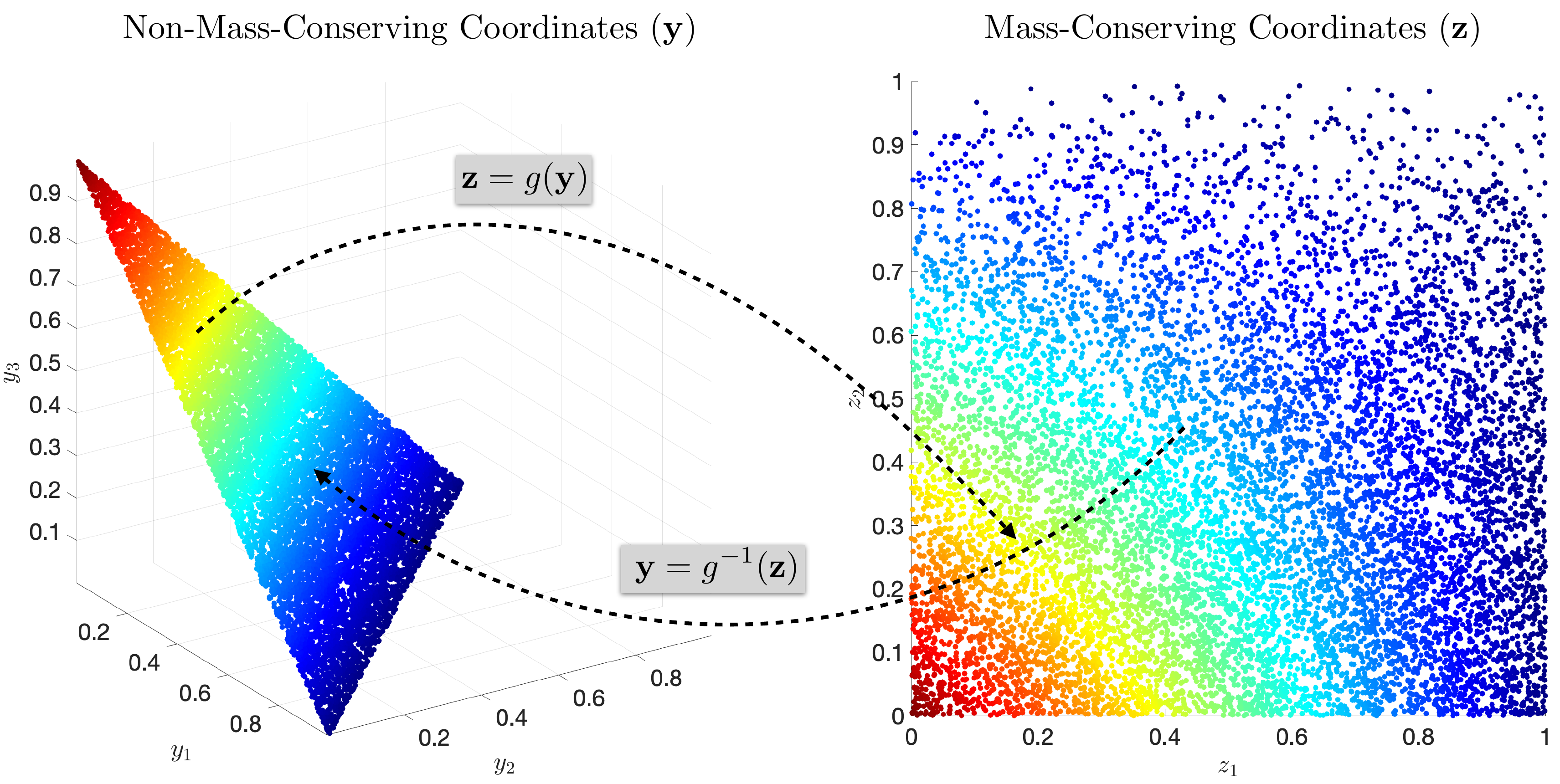}
    \caption{\textbf{Mass conserving map} showing the species mass fraction vector ($\bm{y} \in \mathbb{R}^n$) is mapped uniquely to a mass-conserving coordinate ($\bm{z} \in \mathbb{R}^{n-1}$) and vice versa.} 
    \label{fig:massconmap}
\end{figure}
\par The CoM is expressed as a linear constraint on the mass fraction
\begin{equation}
\sum_{i=1}^n y_i = 1.
\end{equation}
where $0 \leq y_i \leq 1$. As a result of the above constraint, the mass fraction vector $\bm{y}$ cannot lie anywhere within the $n$-dimensional hypercube; instead, it must reside within a simplex to ensure the conservation of mass.
\par Given the above constraint, it is evident that the species mass fraction vector can be encoded, without any loss of information, with an $(n-1)$-dimensional vector  $\mathbf z = (z_1,z_2, \cdots, z_{n-1})$ as follows: 
\begin{equation}
    z_1, z_2, \cdots z_{n-1} = g(y_1, y_2, \cdots y_{n})
    \label{Eq:CoM DeepONet forward map}
\end{equation}
where the function $g(.)$ can be considered as an encoder of the input data. For reasons that will become clearer later in this section, we also require that the function $g(.)$ be invertible, such that
\begin{equation}
    y_1, y_2, \cdots y_{n} = g^{-1}(z_1, z_2, \cdots z_{n-1})
    \label{Eq:CoM DeepONet inverse map}
\end{equation}
where  $0 \leq z_i \leq 1$. Therefore, any mass fraction vector ($\bm{y}$) can be uniquely mapped to a unique vector ($\bm{z}$) and vice versa. The schematic of such mapping for $n=3$ species is shown in \Cref{fig:massconmap}. We construct the maps using the popular spectral element simulation of PDES \citep{Karniadakis_2005_Spectral_book} and discuss it in detail in \Cref{Appendix:Subsection:Mass conserving DeepONet}
\subsection{Recursive prediction}
\label{Subsection:Recursive prediction}
As discussed in previous sections, the objective of the present study is to develop an accurate neural operator for stiff chemical kinetics so that it can be integrated with a CFD solver in future studies. In this context, it is possible that the time series predicted by the neural operator may not progress sufficiently in time to be considered as input to the CFD solver. In such a case, we propose considering a recursive scheme which predicts the output auto-regressively from the prediction.
\par As discussed in \Cref{Subsection:Problem statement}, \Cref{Figure:Problem statement}, the pre-trained neural operator predicts the time series from $t=0$ to $t=n_t\Delta t$. The output states at time $n_t\Delta t$, i.e., $\bm{Y}_{n_t}$ are considered as input to the same pre-trained neural operator again to predict the next time segment of the same length as the initial segment. The process continues until the predicted time series has progressed sufficiently to be considered as input to the CFD solver. We would like to mention two important aspects in this case. The first aspect is that an appropriate normalization and denormalization scheme must be considered while performing a recursive prediction. The second one being that the input to the trunk is the same $t=0, \Delta t, 2\Delta t, \cdots, n_t\Delta t$ in all recursive predictions. We can consider it as translating the time axis such that the end point of the previous time series becomes the first point of the newly translated time series. In the present study, since we have not integrated our operator with the CFD solver, we perform recursive predictions for the entire time series.
\subsection{Error metric}
\label{Subsection:Error matric}
In order to check the accuracy of DeepONet during training and testing, we consider the relative $L_2$ error defined as,
\begin{equation}
    \text{Relative}\;\; L_2\;\;\text{error} = \dfrac{||y - \hat{y}||_2}{||y||_2} = \dfrac{\sqrt{\sum\limits_{i=0}^{n_t} \left(y_{i} -  \hat{y}_{i}\right)^2}}{{\sqrt{\sum\limits_{i=0}^{n_t} y_{i}^2}}},
    \label{Eq:L2 error}
\end{equation}
where $n_t+1$ is the number of time points considered. It is important to note that the output of DeepONet and the true data are 3-dimensional, with size $bs\times (n_t+1)\times j$, where $bs$, $n_t+1$, and $j$ are the sample size, number of time points, and number of state variables, respectively. In \Cref{Eq:L2 error}, we consider the relative $L_2$ error only in the direction of time points. Thus, to get a global error,  we consider the mean of the relative $L_2$ error for each state variable and take the mean again to obtain an overall scalar error value, which is the global error. However, calculating a global error in this way may not reflect the robustness of the proposed method. Thus, we also study the mean and standard deviation of the relative $L_2$ error for each state variable, along with studying the violin plot of the relative $L_2$ error for each state variable. We would also like to mention that, unless otherwise stated, we calculate all the errors in the physical space, and not by using the normalized data.
\section{Computational experiment: Syngas problem}
\label{Section:Numerical results:Syngas Problem}
The applicability and accuracy of the methods proposed in this study are evaluated by considering computational examples. The first computational example considered is the syngas (synthetic gas) problem involving 11 species and 21 reactions \citep{Paykani_2022_Synthesis}. Syngas is a mixture of carbon monoxide (CO) and hydrogen (H$_2$), and often, carbon dioxide (CO$_2$). It has various applications, including usage as a fuel or an intermediate for producing chemicals. The total number of state variables for the syngas problem is 12 (i.e, temperature and 11 species' mass fractions).
\par  To train the proposed DeepONet, we generated samples of the dynamics of the system with different initial conditions of the state variables. We considered 51 initial temperature values uniformly spaced between 1000 and 1500 K. The initial conditions for the species' mass fractions are generated from equivalence ratios ranging from 0.7 to 1.3. We sampled 51 equivalence ratios uniformly spaced within the aforementioned range. Thus, we generated a total of $51\times51=2601$ different samples of initial conditions of the system, i.e., initial temperature and initial species mass fractions. We solved the ODEs (\cref{Eq:ODE general}) of the syngas problem using Cantera's \citep{Goodwin_2017_Cantera} implicit solver with the different initial conditions generated. Within the Cantera solver used for data generation, the time integration was performed using the VODE solver with a backward differentiation formula (BDF) scheme, suitable for stiff systems. The solver employs an implicit multi-step method with user-specified relative tolerance of $10^{-8}$ and absolute tolerances $10^{-12}$, and incorporates the Jacobian to enhance stability and efficiency. We consider a constant and the same pressure of 1.0 atm for all generated samples. The solution is obtained on a uniform time step of $\Delta t = 10^{-7}$ s. Thus, we generated a total of $2601$ sets of time series of the state variables with different initial conditions. We consider $2080$ randomly sampled time series for training, and $520$ non-overlapping testing samples.
\par The chemical process is a continuous-autonomous process, and our dataset starts from  $t_0 = \Delta t = 10^{-7} $ s to $t_{end} = 10^4\Delta t = 10^{-3}$ s. The total number of time steps is $10^{4}$ with a uniform $\Delta t=10^{-7}$ s. Without loss of generality, we translate the origin from zero to $\Delta t$. Thus, the time series data describing the dynamics of the species lie within the temporal range $t=[0, 10^{4}-1]\Delta t$. As discussed in \Cref{Subsection:Problem statement,Section:Methodology}, for the operator learning, we consider a time decomposition. We divide the time series into $99$ segments with each segment of length $100\times \Delta t$ s (i.e., consisting of $n_t+1 = 101$ time points) and assume that the initial conditions are known for each segment. We would also like to mention that in the time discretization, we have considered the last time point of a segment as the first time point of the next segment, with known initial conditions for each segment (generally coming from the CFD solver). The number of training and testing samples considered is $2080$ and $520$, respectively, which becomes $2080\times99=205920$ and $520\times99=51480$, respectively, after performing time decomposition.
\par We observed the training data set for the syngas problem; the temperature and mass fraction of the species have different orders of magnitude. The temperature is of the order $10^3$, and the mass fraction is in the smallest order of $10^{-17}$. The minimum and maximum values for the state variables corresponding to the training and testing datasets are shown in \Cref{Table:Syngas problem:Min max training-testing data} (in \Cref{Appendix:Subsection:Syngas problem}). In order to get a better accuracy of the DeepONet prediction, we normalized the output data as follows: we take the logarithm  (base 10) of the state variable and after that normalize each of the state variables separately between $[-1, 1]$ using
\begin{equation}
    y_{\text{Norm}} = 2\times\left(\frac{y - y_{\text{min}}}{y_{\text{max}} - y_{\text{min}}}\right) -1.
    \label{Eq:Normalization output}
\end{equation}
where $y$ and $y_{\text{Norm}}$ are the values of the state variable before and after the normalization, $y_{\text{max}}$ and $y_{\text{min}}$ are the maximum and value of the state variable (each state variable individually after taking the logarithm) respectively. Similarly, the inputs to the branch network are also normalized. The input to the trunk network is $t ={ 0, \Delta t, 2\Delta t \dots, 100\Delta t}$, where $\Delta t = 10^{-7}$ s, which is normalized to $[-1,1]$ using $t_0=0$ and $t_{n_t} = 100\Delta t$. In this case, we consider the total number of time points $n_t+1 = 101$.
\par We first discuss the training and the results for the standard DeepONet with one-step training, and then discuss the training and the results for two-step training. As discussed earlier, we consider two types of DeepONet architectures. In the first one, we consider ResNet as the choice of network architecture for both the trunk and branch networks, and call it DON. In the second one, we consider ResNet as the choice of network architecture for the branch and KAN as the choice of network architecture for the trunk, and call it DOK. The details of the network (size, activation function, etc.) considered are shown in \Cref{Table:Syngas: network size} (in \Cref{Appendix:Subsection:Syngas problem}). We train the DeepONets in both single-step and two-step training (including the $\bm{A}$ in two-step training) using the Adam \cite{Kingma_2014adam} optimizer in a TensorFlow 2 environment with 64-bit data precision. To assess the influence of DeepONet parameter initialization on accuracy, we run three independent training runs for each case. After that, we consider one of the trained models and its results for further analysis, such as reconstructed error. We update the self-adaptive weights, $\mathcal{W}_a$ and $\mathcal{W}_{ba}$ after $100$ epochs and at every $50$ epochs. We have included the convergence of relative $L_2$ error with epoch in \Cref{Appendix:Subsubsection:Error Convergence for Syngas problem}, along with a brief on the hyperparameter setup, like learning rate scheduler, minibatch size.
\subsection{Prediction using DeepONet for syngas problem}
\label{Subsection:Numerical:Syngas:DeepONet}
In this section, we will discuss the training and results of DeepONet for the syngas problem when trained using the one-step training method of DeepONet (discussed in \Cref{Subsection:DeepONet}). As discussed, we have normalized the output to $[-1,1]$. We observed that both the training and testing dataset lies between $-1$ and $1$ after normalization. Thus, in the case of one-step training, we restricted the output of DeepONet as,
\begin{equation}
    \widehat{\bm{Y}}\left(\bm{\Theta, \Phi}\right) = 1.05\times\tanh(\widehat{\bm{Y}}(\bm{\Theta, \Phi})).
    \label{Eq:Output restriction}
\end{equation}
We know that $\tanh(\:.\:)$ is bounded in $(-1,1)$. Thus, in order to account for the exclusion of the limit, i.e., -1 and 1, we consider the multiplication factor $1.05$. Furthermore, 1.05 also acts as a limit for how much extrapolation we expect in case we need to predict extrapolated data. We implemented the PoU in the DeepONet and incorporated loss due to conservation of mass in the loss function as discussed in \Cref{Subsection:POU and CoM}. We trained the DeepONet with the three different loss functions discussed earlier. First, the standard MSE loss function with additional loss due to CoM (ref. \Cref{Eq:Loss:Data_loss,Eq:Combined loss data+CoM,Eq:Loss conservation of mass}).In this case, we consider the weighting coefficient for $\mathcal{L}_\text{CoM}$ ($\mathcal{W}_{\text{CoM}}$) in \Cref{Eq:Combined loss data+CoM} as 0.1. The second loss function is the adaptive loss function ``Type-A" given by \Cref{Eq:Adaptive loss:Type-A} with $\mathcal{R}_1 = j = 12$. As discussed earlier, we initialized each weight as $\mathcal{W}_a = 1$. In this case, we consider the weight ($\mathcal{W}_{\text{CoM}}$) in \Cref{Eq:Combined loss data+CoM} as the $0.1$ times the sum of the weights of the data loss, i.e., $\mathcal{W}_\text{CoM} = 0.1\times \sum  \mathcal{W}_a$. The third loss function is the adaptive loss function ``Type-B" given by \Cref{Eq:Adaptive loss:Type-B} with $\mathcal{R}_2 = j\times bs = 12\times 205920$, with initial value of each weight as $\mathcal{W}_{ba}=1$. In this case, we consider the weight ($\mathcal{W}_{\text{CoM}}$) in \Cref{Eq:Combined loss data+CoM} as the $0.1$ times the sum of the weights of the data loss, i.e., $\mathcal{W}_\text{CoM} = 0.1\times \sum  \mathcal{W}_{ba}$. We also like to mention that as we consider mini-batch, the $\mathcal{W}_\text{CoM}$ changes with each mini-batch, even in the same epoch. This is because we have calculated $\mathcal{W}_\text{CoM}$ for a respective $\mathcal{W}_{ba}$ corresponding to a particular mini-batch. We update the self-adaptive weights, $\mathcal{W}_a$ and $\mathcal{W}_{ba}$ after $100$ epochs and at every $50$ epochs.
\par The mean of relative $L_2$ error (as discussed in \Cref{Subsection:Error matric} with $n_t+1=101$) for each of the three different runs and different methods is shown in \Cref{Table:Syngas:L2 error one step}. We observed that in the case of DON, the error reduces from approx $0.154\%$ to approx $0.064\%$ when considering adaptive weight in the loss function (i.e., adaptive loss function Type-B). In the case of DOK, the error reduces from approx $0.101\%$ to approx $0.043\%$. The difference between the mean of relative $L_2$ error for the three independent runs is very small, showing the robustness of the training and the minimum effect of initialization of the DeepONet parameters. 
\begin{table}[H]
    \centering
    \caption{\textbf{Training and testing error for Syngas problem.} Relative $L_2$ error in training and testing for DON (DeepONet) and DOK (DeepKAN) when trained using standard training (one-step) with different loss functions. The second and third rows indicate the neural network architecture considered in the trunk and branch networks, respectively. The network size and other details are shown in \Cref{Table:Syngas: network size} (in \Cref{Appendix:Subsection:Syngas problem}). In the fourth column, we have indicated the type of loss function considered for each case. The last two columns show the mean relative $L_2$ error (\%) for training and testing, respectively, for three independent training runs for each case. The convergence of the mean of relative $L_2$ error with epoch is shown in \Cref{Figure:Syngas:Convergence L2 error one step} (in \Cref{Appendix:Subsection:Syngas problem}).}
    \label{Table:Syngas:L2 error one step}
    \begin{tabular}{L{2cm}|C{1.25cm}|C{1.25cm}|C{2.65cm}|C{3cm}|C{3cm}} \hline
    \rowcolor{orange!45} \multicolumn{6}{c}{One-step training of DeepONet} \\ \hline
    \rowcolor{orange!45} & & & & \multicolumn{2}{c}{Relative $L_2$ error (\%) } \\ \cline{5-6}
    \rowcolor{orange!45} \multirow{-2}{*}{Case} & \multirow{-2}{*}{Trunk} & \multirow{-2}{*}{Branch} & \multirow{-2}{*}{Loss function} & Training & Testing \\  \hline
DON-NA & ResNet & ResNet & Non-Adaptive & 0.154, 0.152, 0.154 & 0.154, 0.151, 0.153 \\ \hline
\rowcolor{magenta!25} DON-Ad-A & ResNet & ResNet & Adaptive: Type-A & 0.077, 0.077, 0.076 & 0.076, 0.077, 0.076 \\ \hline
\rowcolor{cyan!25} DON-Ad-B & ResNet & ResNet & Adaptive: Type-B & 0.064, 0.062, 0.061 & 0.064, 0.062, 0.061 \\ \hline
DOK-NA & KAN & ResNet & Non-Adaptive & 0.099, 0.099, 0.101 & 0.099, 0.099, 0.101 \\ \hline
\rowcolor{magenta!25} DOK-Ad-A & KAN & ResNet & Adaptive: Type-A & 0.052, 0.051, 0.054 & 0.052, 0.050, 0.054 \\ \hline
\rowcolor{cyan!25} DOK-Ad-B & KAN & ResNet & Adaptive: Type-B & 0.042, 0.043, 0.042 & 0.042, 0.043, 0.043 \\ \hline
\end{tabular}
\end{table}
\subsection{Prediction using two-step training of DeepONet for syngas problem}
\label{Subsection:Numerical:Syngas:Two-step DeepONet}
\par We train our DeepONets with the two-step training method discussed in \Cref{Subsection:Two-step training}. We have considered the same normalization scheme discussed earlier; however, we have not considered the bound discussed in the one-step straining given by \Cref{Eq:Output restriction} as we need to separate the trunk and branch during training. There can be many combinations of network types for both branch and trunk networks, along with varied types of loss functions. Therefore, we first train the trunk network with the two proposed network architectures, i.e., ResNet and KAN, and the three different loss functions considered. After that, we train the branch with the best trunk network obtained from the first step. The size of the $\bm{A}$ tensor (in \Cref{Eq:2-Step trunk output einsum}) is $12\times 95\times (2080\times99)$. Similarly to the one-step training, in the trunk training of the two-step training, we observed that the error was reduced when we considered the adaptive loss function in both cases, ResNet and KAN. Furthermore, the error in the case of KAN is smaller than that in ResNet. The errors between different runs are small. The relative error in training of trunk in case of KAN network with adaptive weight Type-B (KAN-Ad-B) is $0.008\%$. The mean of relative $L_2$ error in training for the trunk training for the three different runs is shown in \Cref{Table:Syngas:2-step Trunk trining error} (in \Cref{Appendix:Subsection:Syngas problem}) and the convergence with epoch for the trunk training is shown in \Cref{Figure:Syngas:Covergence 2-step training L2 trunk training} (in \Cref{Appendix:Subsection:Syngas problem}).
\par After training the trunk network with different types of network architecture and different loss functions, we trained the branch using the reconstructed output labeled data ($\bm{U} = \bm{R}^*\bm{A}^*$). We consider three different loss functions, non-adaptive and two types of adaptive loss functions, as discussed in \Cref{Subsection:Two-step training}. The mean of relative $L_2$ error for three independent runs is shown in \Cref{Table:Syngas:2-step branch training testing error}. Similar to the previous observation, in this case as well, we observed that the error reduces as the adaptive loss function is considered. 
\begin{table}[H]
\centering
\caption{\textbf{Training and testing error for Syngas problem after branch training.} Relative $L_2$ error in training and testing for branch training in two-step training of DeepONet. The second and third columns show the type of network architecture considered for the branch and the loss function considered, respectively. The last two columns show the mean relative $L_2$ error (\%) in training and testing for three independent training runs. The convergence of the relative $L_2$ error with epoch is shown in  \Cref{Figure:Syngas:Convergence:2-step:L2 branch training} (in \Cref{Appendix:Subsection:Syngas problem}).}
\label{Table:Syngas:2-step branch training testing error}
\begin{tabular}{l|c|c|c|c} \hline
\rowcolor{orange!45} \multicolumn{5}{c}{Two-step training of DeepONet: After branch training.} \\ \hline
\rowcolor{orange!45} & & & \multicolumn{2}{c}{Relative $L_2$ error (\%) } \\ \cline{4-5} 
\rowcolor{orange!45} \multirow{-2}{*}{Case} & \multirow{-2}{*}{Branch} & \multirow{-2}{*}{Loss function} & Training & Testing \\  \hline
2S-NA & ResNet & Non-Adaptive & 0.086, 0.084, 0.084 &0.086, 0.084, 0.084 \\ \hline
\rowcolor{magenta!25} 2S-Ad-A & ResNet & Adaptive: Type-A & 0.044, 0.046, 0.045 & 0.044, 0.046, 0.045 \\ \hline
\rowcolor{cyan!25} 2S-Ad-B & ResNet & Adaptive: Type-B & 0.041, 0.040, 0.041 & 0.041, 0.040, 0.041 \\ \hline
\end{tabular}
\end{table}
\subsection{Results and discussion}
\label{Subsection:Numerical:Syngas:Result analysis}
\begin{figure}[H]
    \centering
    \includegraphics[width=1\textwidth]{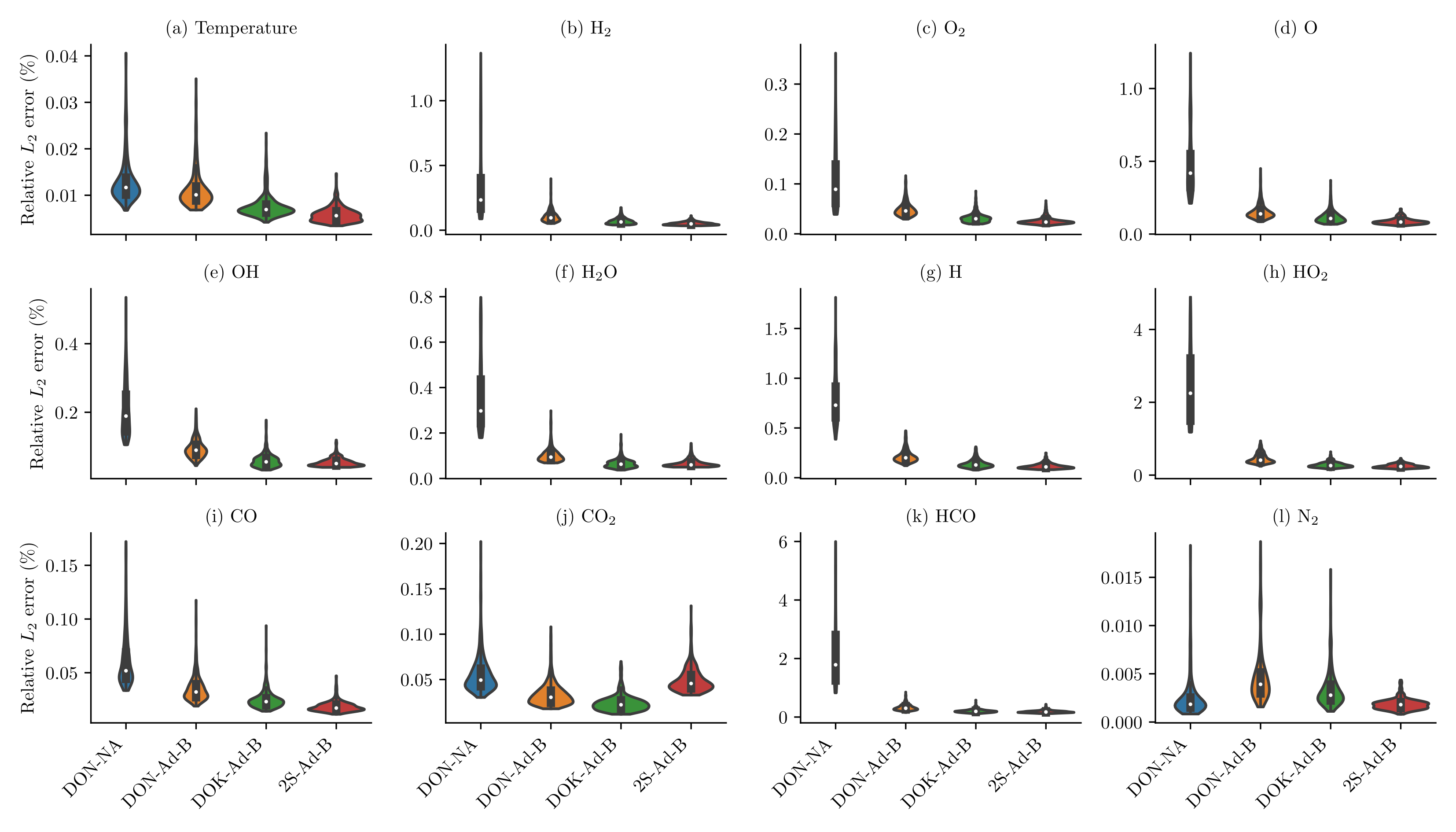}
    \caption{\textbf{Syngas problem, Violin plot} of the relative $L_2$ error of the reconstructed prediction of the state variables of the test dataset. Here, we have shown only four methods; the violin plots for all the methods considered are shown in \Cref{Figure:Syngas:Violin plot all 9 method} (in \Cref{Appendix:Subsection:Syngas problem}). We observed that the predicted results using DON-NA have the highest mean of the relative $L_2$ error compared to the other methods. Furthermore, DON-NA also has a higher standard deviation of the relative $L_2$ error. The mean and standard deviation are reduced as adaptive loss functions are considered. We would particularly like to mention that the state variables corresponding to the dynamics of HCO and HO$_2$ shown in (k) and (h), respectively, have higher errors, which are reduced when we consider the adaptive loss function. We have observed that the DON-Ad-B and 2S-Ad-B have similar accuracy; however, 2S-Ad-B has smaller standard deviations. The relative $L_2$ errors are calculated as discussed in \Cref{Subsection:Numerical:Syngas:Result analysis}.}
    \label{Figure:Syngas:Violin 4 method}
\end{figure}

\begin{figure}[H]
    \centering
    \begin{subfigure}[b]{1\textwidth}
    \includegraphics[width=1\textwidth]{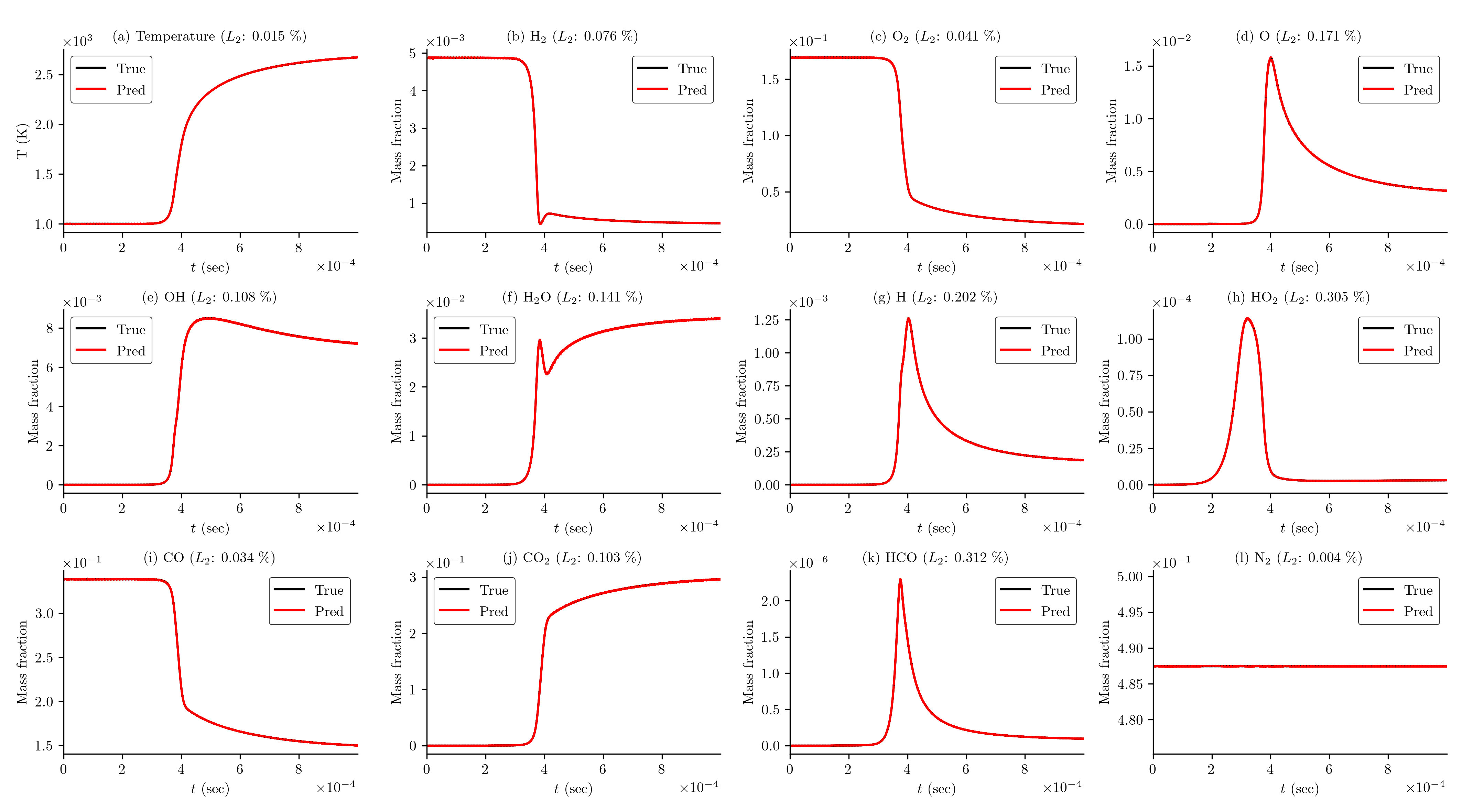}
    \caption{Test Sample 1 (2S-Ad-B)}
    \end{subfigure}
    \caption{\textbf{Syngas problem, sample test result, 2S-Ad-B:} Plot showing a representative sample result   from the test dataset when predicted using 2S-Ad-B, i.e., DeepOKAN trained using the two-step training method and adaptive loss function Type-B. The number in the bracket indicates the relative $L_2$ errors for each species (calculated as discussed in \Cref{Subsection:Numerical:Syngas:Result analysis}). The sample result corresponds to one of the higher errors in prediction. The predicted sample shows good accuracy with the true value. Two additional sample results are shown in \Cref{Figure:Syngas:2 step: Sample results appendix} (in Appendix \Cref{Appendix:Subsection:Syngas problem})}
    \label{Figure:Syngas:2 step: Sample 1}
\end{figure}

We trained two different types of DeepONet, the first one with ResNet as both the branch and trunk networks, the second one with ResNet as the branch and KAN as the trunk network. We trained these DeepONets with three different types of loss functions: non-adaptive and two adaptive loss functions. We also trained them with two different training paradigms, i.e., one-step training and two-step training. We observed that the two-step training with KAN as trunk and ResNet as branch gives the smallest mean of the relative $L_2$ error when trained using adaptive loss function Type-B. However, a single metric (the mean, in this case) does not accurately represent the methods. Thus, we study the results with the test dataset in more detail. For this, we will study the statistics of the error in more detail. Note that the test dataset consists of $520$ samples. As discussed earlier, we had divided these into $99$ segments of $100\Delta t$ length, i.e., $n_t+1 = 101$. We know the initial conditions of each of the $99$ segments, and the DeepONet predicts the $100\Delta t $ s advance when the initial conditions are given. In order to study the dynamics, we reconstructed the time series from the DeepONet prediction, i.e., we flattened the time series to a length equal to $99*101$. We would like to mention that there are predictions at common time points, as discussed earlier. We calculate the relative $L_2$ error using \Cref{Eq:L2 error} for each state variable and $520$ samples (assuming $n_t+1=99*101$). The mean and standard deviation of the relative $L_2$ error for each state variable and method are shown in \Cref{Table:Syngas:Mean and std test data}. The violin plots of the relative $L_2$ error for DON-NA, DON-Ad-B, DOK-Ad-B, and 2S-Ad-B for each of the state variables are shown in \Cref{Figure:Syngas:Violin 4 method} (violin plots for all the methods considered are shown in \Cref{Figure:Syngas:Violin plot all 9 method} (in \Cref{Appendix:Subsection:Syngas problem})). We observed that state variables describing the dynamics of HO$_2$ and HCO have the higher mean and standard deviation of errors compared to the rest of the state variables, as seen in \Cref{Table:Syngas:Mean and std test data}. As expected, the mean and standard deviation of error reduce as we consider the adaptive loss function. Though the overall error in the case of DOK-Ad-B and 2S-Ad-B is nearly the same, the 2S-Ad-B has a smaller standard deviation as observed in \Cref{Figure:Syngas:Violin 4 method}, particularly for O, H, HO$_2$, and HCO. However, the training time for two-step training is higher than the standard one-step training. The training time for different cases is discussed in \Cref{Section:Numeical:Computational cost}. A representative sample dynamics from the test dataset is shown in \Cref{Figure:Syngas:2 step: Sample 1} and two additional sample results are shown in \Cref{Figure:Syngas:2 step: Sample results appendix} (in \Cref{Appendix:Subsection:Syngas problem}) when predicted using 2S-Ad-B. We observed that the predicted dynamics of the state variables show good accuracy with the true dynamics. The point-wise error values over time for three representative samples from the test dataset when predicted 2S-Ad-B are shown in \Cref{Figure:Syngas:Pointwise error 2S-Ad-B}.
\begin{table}[H]
\caption{\textbf{Syngas problem, mean and standard deviation of the \% relative $L_2$ error} for the test samples. The state variables describing the dynamics of HO$_2$ and HCO have the maximum mean relative $L_2$ error. We observed that the errors in these two variables were reduced when the adaptive loss functions were considered. Furthermore, the standard deviation of the relation $L_2$ error is also reduced when adaptive loss functions are considered. The violin plots for the relative $L_2$ error are shown in \Cref{Figure:Syngas:Violin 4 method,Figure:Syngas:Violin plot all 9 method}. The relative $L_2$ errors are calculated as discussed in \Cref{Subsection:Numerical:Syngas:Result analysis}. }
\label{Table:Syngas:Mean and std test data}
\centering
\resizebox{\textwidth}{!}{%
\begin{tabular}{c|c|c|c|c|c|c|c|c|c|c} \hline
\rowcolor{orange!45} & & \multicolumn{3}{c|}{DeepONet} & \multicolumn{3}{c|}{DeepOKAN} & \multicolumn{3}{c}{Two-step training} \\ \hline
\rowcolor{orange!45} \rotatebox{90}{\begin{tabular}{c}State \\ variable\end{tabular}} &  & \rotatebox{50}{DON-NA} & \rotatebox{50}{DON-Ad-A} & \rotatebox{50}{DON-Ad-B} & \rotatebox{50}{DOK-NA} & \rotatebox{50}{DOK-Ad-A} & \rotatebox{50}{DOK-Ad-B} & \rotatebox{50}{2S-NA} & \rotatebox{50}{2S-Ad-A} & \rotatebox{50}{2S-Ad-B} \\ \hline
& $\mu$ & 0.013 & 0.014 &  0.011 &  0.009 &  0.010 &  0.008 &  0.006 &  0.007 &  0.006 \\ 
\multirow{-2}{*}{T} & $\sigma$ & 0.005 & 0.006 &  0.004 &  0.003 &  0.004 &  0.002 &  0.002 &  0.002 &  0.002 \\ \hline
\rowcolor{cyan!25} & $\mu$ & 0.340 & 0.196 &  0.099 &  0.176 &  0.121 &  0.069 &  0.099 &  0.081 &  0.052 \\ 
\rowcolor{cyan!25} \multirow{-2}{*}{H$_2$} & $\sigma$ & 0.267 & 0.118 &  0.037 &  0.103 &  0.069 &  0.024 &  0.034 &  0.032 &  0.013 \\ \hline
& $\mu$ & 0.108 & 0.097 &  0.045 &  0.072 &  0.060 &  0.031 &  0.044 &  0.038 &  0.026 \\ 
\multirow{-2}{*}{O$_2$} & $\sigma$ & 0.063 & 0.054 &  0.013 &  0.036 &  0.026 &  0.009 &  0.017 &  0.015 &  0.007 \\ \hline
\rowcolor{cyan!25} & $\mu$ & 0.482 & 0.222 &  0.151 &  0.319 &  0.180 &  0.115 &  0.203 &  0.114 &  0.088 \\ 
\rowcolor{cyan!25} \multirow{-2}{*}{O} & $\sigma$ & 0.222 & 0.098 &  0.047 &  0.143 &  0.080 &  0.040 &  0.062 &  0.031 &  0.021 \\ \hline
& $\mu$ & 0.207 & 0.098 &  0.090 &  0.139 &  0.069 &  0.058 &  0.107 &  0.051 &  0.054 \\ 
\multirow{-2}{*}{OH} & $\sigma$ & 0.077 & 0.039 &  0.022 &  0.047 &  0.026 &  0.019 &  0.034 &  0.017 &  0.013 \\ \hline
\rowcolor{cyan!25} & $\mu$ & 0.351 & 0.136 &  0.097 &  0.175 &  0.089 &  0.066 &  0.147 &  0.070 &  0.066 \\ 
\rowcolor{cyan!25} \multirow{-2}{*}{H$_2$O} & $\sigma$ & 0.150 & 0.051 &  0.031 &  0.062 &  0.035 &  0.021 &  0.047 &  0.025 &  0.017 \\ \hline
& $\mu$ & 0.785 & 0.350 &  0.201 &  0.487 &  0.255 &  0.139 &  0.293 &  0.161 &  0.116 \\ 
\multirow{-2}{*}{H} & $\sigma$ & 0.244 & 0.138 &  0.057 &  0.162 &  0.097 &  0.041 &  0.084 &  0.046 &  0.027 \\ \hline
\rowcolor{cyan!25} & $\mu$ & 2.443 & 0.972 &  0.432 &  1.322 &  0.631 &  0.280 &  0.880 &  0.451 &  0.250 \\ 
\rowcolor{cyan!25} \multirow{-2}{*}{{\color{red}HO$_2$}} & $\sigma$ & 1.012 & 0.450 &  0.117 &  0.578 &  0.264 &  0.073 &  0.254 &  0.151 &  0.056 \\ \hline
& $\mu$ & 0.061 & 0.055 &  0.034 &  0.044 &  0.039 &  0.025 &  0.028 &  0.025 &  0.018 \\ 
\multirow{-2}{*}{CO} & $\sigma$ & 0.023 & 0.019 &  0.011 &  0.018 &  0.017 &  0.008 &  0.009 &  0.007 &  0.005 \\ \hline
\rowcolor{cyan!25} & $\mu$ & 0.055 & 0.036 &  0.033 &  0.038 &  0.026 &  0.024 &  0.093 &  0.043 &  0.049 \\ 
\rowcolor{cyan!25} \multirow{-2}{*}{CO$_2$} & $\sigma$ & 0.021 & 0.018 &  0.011 &  0.015 &  0.012 &  0.009 &  0.032 &  0.015 &  0.013 \\ \hline
& $\mu$ & 2.197 & 0.654 &  0.305 &  0.902 &  0.413 &  0.205 &  0.499 &  0.261 &  0.182 \\ 
\multirow{-2}{*}{{\color{red}HCO}} & $\sigma$ & 1.250 & 0.355 &  0.085 &  0.413 &  0.189 &  0.060 &  0.138 &  0.071 &  0.044 \\ \hline
\rowcolor{cyan!25} & $\mu$ & 0.002 & 0.007 &  0.004 &  0.002 &  0.005 &  0.003 &  0.001 &  0.002 &  0.002 \\ 
\rowcolor{cyan!25} \multirow{-2}{*}{N$_2$} & $\sigma$ & 0.002 & 0.007 &  0.003 &  0.002 &  0.005 &  0.002 &  0.000 &  0.001 &  0.001 \\ \hline
\end{tabular}}
\end{table}

\subsection{Mass conserving DeepONet}
\label{Subsection:Numerical:Syngas:Mass conserving DeepONet}
In the previous sections, we have discussed the results of the proposed DeepONets with different loss functions and training strategies. We have considered a loss function that weakly incorporates CoM in the single-step training strategy; however, this does not ensure that the DeepONet outputs always follow the CoM constraint. In \Cref{Subsection:Mass conserving DeepONet}, we have proposed an operator that automatically satisfies the CoM constraint of the species. In this section, we will discuss the implementation and results of the mass-conserving DeepONet. In this case, we consider DeepOKAN (DOK) trained with the adaptive loss function Type-B and trained using two-step training. The number of species for the syngas problem is 11, which, after transformation using forward mapping (ref. \Cref{Eq:CoM DeepONet forward map:Append,Eq:CoM DeepONet forward map detail:Append}), reduces to 10. Thus, the number of reduced state variables is 11 (temperature and 10 mass fractions of the reduced species). We have considered a similar normalization scheme for the input and output data, as discussed earlier (on the 11 transformed state variables).
\par
\begin{figure}[H]
    \centering
    \includegraphics[width=1\textwidth]{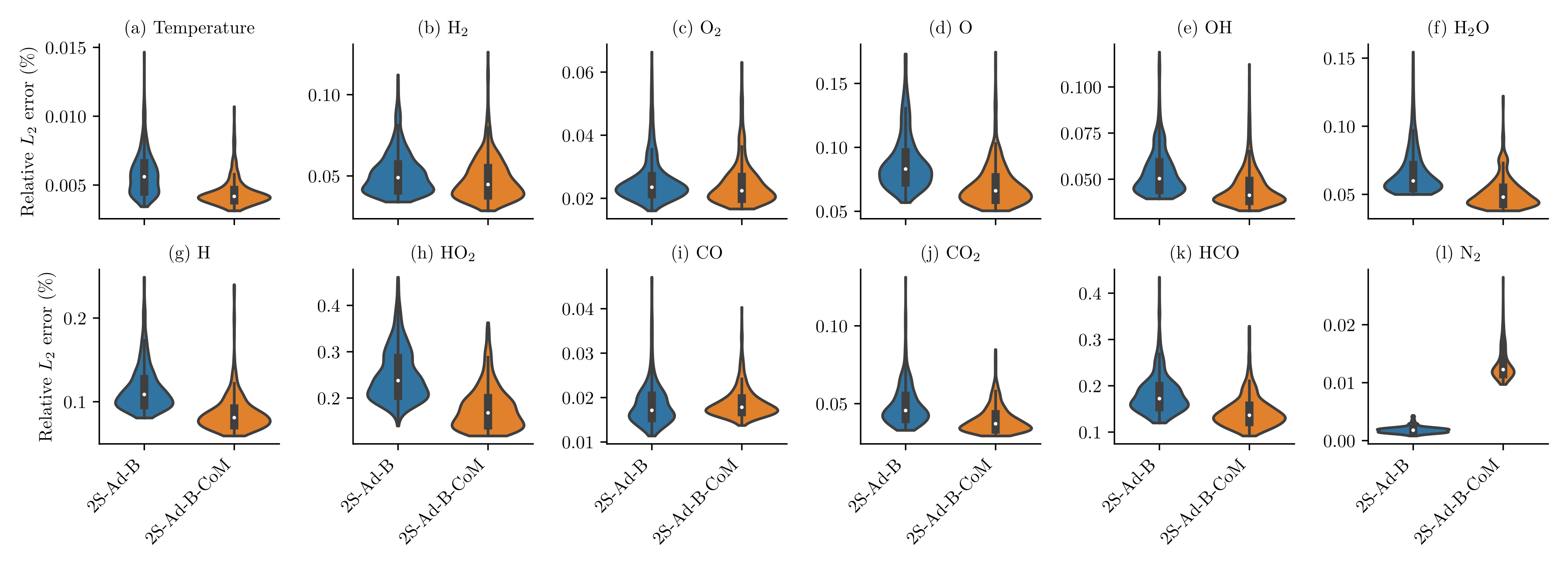}
    \caption{\textbf{Syngas problem, mass conserving DeepONet, Violin plot} for relative $L_2$ error of the reconstructed prediction of the state variables of the test dataset when predicted using the mass conserving DeepONet (2S-Ad-B-CoM) and compared with 2S-Ad-B discussed in \Cref{Subsection:Numerical:Syngas:Two-step DeepONet}. The relative $L_2$ errors are calculated similarly to those discussed in \Cref{Subsection:Numerical:Syngas:Result analysis}.}
    \label{Figure:Syngas:Mass conserving DeepONet:Violin plot}
\end{figure}

 We have considered the trunk with KAN architecture with the same network size as considered in \Cref{Table:Syngas: network size} (in \Cref{Appendix:Subsection:Syngas problem}), except for the output layer. The number of neurons in the output layer is $11\times95=1045$ instead of $1140$. Similarly, we consider the branch with ResNet architecture with the same network size as considered in \Cref{Table:Syngas: network size}, except the input layer is $11$ instead of $12$ and the output layer is $11\times95=1045$ instead of $1140$. The size of the $\bm{A}$ tensor (in \Cref{Eq:2-Step trunk output einsum}) is $11\times95\times(2080\times99)$. We named this DeepONet as 2S-Ad-B-CoM. We train the parameters of the trunk ($\bm{\Phi}$) and $\bm{A}$ up to 15,000 epochs. The mean of the relative $L_2$ error in training for the trunk training is $0.0080\%$, $0.0083\%$, and $0.0081\%$ in the transformed coordinate system, respectively (not normalized), for three independent runs. After the trunk training, we train the branch using the reconstructed labeled data, similar to established procedures discussed earlier. The mean of the relative $L_2$ error in training is $0.0344\%$, $0.0362\%$, $0.0344\%$, and testing is $0.0344\%$, $0.0362\%$, $0.0344\%$ after the branch training for three independent runs in the transformed coordinate system. The predicted output after training is converted back to the physical system using inverse mapping discussed in \Cref{Subsection:Mass conserving DeepONet,Appendix:Subsection:Mass conserving DeepONet}. The error in the original (physical) system in training is $0.0354$, $0.0370$, $0.0342$, and in testing is $0.0355$, $0.0370$, $0.0342$, respectively. For comparison, we have shown the training and testing error in \Cref{Table:Syngas:Mass conserve DeepONet:Relative L2} (in \Cref{Appendix:Subsection:Syngas problem}). The violin plots for the reconstructed results of the test dataset for the state variable (original) are shown in \Cref{Figure:Syngas:Mass conserving DeepONet:Violin plot}. We also study the mean and standard deviation of each state variable, similar to the discussion in \Cref{Subsection:Numerical:Syngas:Result analysis}. The mean and standard deviation of each state variable are shown in \Cref{Table:Syngas:Mass conserve DeepONet:Mean and Standard deviation} (in \Cref{Appendix:Subsection:Syngas problem}) along with results of 2S-Ad-B. We observed that the results of 2S-Ad-B-CoM are marginally better than the results of 2S-Ad-B. We would also like to mention that while in the mass-conserving DeepONet, the sum of the species mass fraction is one, the 2S-Ad-B does not follow CoM constraint. The minimum and maximum values of CoM in the case of 2S-Ad-B in the test dataset are $0.99852$ and $1.00162$, respectively. Furthermore, we want to mention that additional computation is required in the case of mass-conserving DeepONet for the inverse mapping. The convergence of relative $L_2$ error with epoch is shown in \Cref{Figure:Syngas:Convergence mass conserve DOK} (in \Cref{Appendix:Subsection:Syngas problem}).

\subsection{Mass conservation using softmax function}
\label{subsection:Numerical:Syngas:Mass conservation using softmax function}
In the previous sections, we have discussed two ways to enforce CoM constraint, either as a soft constraint in the form of an additional loss term in the single step training or enforcing it by construction such that it is automatically satisfied. Alternatively, CoM can also be enforced by applying the softmax function to the species' mass fraction output, which inherently ensures that the mass fractions always sum to one and are positive. In this case, we need to modify the DeepONet output as well as the loss function: the temperature is restricted to be positive using an $\exp$ function, and the species mass is conserved using softmax.
\begin{equation}
Y_{bca} = \begin{cases}
    \exp(Y_{bca}), \;\;\; a=0\\
    \text{softmax}(Y_{bca},\;\; \text{axis}=a), \;\;\; \forall a \neq 0 
\end{cases}
\end{equation}

We would like to mention that, in this case, while the inputs are normalized as discussed earlier, the DeepONet predicts the output
directly in physical space, i.e., no normalization of the output is considered. 
We also need to marginally change the loss function for accuracy and convergence. Instead of minimizing the data loss, we need to minimize the natural logarithm ($\ln$) of the data. For a non-adaptive loss function,
\begin{equation}
    \mathcal{L}_{\text{data}}(\bm{\Theta}, \bm{\Phi})
    = \dfrac{1}{j\times bs\times (n_t+1)}\sum_{a=0}^{j-1}\sum_{b=1}^{bs}\sum_{c=0}^{n_t}\left(\ln{Y}_{bca} - \ln{\widehat{Y}}_{bca}\right)^2, \;\;\;\;\;\; 
    \begin{array}{l}
    j\rightarrow \text{No. of states}, \\
    n_t+1\rightarrow \text{No. of time points} \\
    bs\rightarrow \text{No. of samples} \\
    \end{array}
\label{Eq:Loss:Data_loss_ln}
\end{equation}
Similarly, the adaptive loss function Type-B is
\begin{equation}
    \mathcal{L}_{\text{data}}(\bm{\Theta}, \bm{\Phi}) = \sum_{a=0}^{j-1}\sum_{b=1}^{bs}\mathcal{W}_{ba}\left[\dfrac{1}{ n_t+1}\sum_{c=0}^{n_t}\left(\ln{Y}_{bca} - \ln{\widehat{Y}}_{bca}\right)^2\right]. \;\;\;\;\;\; \begin{array}{l}
    j\rightarrow \text{No. of states variable}, \\
    n_t+1\rightarrow \text{No. of time points} \\
    bs\rightarrow \text{No. of samples} \\
    \end{array}
\label{Eq:Adaptive loss:Type-B_ln}
\end{equation}
\par We trained the DeepONet (DOK) (single step) with the same network size and hyperparameters considered earlier and the rsults are shown in \Cref{Table:Syngas:L2 error one step_softmax}. In this case as well, we observe an improvement in accuracy when considered adaptive loss function. 

\par
\begin{figure}[H]
    \centering
    \includegraphics[width=1\textwidth]{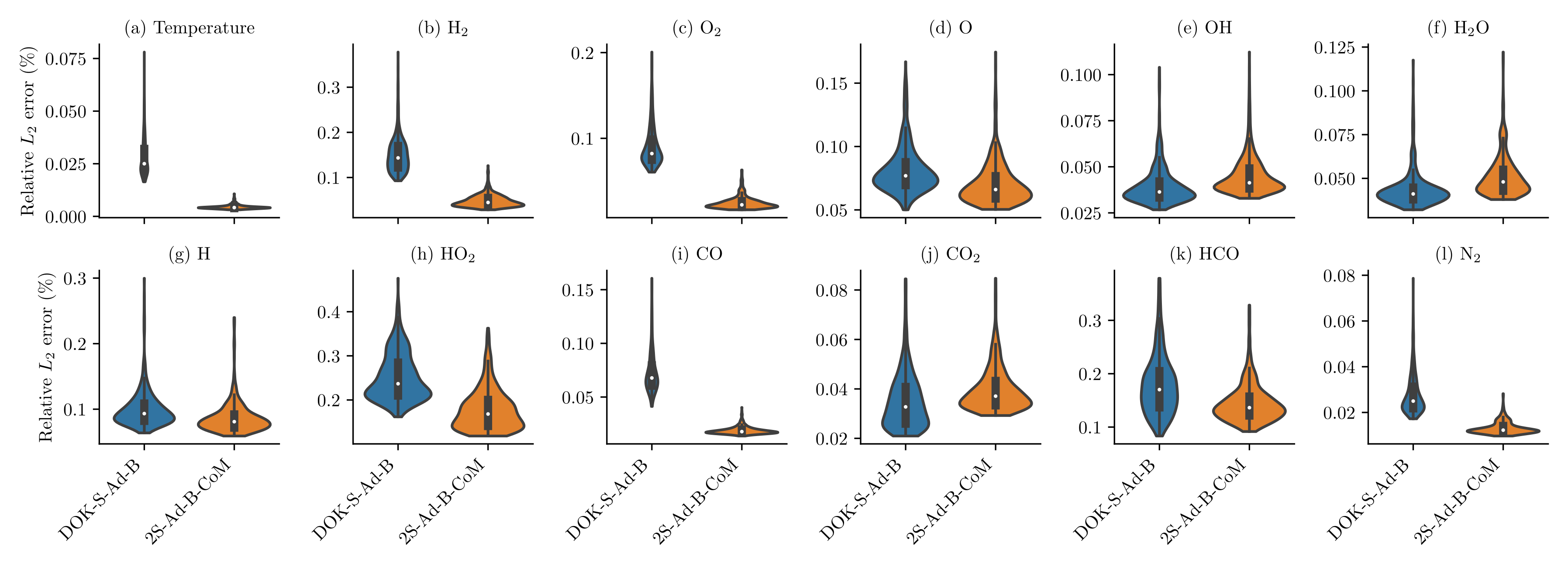}
    \caption{\textbf{Syngas problem, softmax mass conserving DeepOKAN, Violin plot} for relative $L_2$ error of the reconstructed prediction of the state variables of the test dataset when predicted using the softmax mass conserving DeepOKAN with adaptive weights Type-B (DOK-S-Ad-B) compared with 2S-Ad-B-CoM. The relative $L_2$ errors are calculated similarly to those discussed in \Cref{Subsection:Numerical:Syngas:Result analysis}.}
    \label{Figure:Syngas:Softamx_DOK_CoM_Violin}
\end{figure}

 We observe almost similar accuracy for the adaptive loss case between as DOK-Ad-B (DeepOKAN with single step training with adaptive loss function Type-B), 2S-Ad-B (DeepOKAN two step training with adaptive loss Type-B) for DOK-S-Ad-B (DeepOKAN single step training with adaptive loss function Type-B with softmax). However, for non adaptive loss function, the CoM with softmax shows better accuracy than soft constraint. Furthermore, the mass conservation using the coordinate transform (2S-Ad-B-CoM) (\Cref{Subsection:Mass conserving DeepONet,Subsection:Numerical:Syngas:Mass conserving DeepONet}) shows better accuracy than softmax mass conservation. We would also like to mention that the mass conservation using softmax is not possible in the two-step training, as we need to separate the trunk and branch networks during training. In \Cref{Figure:Syngas:Softamx_DOK_CoM_Violin}, we have shown the violin plot for the relative $L_2$ error of the reconstructed error output for the test dataset with a comparison with 2S-Ad-B-CoM.
\begin{table}[H]
    \centering
    \caption{\textbf{Training and testing error for Syngas problem, softmax mass conservation.} Relative $L_2$ error in training and testing for DOK (DeepKAN) with softmax CoM constraint and trained using standard training (one-step) with different loss functions. The second and third rows indicate the neural network architecture considered in the trunk and branch networks, respectively. The network size and other details are shown in \Cref{Table:Syngas: network size} (in \Cref{Appendix:Subsection:Syngas problem}). In the fourth column, we have indicated the type of loss function considered for each case. The last two columns show the mean relative $L_2$ error (\%) for training and testing, respectively, for three independent training runs for each case. The convergence of the mean of relative $L_2$ error with epoch is shown in \Cref{Figure:Syngas:Softmax_mass_Conseving_Convergence} (in \Cref{Appendix:Subsection:Syngas problem}).}
    \label{Table:Syngas:L2 error one step_softmax}
    \begin{tabular}{L{2cm}|C{0.9cm}|C{1.1cm}|C{2.65cm}|C{3.25cm}|C{3.25cm}} \hline
    \rowcolor{orange!45} \multicolumn{6}{c}{One-step training of DeepONet (softmax mass conservation)} \\ \hline
    \rowcolor{orange!45} & & & & \multicolumn{2}{c}{Relative $L_2$ error (\%) } \\ \cline{5-6}
    \rowcolor{orange!45} \multirow{-2}{*}{Case} & \multirow{-2}{*}{Trunk} & \multirow{-2}{*}{Branch} & \multirow{-2}{*}{Loss function} & Training & Testing \\  \hline
    DOK-S-NA & KAN & ResNet & Non-Adaptive & 0.0468, 0.0466,  0.046 & 0.0464, 0.0462, 0.046 \\ \hline
    DOK-S-Ad-B & KAN & ResNet & Adaptive: Type-B & 0.0431, 0.0432, 0.0427 & 0.0431, 0.0431, 0.0426 \\ \hline
\end{tabular}
\end{table}

\subsection{Test for extrapolation: Extrapolation using trained DeepONet}
\label{Subsection:Numerical:Syngas:Test for extrapolation}
In the previous sections, we have discussed the results of the syngas problem predicted using DeepONet, where the test dataset is within the training distribution. In this section, we will test our proposed model on extrapolated data. As mentioned earlier, the training-testing dataset is generated with an initial temperature range of 1000 K to 1500 K and an equivalence ratio range of [0.7, 1.3]. We generated two new datasets with (i) an initial temperature of 1800 K and an equivalence ratio of 0.9, (ii) an initial temperature of 1800 K and an equivalence ratio of 1.5. We predict the results using the trained model DOK-Ad-B and 2S-Ad-B in the previous sections, assuming the initial conditions are known for each time segment. The first set of results is shown in \Cref{Figure:Syngas:Extrapolation sample 1} and the second set of results in \Cref{Figure:Syngas:Extrapolation sample 2} (in \Cref{Appendix:Subsection:Syngas problem}). In the case of the first dataset, we also tested our model beyond the training time zone. The training time zero is 0 to $\sim1\times 10^{-3}$ s. We tested up to $\sim2\times 10^{-3}$ s. The results show good accuracy. In the second dataset, we observed that the prediction using DOK-Ad-B is marginally poor. Particularly, the predicted maximum output is smaller than the true values. We would like to mention that in the case of DOK-Ad-B, we have restricted our output using $1.05\times\tanh(\:.\:)$ function (ref. \Cref{Eq:Output restriction}). This restricts the maximum allowable extrapolation. We can modify the $1.05$ and allow maximum allowable extrapolation. Furthermore, we also observed in both cases that predictions using 2S-Ad-B are noisier than the predictions using DOK-Ad-B. A rigorous theoretical or mathematical analysis of why the predictions using 2S-Ad-B in these cases are noisier is beyond the scope of the present study. However, we believe that the $\tanh(\:.\:)$ (ref. \Cref{Eq:Output restriction}) function in DOK-Ad-B in the output of DeepONet makes the output smoother or less noisy. Furthermore, in the case of DOK-Ad-B, we also consider the PoU in the trunk output.
\begin{figure}[H]
    \centering
    \begin{subfigure}[b]{1\textwidth}
    \includegraphics[width=1\textwidth]{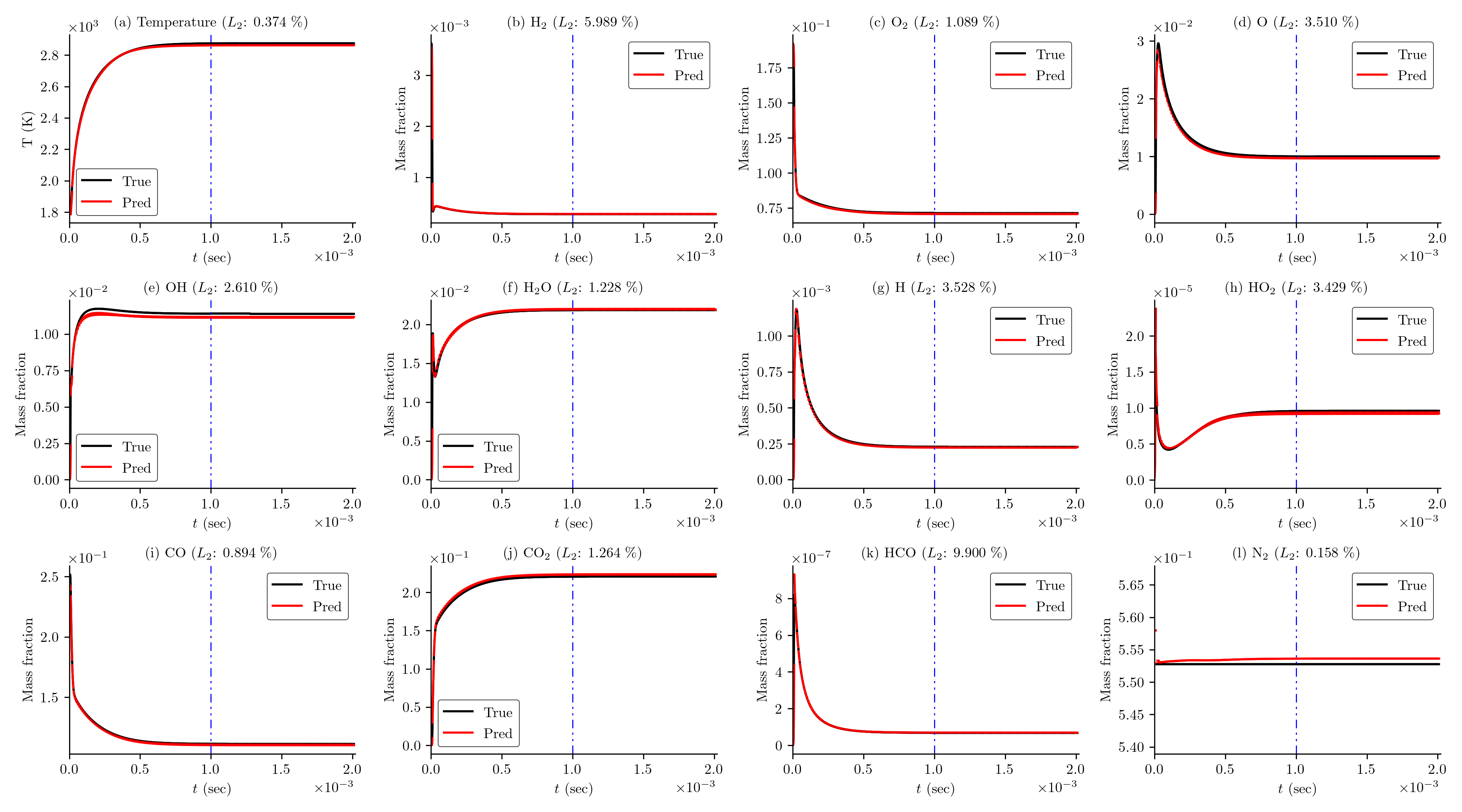}
    \caption{DOK-Ad-B, Extrapolation sample \# 1}
    \end{subfigure}
\end{figure}
\begin{figure}[H]\ContinuedFloat
    \begin{subfigure}[b]{1\textwidth}
    \includegraphics[width=1\textwidth]{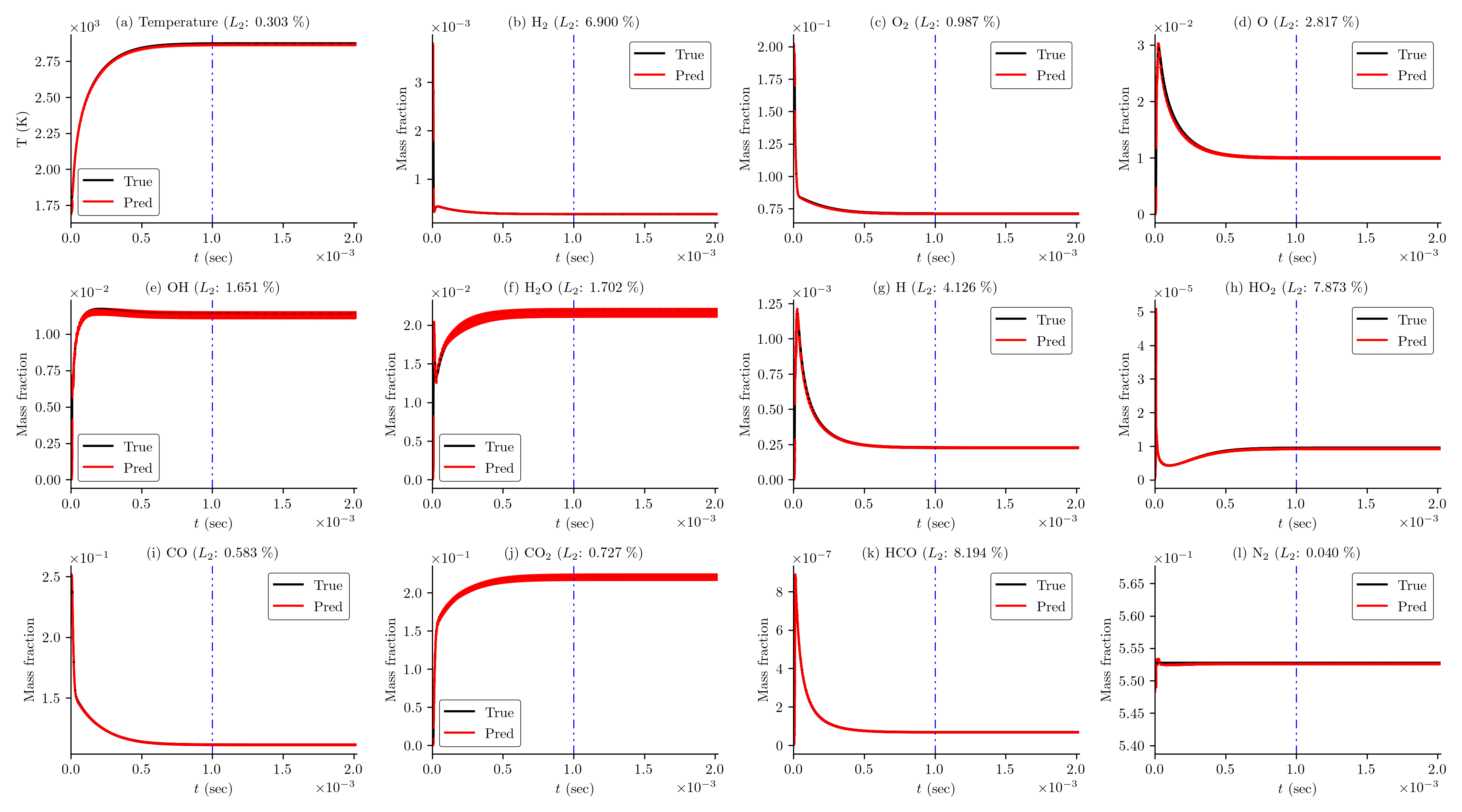}
    \caption{2S-Ad-B, Extrapolation sample \# 1}
    \end{subfigure}
    \caption{\textbf{Syngas problem, DeepONet extrapolation result \# 1.} Plot showing the dynamics of the state variables of the syngas problem for extrapolation dataset \#1 when predicted using trained (a) DOK-Ad-B and (b) 2S-Ad-B method. The predicted dynamics show good accuracy with the true dynamics.}
    \label{Figure:Syngas:Extrapolation sample 1}
\end{figure}
\subsection{Recursive prediction using DeepONet}
\label{Subsubsection:Numerical:Syngas:Recursive prediction}
\begin{figure}[H]
    \centering
    \begin{subfigure}[b]{1\textwidth}
    \includegraphics[width=1\textwidth]{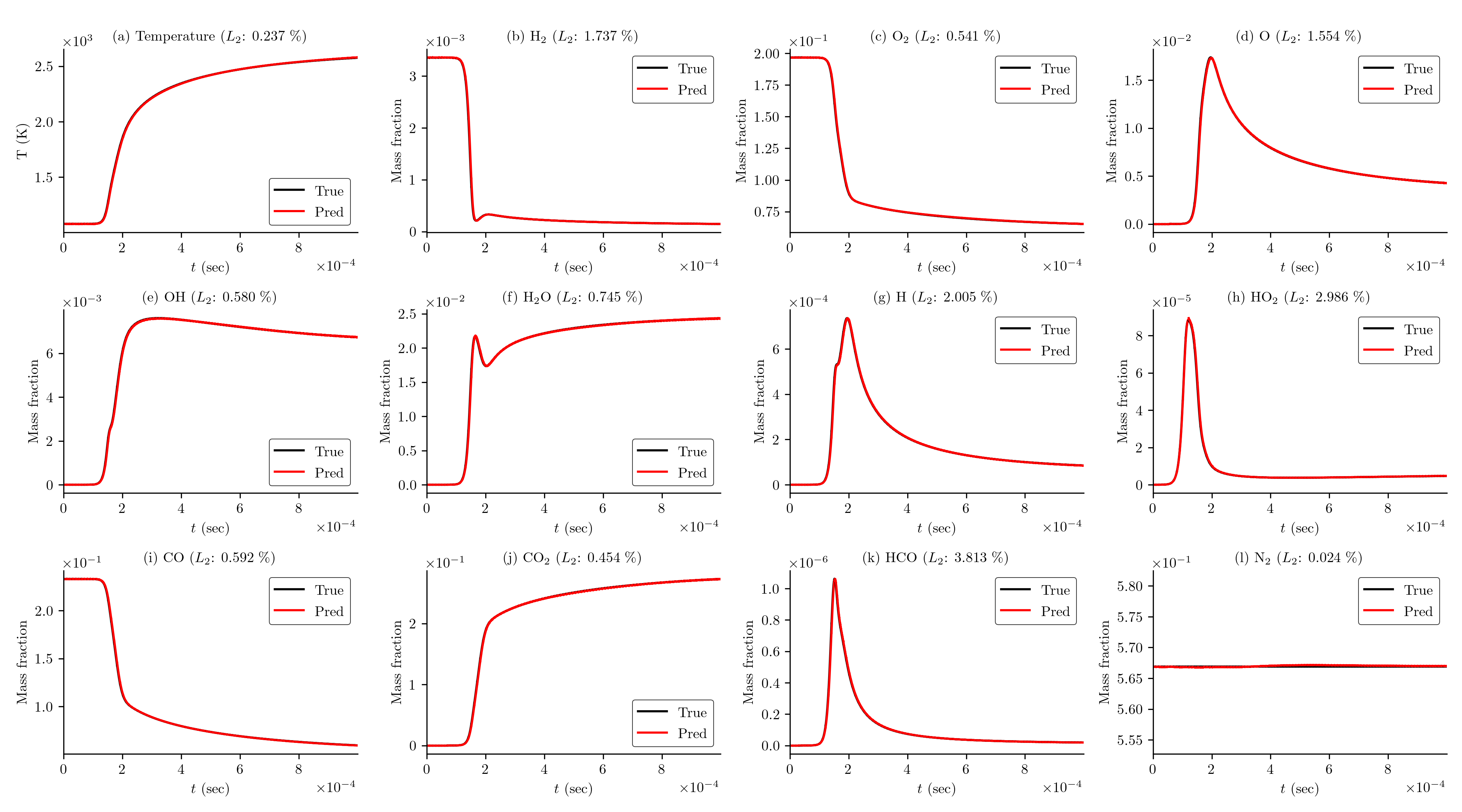}
    \caption*{Recursive prediction, Sample result 1. 90 percentile error in HCO.}    
    \end{subfigure}
    \caption{\textbf{Syngas problem, Recursive prediction, sample result.} Plot showing the dynamics of the state variables of the syngas problem for test dataset when predicted using recursive prediction. Plot shows the dynamics of the state variables corresponding to 90 percentile error in predicted state variable HCO (3.813\%), The plots show the dynamics of the predicted state variable when predicted recursively using the trained model 2S-Ad-B, which we called 2S-Ad-B-R. The plots show good accuracy in prediction. Additional results are shown in \Cref{Figure:Syngas:Recursive prediction:Sample results: Appendix} (in \Cref{Appendix:Subsection:Syngas problem}).}
    \label{Figure:Syngas:Recursive prediction:Sample results}
\end{figure}

\begin{figure}[H]
    \centering
    \includegraphics[width=1\textwidth]{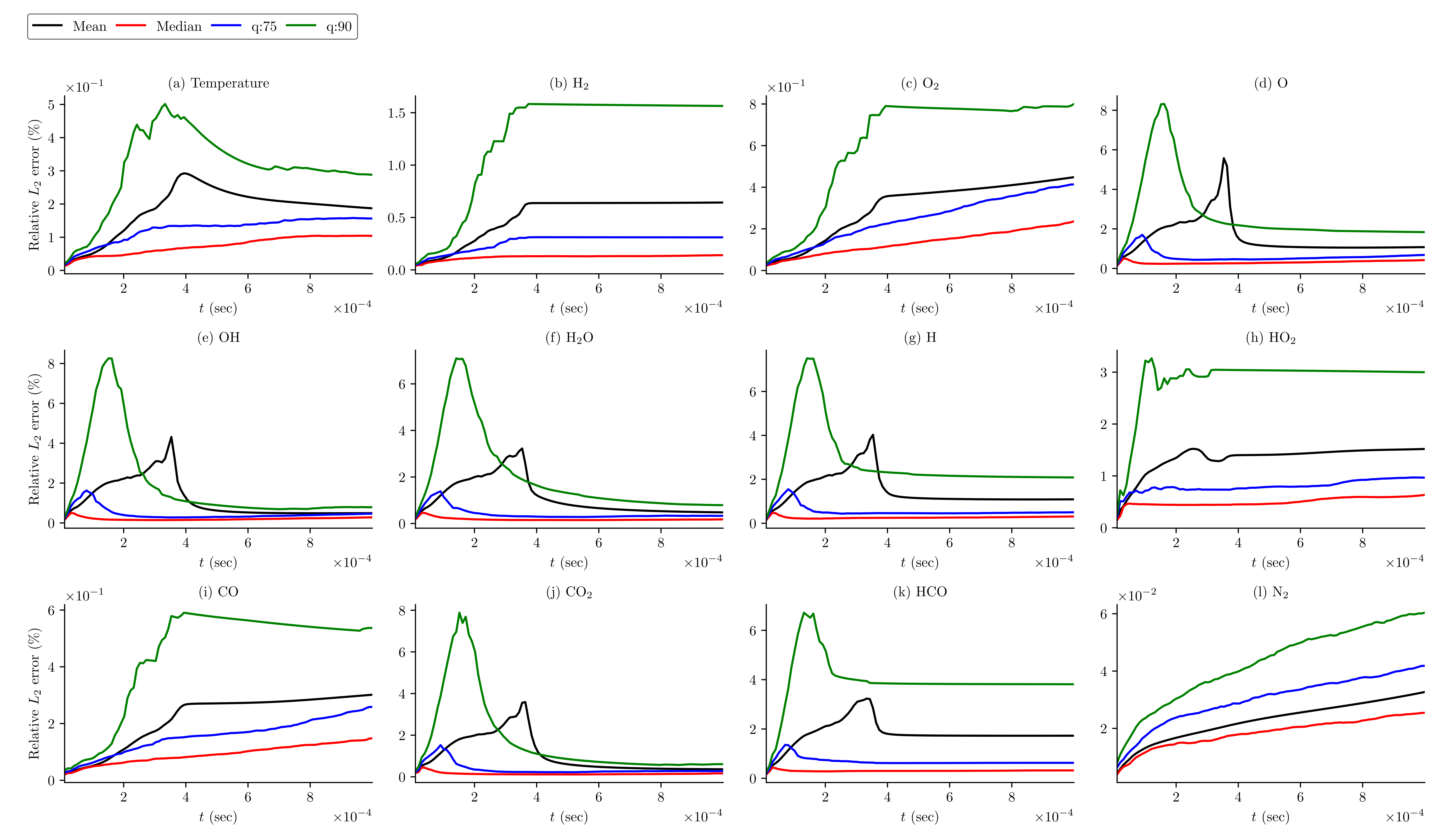}
    \caption{\textbf{Syngas problem, recursive prediction, error accumulation:} Plots show the accumulation of error over time for each state variable when predicted recursively in an autoregressive manner as discussed in \Cref{Subsubsection:Numerical:Syngas:Recursive prediction}. The relative $L_2$ errors are calculated similarly to the discussion in \Cref{Subsection:Numerical:Syngas:Result analysis} using the complete output up to that autoregressive step. Since the error is skewed, we have shown the median, $75$ percentile (q:75), and $90$ percentile (q:90) error growth along with mean error accumulation for the test dataset. We observed that the mean error is higher than the median and $75$ percentile error, indicating a high skewness. From the $90$ percentile error, we observed that maximum errors occur in the time zone $1\times 10^{-4}$ s to $4\times 10^{-4}$ s. The violin plots of the relative error at the end of time are shown in \Cref{Figure:Syngas:DeepONet:Recursive prediction:Violin plot}.}
    \label{Figure:Syngas:Recursive prediction:Error accumulations}
\end{figure}

In the previous sections, we have discussed the training process and the results of DeepONets when the initial conditions for each of the time segments are known. As discussed in \Cref{Subsection:Problem statement}, we will test our DeepONet models for long-time prediction using the recursive/auto-regressive prediction method discussed in \Cref{Subsection:Recursive prediction}. For this purpose, we consider the DeepONet models trained in the previous section to predict the output auto-regressively by considering the last output of the DeepONet as the input to predict the dynamics of the second segment, and so on. We consider the DeepONet model 2S-Ad-B trained in the previous section to predict the auto-regressive dynamics and call it 2S-Ad-B-R. In \Cref{Subsection:Numerical:Syngas:Mass conserving DeepONet}, we also discussed the mass-conserving DeepONet results. We also consider this model for recursive prediction and call it 2S-Ad-B-CoM-R. The mean of the test errors for the two methods for the three models (parameters of three independent runs) are shown in \Cref{Table:Syngas:Recursive:Test error} (in \Cref{Appendix:Subsection:Syngas problem}), when calculated using \Cref{Eq:L2 error} with $n_t+1=101$. We consider the complete test dataset, i.e., 520 test samples discussed earlier. Similar to \Cref{Subsection:Numerical:Syngas:Result analysis}, we study the statistics of the relative $L_2$ error of the reconstructed dynamics. The mean, standard deviation, median, $75$ percentile (q:75), $90$ percentile (q:90), and maximum error of the relative $L_2$ error of the reconstructed test samples for 2S-Ad-B-R and 2S-Ad-B-CoM-R are shown in \Cref{Table:Syngas:DeepONet:Recursive prediction error}. The violin plots for $L_2$ error for the reconstructed output dynamics are shown in \Cref{Figure:Syngas:DeepONet:Recursive prediction:Violin plot} (in \Cref{Appendix:Subsection:Syngas problem}). We observed that while most of the dynamics show good accuracy, there are high errors in a few of the samples, as long tails are observed in the violin plots. Dynamics of a test samples corresponding to 90 percentile error in HCO is \Cref{Figure:Syngas:Recursive prediction:Sample results} and additional dynamics corresponds to test sample are shown in \Cref{Figure:Syngas:Recursive prediction:Sample results: Appendix} (in \Cref{Appendix:Subsection:Syngas problem}). The dynamics show good accuracy. The accumulation of error over time for each state variable is shown in \Cref{Figure:Syngas:Recursive prediction:Error accumulations}.
\begin{table}[H]
\centering
\caption{\textbf{Syngas problem, Recursive prediction}, mean, standard deviation, median,  $75$ percentile (q:75), $90$ percentile (q:90), and maximum error of the relative $L_2$ error (\%) of the reconstructed test samples for 2S-Ad-B-R and 2S-Ad-B-CoM-R. The state variables describing the dynamics of HO$_2$ and HCO have the maximum mean relative $L_2$ error. We observed that 90 percentile of all the state variables have an error smaller than 5\% relative $L_2$ error. The relative $L_2$ errors are calculated similarly to the discussion in \Cref{Subsection:Numerical:Syngas:Result analysis}. The violin plots of relative $L_2$ error for each state variable are shown in \Cref{Figure:Syngas:DeepONet:Recursive prediction:Violin plot} (in \Cref{Appendix:Subsection:Syngas problem})}
\label{Table:Syngas:DeepONet:Recursive prediction error}
\renewcommand{\arraystretch}{1.25}
\begin{tabular}{C{0.95cm}?c|c|c|c|c|c?c|c|c|c|c|c}
\hline
\rowcolor{orange!45} & \multicolumn{6}{c?}{2S-Ad-B-R} & \multicolumn{6}{c}{2S-Ad-B-CoM-R} \\ \cline{2-13}
\rowcolor{orange!45}\multirow{-2}{*}{States } & $\mu$ & $\sigma$ & Median & q:75 & q:90 & Max& $\mu$ & $\sigma$ & Median & q:75 & q:90 & Max \\ \hline
T & 0.19 & 0.36 & 0.10 & 0.16 & 0.29 & 3.48 & 0.17 & 0.37 & 0.07 & 0.12 & 0.32 & 4.54 \\ \hline
\rowcolor{cyan!25} H$_2$ & 0.64 & 1.59 & 0.14 & 0.31 & 1.56 & 13.41 & 0.72 & 1.59 & 0.20 & 0.39 & 2.06 & 15.67 \\ \hline
O$_2$ & 0.45 & 0.79 & 0.24 & 0.41 & 0.80 & 7.44 & 0.52 & 0.83 & 0.30 & 0.50 & 0.97 & 11.17 \\ \hline
\rowcolor{cyan!25} O & 1.08 & 2.62 & 0.42 & 0.68 & 1.84 & 24.56 & 1.13 & 2.75 & 0.43 & 0.67 & 2.26 & 35.28 \\ \hline
OH & 0.49 & 0.88 & 0.28 & 0.47 & 0.79 & 8.44 & 0.48 & 0.90 & 0.27 & 0.41 & 0.85 & 11.33 \\ \hline
\rowcolor{cyan!25} H$_2$O & 0.48 & 1.16 & 0.18 & 0.33 & 0.79 & 10.60 & 0.52 & 1.17 & 0.20 & 0.37 & 0.98 & 12.66 \\ \hline
H & 1.09 & 2.94 & 0.31 & 0.50 & 2.08 & 27.02 & 1.13 & 3.04 & 0.30 & 0.49 & 2.77 & 35.83 \\ \hline
\rowcolor{cyan!25} {\color{red}HO$_2$} & 1.52 & 3.19 & 0.64 & 0.97 & 3.00 & 28.60 & 1.45 & 3.26 & 0.49 & 0.74 & 3.70 & 37.62 \\ \hline
CO & 0.30 & 0.58 & 0.15 & 0.26 & 0.54 & 5.22 & 0.38 & 0.58 & 0.20 & 0.38 & 0.76 & 5.52 \\ \hline
\rowcolor{cyan!25} CO$_2$ & 0.36 & 0.82 & 0.17 & 0.28 & 0.61 & 7.78 & 0.38 & 0.83 & 0.16 & 0.28 & 0.69 & 10.39 \\ \hline
{\color{red}HCO} & 1.73 & 4.95 & 0.32 & 0.63 & 3.81 & 43.91 & 1.84 & 5.02 & 0.38 & 0.70 & 4.66 & 52.83 \\ \hline
\rowcolor{cyan!25} N$_2$ & 0.03 & 0.02 & 0.03 & 0.04 & 0.06 & 0.14 & 0.06 & 0.04 & 0.04 & 0.06 & 0.11 & 0.30 \\ \hline
\end{tabular}
\end{table}


\section{Computational experiment: GRI - Mech 3.0 problem}
\label{Section:Numerical results:GRI Problem}

In the previous section, we showed that the proposed DeepONet-based framework AMORE can predict the dynamics of the stiff chemical system corresponding to the Syngas mechanism with good accuracy. In this regard, we shall show the scalability of the proposed framework to a larger number of state variables. To this end, we consider the GRI-Mech 3.0 as our second computational example. The GRI-Mech 3.0 is a detailed chemical reaction mechanism for natural gas combustion. GRI-Mech 3.0 is specifically designed to simulate the combustion of methane (CH$_4$) and its mixtures with air. The GRI-Mech 3.0 mechanism includes 53 species and 325 reactions, providing a detailed description of combustion processes for methane-based fuels. Similar to the Syngas problem discussed earlier (\Cref{Section:Numerical results:Syngas Problem}), we generate data for training and testing of DeepONet with the initial condition of temperature within the range of 1000 K to 1500 K and the equivalence ratio between 0.7 to 1.3. We also used Cantera with the VODE implicit time integrator (BDF), analogous to the Syngas problem with the relative tolerance of $10^{-8}$ and absolute tolerance of $10^{-12}$. The reactor is modeled as a constant-pressure system at 1.0 atm for all generated samples. The system has 54 state variables, i.e., temperature, along with the 53 state variables describing the dynamics of the mass fraction of each of the 53 species. From the simulated dataset, we have observed that out of 54 state variables, only 24 have significant values or non-negative and non-zero values. Thus, we have considered only 24 state variables. The minimum and maximum values of each state variable for the training and testing datasets are shown in \Cref{Table:GRI problem:Min max training-testing data} (in \Cref{Appendix:Subsection:GRI problems}). Similar to the syngas problem, we consider 2080 samples for training and 520 samples for testing. Similar to the syngas problem, we consider the time series from $t_0=10^{-7}$ s, which after the shifting of the time axis becomes $t=0$ s. In this computational example, as well, we have divided the time series into 99 segments, each of length $100\Delta t$, s where $\Delta t = 10^{-7}$. Thus, after performing the time decomposition, the number of training samples becomes $2080\times 99 = 205920$, and the number of testing samples becomes $520\times 99 = 51480$. For the sake of convenience, from now onward, we will refer to the GRI-Mech 3.0 problem as the GRI problem.
\subsection{Prediction using DeepONet}
\label{SubsectionL:Numerical:GRI:Prediction using DeepONet}
\par In the previous problem (\Cref{Section:Numerical results:Syngas Problem} for Syngas problem), we have considered different types of DeepONets with different loss functions and training strategies. We observe that DeepOKAN trained using the two-step training method with adaptive loss function Type-B shows better accuracy compared to the other methods. Therefore, for the GRI problem, we only consider DeepOKAN trained using the two-step training method with adaptive loss function Type-B. The network size considered for the GRI problem is shown in \Cref{Table:GRI: network size} (in \Cref{Appendix:Subsection:GRI problems}). First, we trained the trunk network; in this case, the size of $\bm{A}$ (in \Cref{Eq:2-Step trunk output einsum}) is $24\times 95\times (2080\times 99)$. We trained the parameters of the trunk and $\bm{A}$ up to 20,000 epochs. The \% relative $L_2$ error in the training of the trunk network for three independent runs are $0.0303$, $0.0294$, and $0.0288$. After the trunk training, we trained the parameters of the branch network up to 20,000 epochs with the reconstructed output similar to the syngas problem discussed earlier. The \% relative $L_2$ error after the branch training for three independent runs are $0.0853$, $0.0769$ and $0.0847$ in training and $0.0837$, $0.0754$ and $0.0831$ in testing, respectively. The convergence of relative $L_2$ error in trunk training and branch training and testing is shown in \Cref{Figure:GRI:Convergence DOK} (in \Cref{Appendix:Subsection:GRI problems}).
\par
\begin{figure}[H]
    \centering
    \includegraphics[width=1\textwidth]{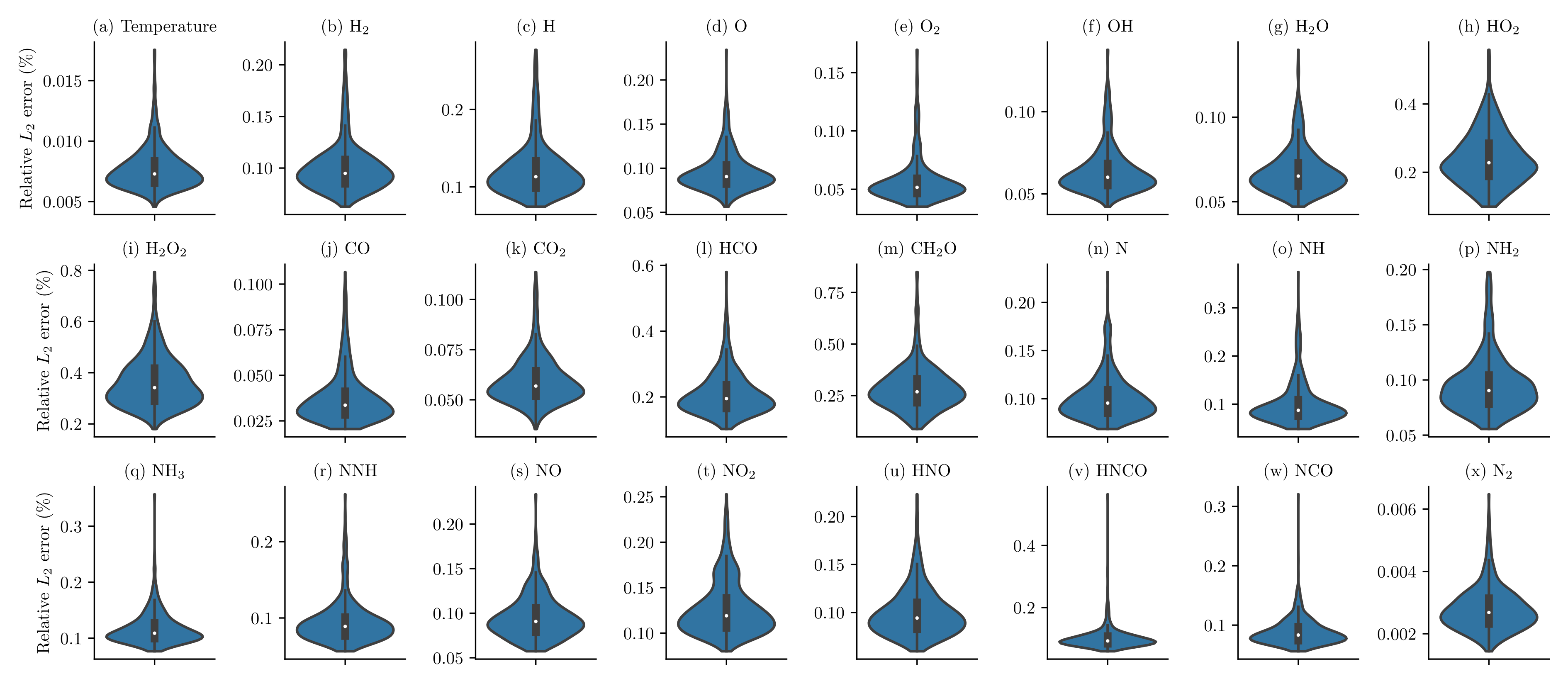}
    \caption{\textbf{GRI problem, Violin plot} of the relative $L_2$ error of the reconstructed prediction of the state variables of the test dataset. The relative $L_2$ errors are calculated similarly as discussed in \Cref{Subsection:Numerical:Syngas:Result analysis}}
    \label{Figure:GRI:Violin plot}
\end{figure}

\begin{figure}[H]
    \centering
    \begin{subfigure}[b]{1\textwidth}
    \includegraphics[width=1\textwidth]{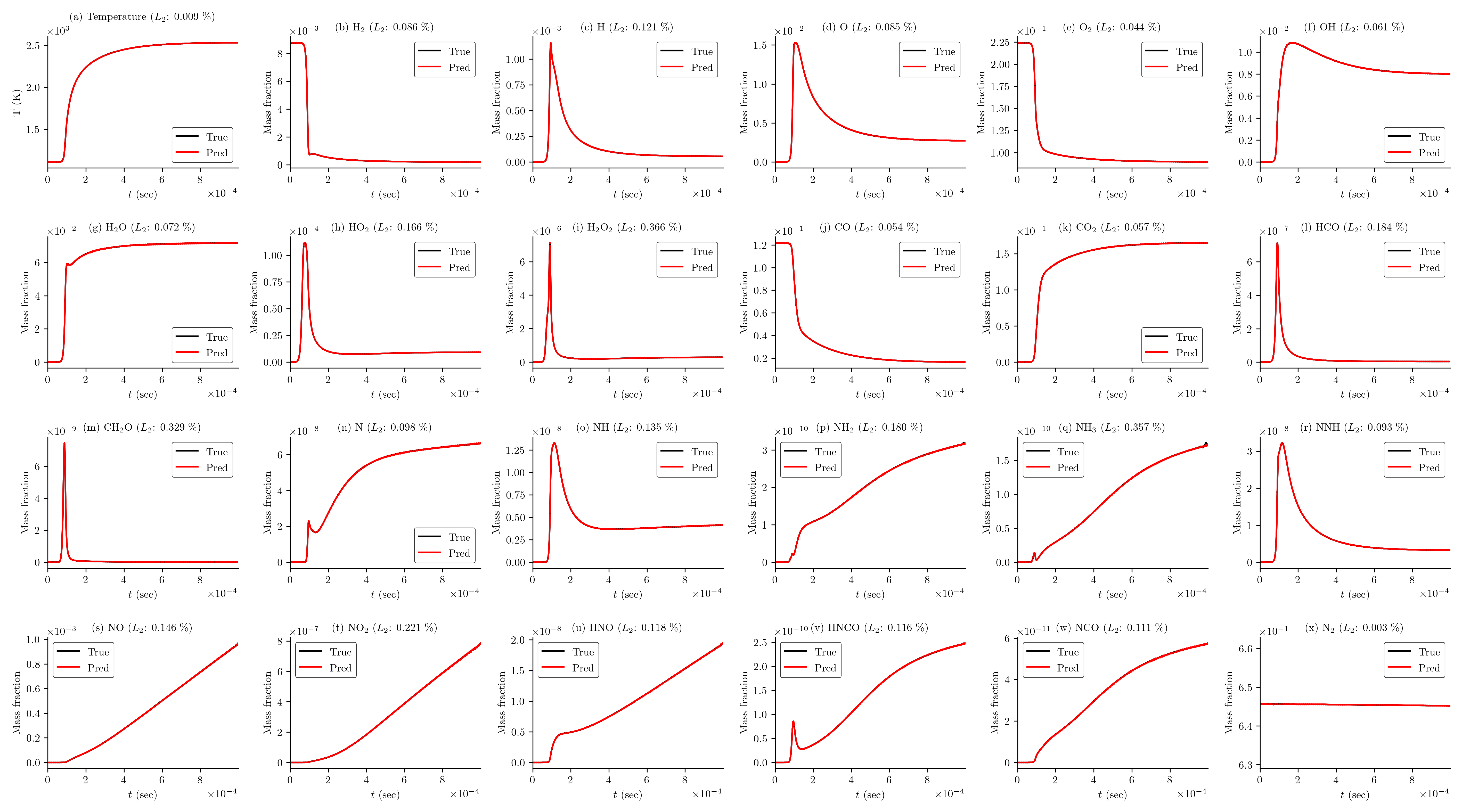}
    \caption{Sample 1}
    \end{subfigure}
    \caption{\textbf{GRI problem, sample result from test dataset.} Plot showing the two sample predicted dynamics of the state variables for GRI problem. The sample result corresponds to one of the higher errors in prediction. The predicted sample shows good accuracy with the true value. Additional sample results are shown in \Cref{Figure:GRI:Sample result:Appendix} (in \Cref{Appendix:Subsection:GRI problems}).}
    \label{Figure:GRI:Sample result}
\end{figure}

 Similar to the discussion in \Cref{Subsection:Numerical:Syngas:Result analysis}, we study the statistics of the relative $L_2$ error of the reconstructed output of the testing dataset. The mean and standard deviation of the relative $L_2$ error for each state variable are shown in \Cref{Table:GRI:Mean and SD for test dataset} (in \Cref{Appendix:Subsection:GRI problems}), and the violin plot of the relative $L_2$ error for each state variable is shown in \Cref{Figure:GRI:Violin plot}. The results show good accuracy across all state variables. A representative dynamics of the test sample is shown in \Cref{Figure:GRI:Sample result}, two additional representative dynamics of the test sample are shown in \Cref{Figure:GRI:Sample result:Appendix} (in \Cref{Appendix:Subsection:GRI problems}).
\section{AMORE implementation in Fourier Neural Operator}
\label{Section:AMORE:FNO:Syngas:FNO}
In the previous sections, we have shown how the proposed AMORE framework can improve the accuracy of DeepONet with multiple outputs. AMORE is a general framework and can be considered with other neural operators. In this section, we will implement AMORE with FNO \citep{Li_2020_FNO}. For this purpose, we will approximate the dynamics of the syngas problem discussed earlier using FNO and train using a non-adaptive MSE loss function (FNO-NA) given by \Cref{Eq:Loss:Data_loss} and an adaptive loss function Type-B (FNO-Ad-B) given by \Cref{Eq:Adaptive loss:Type-B}. We highlight that the objective is not to compare the performance of DeepONet and FNO, as the accuracy depends on a combination of lots of factors, like network size, training strategies, learning rate scheduler, batch size, and number of epochs. The interested reader may refer to a comparative study by \citeauthor{Lu_2022_Comp_deep_fno} \cite{Lu_2022_Comp_deep_fno}.
\par
\begin{figure}[H]
     \centering
    \includegraphics[width=1\textwidth]{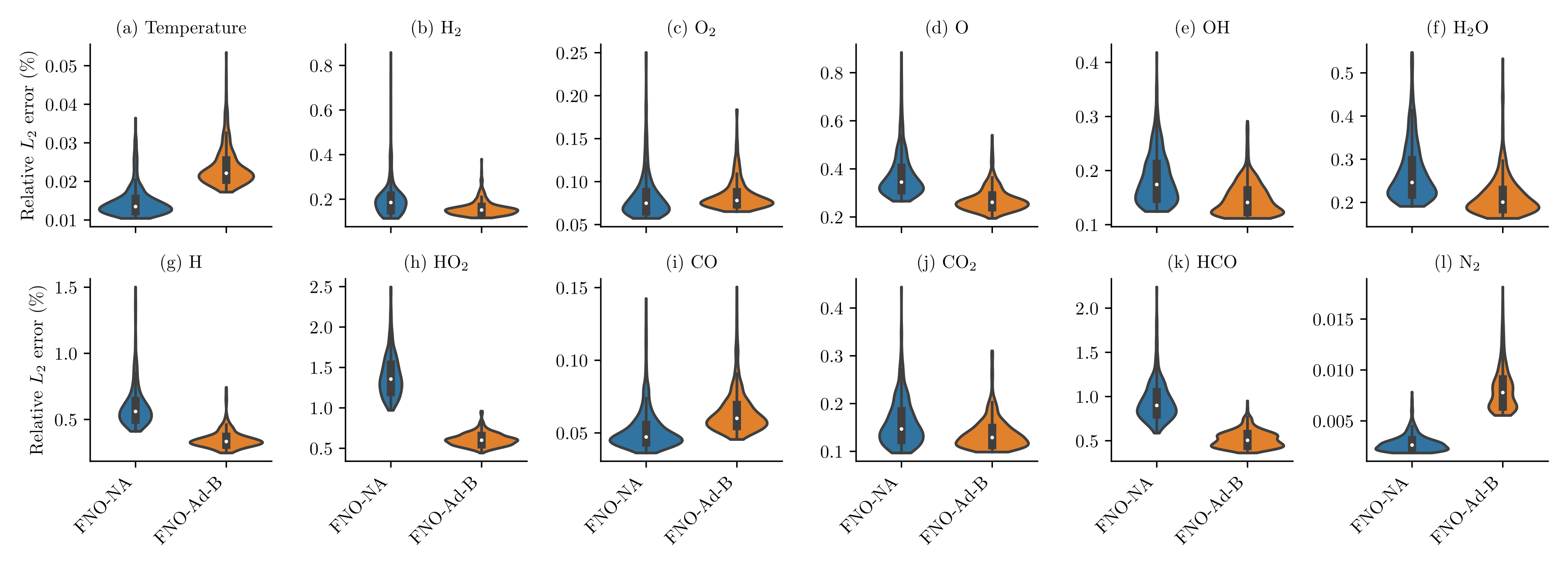}
    \caption{\textbf{Syngas problem, FNO, Violin plot} of the relative $L_2$ error of the reconstructed prediction of the state variables of the test dataset. We observed that the prediction using the adaptive loss function gives better accuracy. The relative $L_2$ errors are calculated similarly as discussed in \Cref{Subsection:Numerical:Syngas:Result analysis}}
    \label{Figure:Syngas:FNO:Violin plot L2 error}
\end{figure}

 We consider an FNO for the syngas problem and train it for 3,000 epochs, which takes around 13.5 hrs, while we trained DeepONet for 15,000 epochs, which takes around 10.5 hrs. The mean relative error in training is 0.17\% and in testing is 0.17\% when considering non-adaptive loss (\Cref{Eq:Loss:Data_loss}). The mean relative error in training reduces to 0.13\% and in testing is reduced to 0.13\% when trained using adaptive loss function Type-B (\Cref{Eq:Adaptive loss:Type-B}). The convergence of the relative $L_2$ error in training and testing are shown in \Cref{Figure:Syngas:FNO:convergence L2 error} (in \Cref{Appendix:Subsection:AMORE in FNO}). As discussed earlier, in \Cref{Subsection:Numerical:Syngas:Result analysis}, we study the reconstructed test data. The violin plots of the relative $L_2$ error of the reconstructed prediction of the test dataset for FNO-NA and FNO-Ad-B are shown in \Cref{Figure:Syngas:FNO:Violin plot L2 error}. We observed that, similar to DeepONet, the accuracy of FNO also improves when an adaptive loss function of Type B is used. The mean and standard deviation of the relative $L_2$ error for the reconstructed test dataset for each state variable is shown in \Cref{Table:Syngas:FNO:Mean SD} (\Cref{Appendix:Subsection:AMORE in FNO}). A brief discussion on FNO and its implementation in the present study, including additional results, is included in \Cref{Appendix:Subsection:AMORE in FNO} 
\section{Computational cost}
\label{Section:Numeical:Computational cost}
The training of DeepONet is offline, and once trained, it can predict the dynamics of a stiff chemical system. In the present study, we have considered two different types of DeepONet, i.e., DON, where both branch and trunk are ResNet, and DOK, where the branch is ResNet and the trunk is KAN, and two different paradigms of training methods. The training time for each of the methods is discussed in \Cref{Appendix:Section:Numeical:Computational cost}. We observed that the training time for the adaptive loss functions is marginally higher than that of the non-adaptive loss function. Furthermore, the training time for the two-step training is higher than for one-step training.
\par The training of DeepONets is offline, thus it does not affect the prediction time or eventual integration of DeepONet with a CFD solver. As discussed earlier, the mass-conserving DeepONet is computationally more expensive than other DeepONets as it requires the inverse mapping. Future studies may focus on reducing computational time in the inverse mapping. However, it may be noted that when integrating with a CFD solver, it is not necessary to consider inverse mapping at all the time points. One may choose to use the mapping only at the time point where the CFD integration is performed.
\section{Summary and Conclusion}
\label{Section:Summary}
We have proposed an adaptive multi-output operator network (AMORE) framework for the accurate prediction of multiple state variables in stiff chemical kinetics using a single neural operator. We consider the Deep Operator Network (DeepONet) to predict all the thermochemical state variables. We consider two DeepONet architectures, the first one with ResNet for both the branch and trunk networks, the second one with ResNet as the branch network and KAN for the trunk network. The proposed DeepONet architectures can predict the dynamics of multiple state variables from a single DeepONet. We design our trunk network such that the partition of unity (PoU) condition is automatically satisfied. To improve accuracy, we proposed two gradient-free adaptive loss functions. The idea behind these adaptive loss functions is that state variables and/or samples with higher error need to be weighted more in the optimization process. The weights are evaluated based on the error in the previous epoch. We consider the two-step training of DeepONet for multiple outputs, where each output variable has its own individual basis function and coefficients. Furthermore, we also extend the proposed adaptive loss functions to two-step training of DeepONet for both the trunk and branch networks. We know that the total mass of a chemical system is conserved, i.e., the total mass of the reactants and products is the same. We consider three different ways of imposing this constraint in the network learning. First, as a soft constraint, an additional loss function is used along with the data loss. Second, using a coordinate transformation map which maps mass fractions to $n-1$ coordinates and guarantees that the sum of the predicted mass fractions is unity. The third one uses a softmax function on the DeepONet output to ensure that the sum of mass fraction is unity. We like to mention that mechanisms such as the GRI-Mech 3.0, the summation of all the species considered is not unity due to non-active or non-significant states. Similarly, in the ROBER problem \citep{Goswami_2024_Stiff_Kinetics}, although mass is conserved, the sum of the state variables is not unity. In these situations, the CoM loss function may be ignored or suitably modified. Furthermore, the CoM using the softmax function may not be possible in the case of two-step training, as the trunk and branch need to be separated during the training of DeepONet. 
\par We have shown that the proposed DeepONet architectures within our AMORE framework can predict the dynamics of the state variables (i.e., temperature and mass fraction) with good accuracy. The computational examples with 12 (Syngas problem) and 24 (GRI-Mech 3.0 problem) output state variables show the accuracy, applicability, and scalability of the proposed method. We have observed that the adaptive weights in the loss function improve prediction accuracy. Furthermore, the DeepONet with KAN as the trunk network provides more reliable predictions, with smaller errors in the mean and standard deviation of the test samples. The DeepONet trained using two-step training provides marginally better results than single-step training. However, the training time for two-step training is higher than that of single-step training. 
\par We tested our proposed DeepONet for out-of-distribution results and recursive/auto-regressive prediction. The results are within acceptable error limits. We also like to mention that in the case of extrapolation, the bounds imposed in the output by $1.05\times\tanh(\bm{\hat{Y}}(\bm{\Theta, \Phi})$ (in \Cref{Eq:Output restriction}) limit the extrapolation limit. However, simply expanding this limit may not improve extrapolation performance, as neural networks are generally poor at extrapolation. In a previous study by \citeauthor{Zhu_2023_Reliable_extrapolation} \cite{Zhu_2023_Reliable_extrapolation}, the authors studied extrapolation using DeepONet and proposed various fine-tuning methods, such as physics-based fine-tuning, sparse new observations, and multi-fidelity learning with sparse new data. Future research may focus on extrapolation using the proposed architecture for larger extrapolation values. In the case of auto-regressive prediction, we considered the entire time series for auto-regressive prediction, which is a highly conservative approach. However, through our computational experiment on the syngas problem using DeepONet-based models, we observed that the relative $ L_2$ error is less than 5\% at the 90th percentile of the test cases, showing their usefulness as a surrogate for stiff chemical kinetics and their potential as a hybrid model for an operator-CFD solver.
\par  The proposed AMORE framework can act as a foundation block for future studies of coupled operators and CFD models, where the operator will simulate the solution of the stiff chemical system. The obvious future research step is to combine neural operators with a CFD solver. In this case, future research can also focus on developing a hybrid model \citep{Nath_2024_Digital} that combines the proposed DeepONet and the Physics-Informed Neural Network (PINN) \citep{Raissi_2019_PINN}. The DeepONet can act as a surrogate for the stiff chemical kinetics, and PINN acts as a solver for fluid flow. We would also like to mention that the mass-conserving DeepONet is computationally expensive compared to the other proposed DeepONet architectures because of the inverse mapping. Future studies may also consider developing strategies for reducing the computation time for the inverse mapping.
\par The proposed AMORE framework (with self-adaptive loss functions) can effectively reduce the error for multiple output state variables for stiff chemical kinetics. The methodology and the proposed framework is universal in a way that it can be considered for other systems and operators. We have demonstrated this by considering FNO within the AMORE framework. Future research directions could also focus on reducing the training cost, particularly for two-step training, and implementing it on other systems.
\par In the present study, we observed that for the considered network configuration and hyperparameters, the adaptive loss functions show better accuracy. Future research may focus on the implementation of the AMORE framework for DeepONet with network configurations such as a CNN in a branch network with multiple outputs. Future research may also focus on extending the AMORE framework for physics-informed operators network with multiple outputs.

\section*{Data availability}
Data will be made available upon request to the corresponding author.

\section*{Acknowledgement}
We want to acknowledge the support from the Small Business Technology Transfer (STTR) program, USA (Grant No: HDTRA224C0001). This research was conducted using computational resources and services at the Center for Computation and Visualization, Brown University. We would also like to thank the anonymous referee who suggested mass conservation using the softmax function.

\section*{Author Contributions Statement}
\textbf{Kamaljyoti Nath}: Conceptualization, Methodology, Validation, Investigation, Visualization, Software, Result generation and analysis, Writing- original draft, Writing – review \& editing; \textbf{Additi Pandey}: Conceptualization, Methodology, Software, Validation, Writing- original draft, Writing – review \& editing; \textbf{Bryan T. Susi}: Conceptualization, Methodology, Supervision, Project administration, Writing – review \& editing; \textbf{Hessam Babaee}: Conceptualization, Methodology, Data curation, Supervision, Writing - original draft, Writing – review \& editing; \textbf{George Em Karniadakis}: Conceptualization, Methodology, Funding acquisition, Project administration, Resources, Supervision, Writing - original draft, Writing – review \& editing. All authors review the manuscript.

\section*{Declaration of competing interest}
The authors declare that they have no known competing financial interests or personal relationships that could have appeared to influence the work reported in this paper.

\bibliography{references}
\bibliographystyle{unsrtnat}

\clearpage
\appendix
\renewcommand{\thesection}{\Alph{section}}
\renewcommand{\thesubsection}{\Alph{section}.\arabic{subsection}}
\renewcommand{\thesubsubsection}{\Alph{section}.\arabic{subsection}.\arabic{subsection}}

\setcounter{equation}{0}
\renewcommand{\theequation}{\thesection.\arabic{equation}}
\setcounter{table}{0}
\renewcommand{\thetable}
{\Alph{section}.\arabic{table}}
\setcounter{figure}{0}
\renewcommand{\thefigure}{\Alph{section}.\arabic{figure}}
\begin{center}
    \Large{Appendices}
\end{center}
\vspace{0.5cm}
\begin{itemize}[leftmargin=*]
    \item \textbf{\Cref{Appedix:Network Architecture}}: A brief discussion on the network architectures considered in the present study.
    \item \textbf{\Cref{Appendix:Section:Additional discussion on Methodology}}: Include additional discussion on the methodology considered.
    \begin{itemize}
        \item \textbf{\Cref{Appendix:Subsection:Two-step training}}: Additional discussion on implementation of two-step training of DeepONet.
        \item \textbf{\Cref{Appendix:SectionCalculation for adaptive loss function}}: Discussion on calculation of adaptive loss function.
        \item \textbf{\Cref{Appendix:Subsection:Mass conserving DeepONet}}: Discussion on implementation of mass-conserving DeepONet.
    \end{itemize}
    \item \textbf{\Cref{Appendix:Section:Additional Results}}: Additional numerical results which are not included in the main content are presented 
    \begin{itemize}
        \item \textbf{\Cref{Appendix:Subsection:Syngas problem}}: Additional results for syngas problem. In \Cref{Appendix:Subsubsection:Error Convergence for Syngas problem}, we have shown the convergence of all the DeepONet method for syngas problem.
        \item \textbf{\Cref{Appendix:Subsection:GRI problems}}: Additional results for GRI-Mech 3.0 problem.
        \item \textbf{\Cref{Appendix:Subsection:AMORE in FNO}}: Implementation and additional result for AMORE in FNO.
    \end{itemize}   
    \item \textbf{\Cref{Appendix:Section:Numeical:Computational cost}}: Additional discussion on computation cost.
\end{itemize}
\section{Network Architecture}
\label{Appedix:Network Architecture}
In the present study, we have considered ResNet as one of the choices for the branch and trunk network in DON, and KAN as one of the choices for the trunk network and ResNet as choice of branch network in DOK. In this section, we briefly discuss the architecture of ResNet and KAN. 
\subsection{ResNet with MLP}
\label{Appendix:Subsection:ResNet}
As discussed earlier, we have considered ResNet (Residual network) with MLP for the branch and the trunk
\begin{wrapfigure}{r}{5.35cm}
    \centering 
    \includegraphics[width=3.5cm]{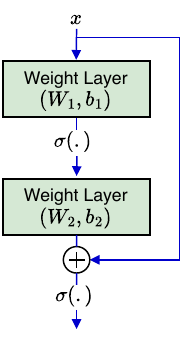}
    \caption{Schematic of a residual block of ResNet showing the residual connection.}
    \label{Figure:Appendix:ResNet}
\vspace{-20pt}
\end{wrapfigure}
networks. ResNet consists of multiple residual blocks with skipped connections after every two layers. We have shown a schematic of a residual block in \Cref{Figure:Appendix:ResNet}. ResNet helps in training very deep networks, preventing overfitting, better gradient flow, and easier optimization. The residual block with input $\bm{x}$ can be written as
\begin{subequations}
\begin{align}
  \bm{z}_1 = & \sigma(\bm{x}\bm{W}_1+\bm{b}_1)
  \label{Eq:ResNet eq 1} \\
  \bm{z}_2 = & \bm{z}_1\bm{W}_2+\bm{b}_2
  \label{Eq:ResNet eq 2} \\
  \bm{z} = & \sigma(\bm{z}_2 + \bm{x})
  \label{Eq:ResNet eq 3}
\end{align}
\end{subequations}
where $\bm{W}_1$ and $\bm{b}_1$ are the weights and biases of the first layer of the residual block and $\bm{W}_2$ and $\bm{b}_2$ are the weights and biases of the second layer of the residual block, $\sigma$ is the activation function considered. We also like to mention that in the first residual block, there might be a mismatch in the dimension. Thus, the summation in \Cref{Eq:ResNet eq 3} may not be possible as in the present study. In this case, we consider a linear projection network (no activation function) with trainable weights and biases. In \Cref{Table:Syngas: network size}, we have shown the number of parameters for the ResNet considered. The second term in the number of parameters refers to the parameters of the projection network.
\subsection{Kolmogorov-Arnold Networks: KAN}
\label{Appendix:Subsection:KAN}
Based on the Kolmogorov Arnold theorem, \citeauthor{Liu_2024_KAN} \cite{Liu_2024_KAN} proposed Kolmogorov Arnold networks (KAN). The basic idea of KAN is that a multivariate continuous function can be written as a sum of the finite composition of continuous functions of the single variables. In the present study, we have considered the Jacobi polynomials as the choice of polynomial function. Below, we have shown the implementation of KAN with the Jacobi polynomials in pseudo-Python code 

\begin{lstlisting}[language=Python]
# Python implementatiaon
A = t   # time after normalization, size: (n_t+1, 1)
for i in range(L):                      # L: number of Layers
    A = sin(A)
    poly = Polynomial(A)                # Output size: (n_t+1, in_dim, order +1)
    A = einsum("ijk,jlk->il", poly, coeff[i])
\end{lstlisting}
where $\text{t}\in\mathbb{R}^{(n_t+1)\times 1}$ is the input time points, L is the number of layers in KAN, and ``Polynomial" is the function that generates the polynomial of A of desired order. The "coeff[i]" are the trainable parameters of the i\textsuperscript{th} layer with size $\text{in\_dim}\times\text{out\_dim}\times\text{(order+1)}$. The size of ''poly" is $(n_t+1)\times \text{in\_dim} \times \text{(order+1)}$. We consider the polynomials defined by a recursive relationship discussed by \citeauthor{Karniadakis_2005_Spectral_book} \citep{Karniadakis_2005_Spectral_book} with $\alpha$ and $\beta$ values of the polynomial as discussed in \Cref{Table:Syngas: network size}. We would also like to discuss the importance of the $\sin(\:.\:)$ function. We know that the domain of a Jacobi polynomial is $[-1,1]$ while the range is $\mathbb{R}$. Thus, after the first layer, the input to the Jacobi function may not be in $[-1,1]$; therefore, to keep the domain within the range $[-1,1]$, we consider the $\sin(\:.\:)$ function. It will ensure that the domain of the Jacobi function is always $[-1,1]$.
\setcounter{table}{0}
\setcounter{figure}{0}
\setcounter{equation}{0}

\section{Additional discussion on Methodology}
\label{Appendix:Section:Additional discussion on Methodology}
We have discussed AMORE, implementation of DeepONet and self-adaptive loss function in the main text. Here, we present additional discussion which are not included in the main text.
\subsection{Two-step training of DeepONet}
\label{Appendix:Subsection:Two-step training}
\begin{figure}[H]
    \centering
    \includegraphics[width=0.95\textwidth]{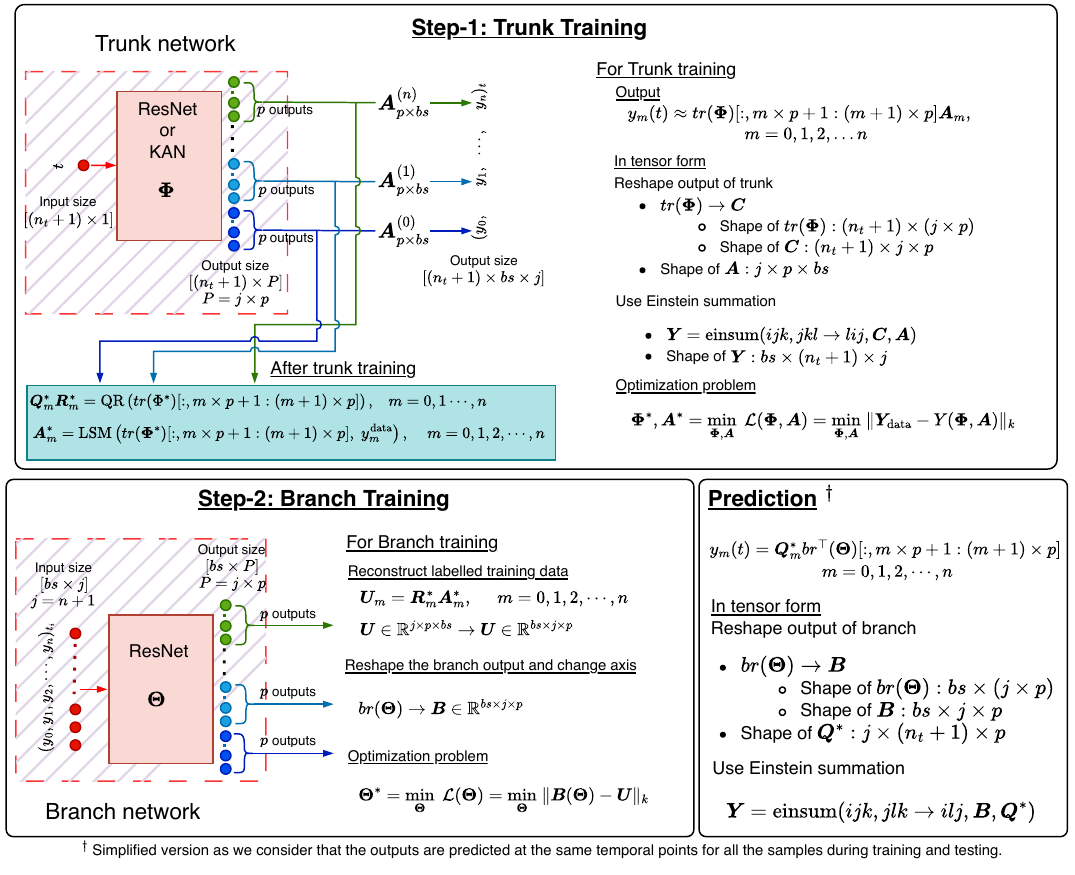}
    \caption{\textbf{Schematic of two-step training of DeepONet for the proposed DeepONet} showing branch and trunk networks and multiple outputs along with training detail. We have shown the trunk network in the top left in the dashed red box with input $t\in \mathbb{R}^{(n_t+1)\times 1}$ and output $tr(\bm{\Phi})\in\mathbb{R}^{(n_t+1)\times P}$. Where $(n_t+1)$ is the number of time points where the output needs to be predicted, and $P=j\times p$ is the number of neurons in the output layer. $p$ is the number of basis functions for each state variable. We have considered ResNet or KAN (with trainable parameters $\bm{\Phi}$) as the choice of network architecture for the trunk network. The output is predicted by multiplying the trunk output with a trainable tensor, as shown in \Cref{Eq:2-Step trunk output einsum}. Thus, in the first step, the optimization process is to obtain the optimal values of the parameters $\bm{\Phi}$ and $\bm{A}$ by minimizing a loss function between the labeled data and predicted output. The output and optimization problem for trunk training is shown in the top right. After the trunk training, the next step is the QR decomposition of the trunk output and recalculation of $\bm{A}$ using the least square approximation. The branch network is shown in the bottom right in the dashed red box with input $\bm{Y}_0\in\mathbb{R}^{bs\times j}$, and output $br(\bm{\Theta})\in\mathbb{R}^{bs\times P}$. Where $bs$ is the number of samples and $P=j\times p$ is the number of neurons in the output layer. $p$ is the number of coefficients for each state variable. We consider ResNet with trainable parameters $\bm{\Theta}$ as the choice of network architecture for the branch network. The parameters of the branch ($\bm{\Theta}$) are optimized using a data loss with respect to a reconstructed labelled data $\bm{U}$ from the trunk output. The optimization process is shown in the bottom-middle. The prediction using the re-parameterized DeepONet is shown in the bottom-right and discussed in \Cref{Subsubsection:Prediction using trained DeepONet}. We also consider adaptive weight in the loss function for both training of trunk and branch networks. These are discussed in \Cref{Subsubsection:Training of trunk network,Subsubsection:Training of branch network} respectively.}
    \label{Figure:2-step DeepONet training}
\end{figure}

We have discussed the implementation of two-step training in \Cref{Subsection:Two-step training}. A schematic diagram for the two-step training of DeepONet is shown in \cref{Figure:2-step DeepONet training}.
We consider three loss functions: the MSE loss function and the two adaptive loss functions. The adaptive loss functions for the trunk training are straightforward and similar to the single-step DeepONet and discussed in \Cref{Subsubsection:Training of branch network}. The implementation of the adaptive loss function for the branch network required marginal modification. The loss function for branch training is discussed below:

\par\noindent\scalebox{1}{$\blacklozenge$} \textbf{Non-adaptive MSE loss function:} This loss function is defined as the mean square error between the branch output ($\bm{B}$) and the reconstructed labeled data ($\bm{U}$) and is similar to the non-adaptive loss function discussed earlier.

\par\noindent\scalebox{1}{$\blacklozenge$} \textbf{Adaptive loss function Type-A:} We define this loss function as data loss with adaptive weights $\mathcal{W}_a$ associated with state variable $a$
\begin{equation}
    \mathcal{L}(\bm{\Theta}) = \sum_{a=0}^{j-1}\mathcal{W}_{a}\left[\dfrac{1}{bs\times p}\sum_{b=1}^{bs}\sum_{c=1}^{p}\left(B_{bac} - U_{bac}\right)^2\right], \;\;\;\;
    \begin{array}{l}
    j\rightarrow \text{No. of states variable}, \\
    p\rightarrow \text{No. of basis functions for each state variable,} \\
    bs\rightarrow \text{No. of samples} \\
    \end{array}
\end{equation}
\par\noindent\scalebox{1}{$\blacklozenge$} \textbf{Adaptive loss function Type-B:} We define this loss function as data loss with adaptive weights $\mathcal{W}_{ba}$ associated with state variable $a$ and sample $b$.
\begin{equation}
    \mathcal{L}(\bm{\Theta}) = \sum_{a=0}^{j-1}\sum_{b=1}^{bs}\mathcal{W}_{ba}\left[\dfrac{1}{ p}\sum_{c=1}^{p}\left(B_{bac} - U_{bac}\right)^2\right], \;\;\;\; \begin{array}{l}
    j\rightarrow \text{No. of states variable}, \\
    p\rightarrow \text{No. of basis functions for each state variable,} \\
    bs\rightarrow \text{No. of samples} \\
    \end{array}
\end{equation}
\par In these cases, as well, the adaptive weights, similar to the weights discussed in \Cref{Subsubsection:Adaptive loss:Type-A,Subsubsection:Adaptive loss:Type-B}, the $\mathcal{W}_a$ and $\mathcal{W}_{ab}$ are updated using a gradient-free update scheme. However, since $\bm{U}$ is not true labeled data, we calculate the relative $L_2$ error from the final output of DeepONet, in the physical domain, and the true labeled data ($\bm{Y}$). The prediction of the final DeepONet output is discussed in the next section (\Cref{Subsubsection:Prediction using trained DeepONet})
\subsection{Calculation for adaptive loss function}
\label{Appendix:SectionCalculation for adaptive loss function}
We have proposed two adaptive loss functions in \Cref{Subsection:Adaptive loss function}, called Type-A (in \Cref{Subsubsection:Adaptive loss:Type-A}) and Type-B (in \Cref{Subsubsection:Adaptive loss:Type-B}). Also, we have considered adaptive loss functions in the two-step training of DeepONet as discussed in \Cref{Subsection:Two-step training,Appendix:Subsection:Two-step training}. In this section, we briefly discussed the calculation of the adaptive loss function.
\par The calculation of the adaptive weights for the loss functions is as follows: the size of the true output and the predicted outputs are $bs\times (n_t+1)\times j$, where $bs$ is the number of samples, $(n_t+1)$ is the number of time points considered (trunk input size), and $j$ is the number of state variables. In the case of adaptive loss of Type-A, the $X_a$ is calculated as 
\begin{equation}
    X_j = \text{mean}\left(\dfrac{L_2\;\text{norm} \left(\bm{Y}_{bs\times (n_t+1)\times j} - \bm{\hat{Y}}_{bs\times (n_t+1)\times j},\;\; \text{axis} = \text{time axis}\right)}{L_2 \;\text{norm} \left(\bm{Y}_{bs\times (n_t+1)\times j},\;\;  \text{axis} = \text{time axis}\right)} , \;\;\;\; \text{axis} = bs\right)
\end{equation}
Similarly, the $X_{ba}$ in the case of adaptive loss function Type-B is calculated as
\begin{equation}
    X_{bs,j} = \dfrac{L_2\;\text{norm} \left(\bm{Y}_{bs\times (n_t+1)\times j} - \bm{\hat{Y}}_{bs\times (n_t+1)\times j},\;\; \text{axis} = \text{time axis}\right)}{L_2 \;\text{norm} \left(\bm{Y}_{bs\times (n_t+1)\times j},\;\;  \text{axis} = \text{time axis}\right)}
\end{equation}
\par As discussed in \Cref{Subsection:Adaptive loss function}, the sum of the adaptive weights is constant. In the case of Type-A, adaptive loss function, the sum of all the weights $\mathcal{W}_a$ at any epoch is always $\mathcal{R}_1$ as shown in \Cref{Eq:Adaptive loss:Type-A W_a sum}.
\begin{equation}
    \sum_{a=0}^{j-1}\mathcal{W}_a = \displaystyle\sum_{a=0}^{j-1} \dfrac{X_a}{\displaystyle\sum_{d=0}^{j-1}X_d} \times \mathcal{R}_1 = \dfrac{\displaystyle\sum_{a=0}^{j-1} X_a}{\displaystyle\sum_{d=0}^{j-1}X_d} \times \mathcal{R}_1 = \mathcal{R}_1
    \label{Eq:Adaptive loss:Type-A W_a sum}
\end{equation}
Similarly, we can show that in the case of Type-B adaptive loss function also the summation of all the adaptive weights $\mathcal{W}_{ab}$ is $\mathcal{R}_2$ at any epoch.

\subsection{Discussion on implementation of mass-conserving DeepONet}
\label{Appendix:Subsection:Mass conserving DeepONet}
In \Cref{Subsection:Mass conserving DeepONet}, we have discussed the basic philosophy of the mass-conserving DeepONet. In this section, we discuss the implementation of the mass-conserving DeepONet. The method involved mapping the mass fraction to a lower dimension, training the DeepONet in the lower dimension, and converting back to the physical domain using an inverse mapping. 
\par Our approach to constructing such maps is inspired by mapping widely used in spectral element simulation of PDES \citep{Karniadakis_2005_Spectral_book}, where a triangle is mapped to a square, i.e., a collapsed two-dimensional coordinate system is mapped to a square. We extend such mappings to $n$-dimensional coordinates. In particular, the forward mapping (ref. \Cref{Eq:CoM DeepONet forward map})
\begin{equation}
    z_1, z_2, \cdots z_{n-1} = g(y_1, y_2, \cdots y_{n}),
    \label{Eq:CoM DeepONet forward map:Append}
\end{equation}
is given by:
\begin{subequations}
\label{Eq:CoM DeepONet forward map detail:Append}
\begin{align}
    z_k &= \frac{y_k}{1-\sum_{j=1, j \neq k}^{n-1} y_j }, \quad 1 \leq k \leq n-2\\
    z_{n-1} &= y_{n-1}
\end{align}
\end{subequations}
\par To find the inverse map ($g^{-1}( \cdot ) $) (ref. \Cref{Eq:CoM DeepONet inverse map})
\begin{equation}
    y_1, y_2, \cdots y_{n} = g^{-1}(z_1, z_2, \cdots z_{n-1}),
    \label{Eq:CoM DeepONet inverse map:Append}
\end{equation}
we start from the forward map
\begin{equation}
    z_k = \frac{y_k}{d_k}, \quad  k = 1, \dots, n-2, \quad \mbox{where} \quad  d_k = 1 - \sum_{j=1, j \neq k}^{n-1} y_j,
\end{equation}
By substituting $y_k = z_k d_k $, we obtain:
\begin{equation}
    d_k = 1 - \sum_{j=1, j \neq k}^{n-1} z_j d_j.
\end{equation}
This results in a system of linear equations:
\begin{equation}
    d_k + \sum_{j \neq k}^{d} z_j d_j = 1, \quad \text{for } k = 1, \dots, n-2.
\end{equation}
This system can be written in matrix form as:
\begin{equation}
\bm{A} \bm{d} = \bm{b},
\end{equation}
where:
 $ \bm{d} = [d_1, d_2, \dots, d_{n-2}]^T $ is the vector of unknowns, $ \bm{b} = [1, 1, \dots, 1]^T $ is a vector of ones, and $ \bm A $ (not same as considered in two-step training) is a $ (n-2) \times (n-2) $ coefficient matrix defined as:
\begin{equation}
    \bm A_{kj} =
    \begin{cases}
        1, & \text{if } k = j, \\
        z_j, & \text{if } k \neq j.
    \end{cases}
\end{equation}
Once the above linear system is solved, the mass fraction vector can be computed via:
\begin{subequations}
\begin{align}
    y_k &= z_k d_k, \quad 1 \leq k \leq n-2\\
    y_{n-1} &= z_{n-1}\\
    y_n &= 1 - \sum_{k=1}^{n-1} y_k.
\end{align}
\end{subequations}
\par To implement the above transformation in the DeepONet, the first step is to convert the mass fraction using the forward map, \Cref{Eq:CoM DeepONet forward map:Append}. The next step is to formulate the DeepONet learning problem, which can be written as
\begin{equation}
    \bm{Z}(t) \approx \mathcal{G}_\theta : \bm{Z}_0 \mapsto \bm{Z}(\bm{Z}_0)(t)
\end{equation}
where $\bm{Z}(t) = [T(t), z_1(t), z_2(t), \cdots z_{n-1}(t)]^\top$ and $\bm{Z}_0 = [T, z_1, z_2, \cdots z_{n-1}]^\top_0$. Once the DeepONet is trained, we convert the predicted output to the physical domain using the inverse mapping, \Cref{Eq:CoM DeepONet inverse map:Append}. We also want to mention that in addition to the forward and inverse mapping, we also consider normalization of the input and output in the transformed domain. These are discussed in computational examples.
\setcounter{table}{0}
\setcounter{figure}{0}
\setcounter{equation}{0}
\section{Additional Results}
\label{Appendix:Section:Additional Results}
\subsection{Additional result for the syngas problem}
\label{Appendix:Subsection:Syngas problem}
\begin{figure}[H]
    \centering
    \includegraphics[width=1\textwidth]{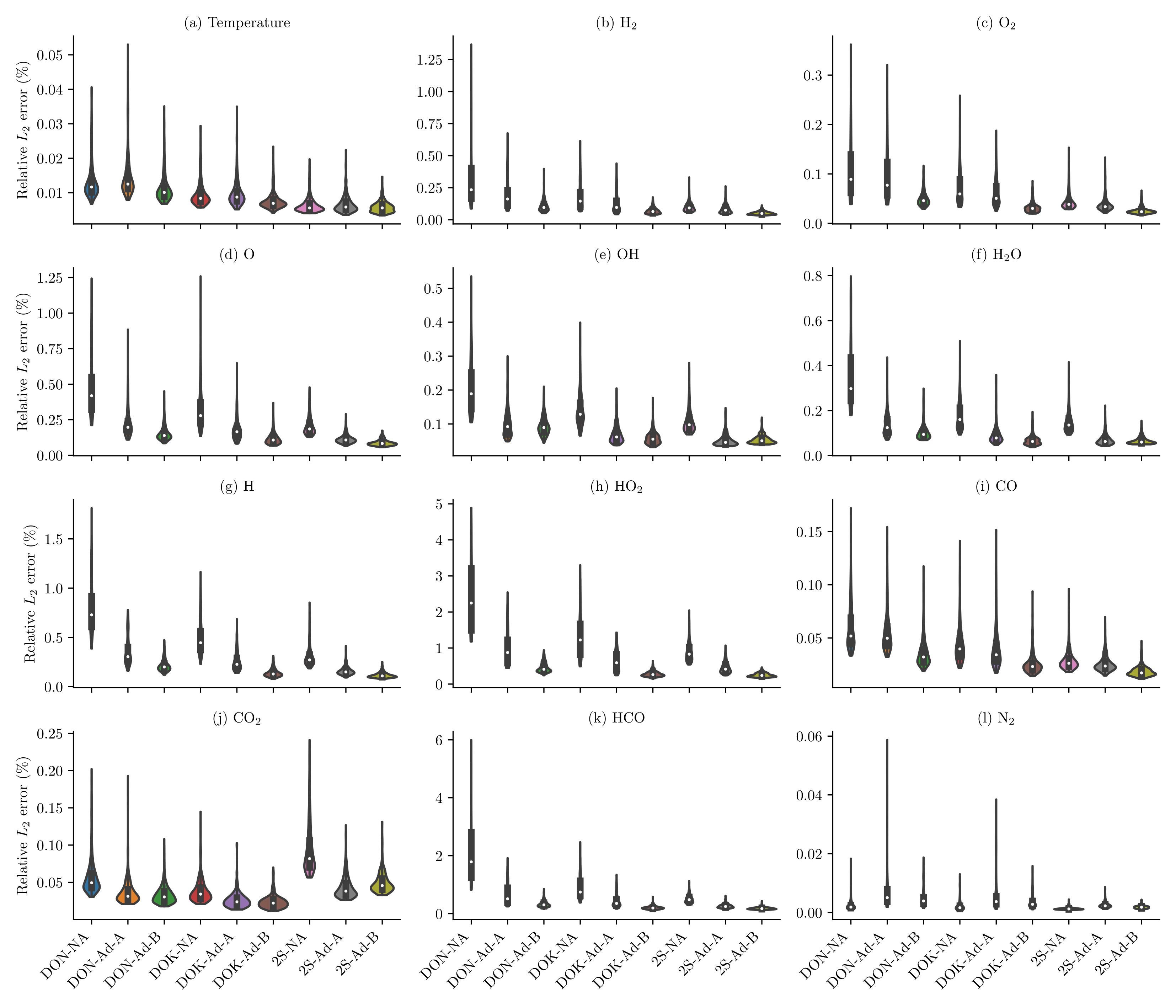}
    \caption{\textbf{Syngas problem, Violin plot} of the relative $L_2$ error of the reconstructed prediction of the state variables of the test dataset. We observed that the predicted results using DON-NA have the highest mean of the relative $L_2$ error compared to the other methods. Furthermore, DON-NA also has a higher standard deviation of the relative $L_2$ error. The mean and standard deviation are reduced as adaptive loss functions are considered. We would particularly like to mention that the state variables corresponding to the dynamics of HCO and HO\textsubscript{2} shown in (k) and (h), respectively, have higher errors, which are reduced when we consider the adaptive loss function. We have observed that the DON-Ad-B and 2S-Ad-B have similar accuracy; however, 2S-Ad-B has smaller standard deviations.}
    \label{Figure:Syngas:Violin plot all 9 method}
\end{figure}

We have discussed the results for the syngas problem in \Cref{Section:Numerical results:Syngas Problem}. In this section, we will present additional results for the syngas problem not included in \Cref{Section:Numerical results:Syngas Problem}. In \Cref{Table:Syngas problem:Min max training-testing data}, we have shown the minimum and maximum values of the training and testing dataset of the state variables of the syngas problem. In \Cref{Table:Syngas: network size}, we have shown the network detail for DON, and DOK, including activation function, layers, number of parameters, etc. In \Cref{Figure:Syngas:Violin 4 method} (in \Cref{Section:Numerical results:Syngas Problem}), we have shown the violin plots for relative $L_2$ error for four methods considered. Here, in \Cref{Figure:Syngas:Violin plot all 9 method}, we have shown the violin plots for all nine methods considered. The following additional results are presented in this section:
\begin{table}[H]
    \centering
    \caption*{Additional tables and figures for results of the syngas problem.}
    \begin{tabular}{L{2cm}|L{12.5cm}} \hline
    Table/Figure & Brief description \\ \hline
    \Cref{Table:Syngas problem:Min max training-testing data} & Syngas problem, minimum and maximum value of training-testing data \\ \hline
    \Cref{Table:Syngas: network size} & Syngas problem, DeepONet network details \\ \hline
    \Cref{Table:Syngas:2-step Trunk trining error} & Syngas problem, Training error in trunk training in the two-step training of DeepONet \\ \hline
    \Cref{Figure:Syngas:Violin plot all 9 method} & Syngas problem, violin plot of relative $L_2$ error. \\ \hline
    \Cref{Figure:Syngas:2 step: Sample results appendix}  & Syngas problem, additional sample test result when considering 2S-Ad-B \\ \hline
    \Cref{Figure:Syngas:Pointwise error 2S-Ad-B} & Syngas problem, point-wise error for sample test results when considering 2S-Ad-B \\ \hline
    \Cref{Table:Syngas:Mass conserve DeepONet:Relative L2} & Syngas problem, mass conserving DeepONet, relative $L_2$ error for in training and testing \\ \hline
    \Cref{Table:Syngas:Mass conserve DeepONet:Mean and Standard deviation} & Syngas problem, mass conserving DeepONet, mean and standard deviation of test sample for each state variable \\ \hline
    \Cref{Figure:Syngas:Extrapolation sample 2} & Syngas problem, extrapolation results \\ \hline
    \Cref{Table:Syngas:Recursive:Test error} & Syngas problem, testing error for recursive prediction \\ \hline
    \Cref{Figure:Syngas:DeepONet:Recursive prediction:Violin plot}  & Syngas problem, Recursive prediction, Violin plot of relative $L_2$ error\\  \hline
    \Cref{Figure:Syngas:Recursive prediction:Sample results: Appendix}  & Syngas problem, Recursive prediction, sample results \\ \hline
    \multicolumn{2}{l}{\Cref{Appendix:Subsubsection:Error Convergence for Syngas problem}: Convergence of relative $L_2$ error for different methods considered} \\ \hline
    \Cref{Figure:Syngas:Convergence L2 error one step} & Convergence of relative $L_2$ error for one-step training. \\ \hline
    \Cref{Figure:Syngas:Covergence 2-step training L2 trunk training} & Convergence of relative $L_2$ error for trunk training using two-step training. \\ \hline
    \Cref{Figure:Syngas:Convergence:2-step:L2 branch training} &  Convergence of relative $L_2$ error for branch training using two-step training. \\ \hline
    \Cref{Figure:Syngas:Convergence mass conserve DOK} & Convergence of relative $L_2$ error for trunk and branch training in two-step training of mass-conserving DeepOKAN. \\ \hline
    \Cref{Figure:Syngas:Softmax_mass_Conseving_Convergence} & Convergence of relative $L_2$ error for Syngas problem for DeepOKAN with softmax Mass conservation. \\ \hline
    \end{tabular}
    
\end{table}
\par In the case of the syngas problem, in the single-step training, we train the DeepONets up to 15,000 epochs with $60$ mini-batches (i.e., each batch size $3432$) up to a 5,000 epoch, after that $120$ mini-batches (i.e., each batch size $1716$) of the branch input. Similarly, in the two-step training as well, we train the parameters of the trunk ($\bm{\Phi}$) and $\bm{A}$ up to 15,000 epochs with $60$ mini-batches (i.e., each batch size $3432$) up to 5,000 epochs, after that $120$ mini-batches (i.e., each batch size $1716$). We want to mention that, in this case, the mini-batches are considered along the batch-size dimension (or the batch-size axis) of the matrix $\bm{A}$. In the two-step training, we considered the best trunk network architecture/model obtained from the first-stage training, i.e., KAN trained using adaptive loss Type-B. In order to check the influence of the initialization of the network parameters on the accuracy, we train the branch with three independent training runs. Furthermore, for each run, we consider different trunk networks (KAN with Adaptive loss Type-B) obtained in the first stage, i.e., run-1 for the branch considers the trunk obtained in run-1 of the trunk network; run-2 for the branch considers the trunk obtained in run-2 of the trunk network, and run-3 for the branch considers the trunk obtained in run-3. In this case, as well, we consider a mini-batch scheme similar to the previous cases.
\begin{table}[H]
    \centering
    \caption{\textbf{Syngas problem, minimum and maximum value} of the training and testing dataset for the syngas problem for the state variables (temperature and mass fraction for each of the $11$ species).}
    \label{Table:Syngas problem:Min max training-testing data}
    \begin{tabular}{c|c|c|c|c}
    \hline
    \rowcolor{orange!45} & \multicolumn{2}{c|}{Training dataset} & \multicolumn{2}{c}{Testing dataset} \\ \cline{2-5} 
    \rowcolor{orange!45} \multirow{-2}{*}{State variable} & Minimum Value & Maximum Value & Minimum Value & Maximum Value \\\hline
    Temperature & 1.000e+03 & 2.855e+03 & 1.000e+03 & 2.852e+03 \\ \hline
    \rowcolor{cyan!25} H$_2$ & 1.060e-04 & 4.938e-03 & 1.060e-04 & 4.938e-03 \\ \hline
    O$_2$ & 1.977e-02 & 2.033e-01 & 1.985e-02 & 2.033e-01 \\ \hline
    \rowcolor{cyan!25} O & 2.403e-12 & 2.492e-02 & 2.454e-12 & 2.492e-02 \\ \hline
    OH & 1.429e-14 & 1.131e-02 & 1.498e-14 & 1.129e-02 \\ \hline
    \rowcolor{cyan!25} H$_2$O & 3.315e-16 & 3.445e-02 & 3.580e-16 & 3.443e-02 \\ \hline
    H & 8.988e-15 & 1.663e-03 & 9.206e-15 & 1.634e-03 \\ \hline
    \rowcolor{cyan!25} HO$_2$ & 2.707e-13 & 1.134e-04 & 2.765e-13 & 1.134e-04 \\ \hline
    CO & 4.326e-02 & 3.431e-01 & 4.356e-02 & 3.431e-01 \\ \hline
    \rowcolor{cyan!25} CO$_2$ & 6.652e-12 & 2.974e-01 & 6.794e-12 & 2.974e-01 \\ \hline
    HCO & 1.820e-17 & 2.406e-06 & 1.936e-17 & 2.405e-06 \\ \hline
    \rowcolor{cyan!25} N$_2$ & 4.840e-01 & 5.860e-01 & 4.840e-01 & 5.860e-01 \\ \hline
    \end{tabular}
\end{table}

\begin{table}[H]
\centering
\caption{\textbf{DeepONet network details considered for the Syngaas problem.} We consider two different types of architectures. In DON (DeepONet), ResNet is considered in both the branch and trunk, while in DOK (DeepOKAN), ResNet is considered in the branch, and KAN is considered in the trunk. The ResNet size $[12-200\times 10-1140]$ indicates that there are $12$ neurons in the input layer, $10$ hidden layers with $200$ neurons in each layer, and the output layer contains $1140$ neurons. The KAN $[1-75\times 4 -1140]$ indicates $4$ layers with $75$ edges. We consider Jacobi polynomials for KAN. We discussed the ResNet and KAN architecture in \Cref{Appedix:Network Architecture}. The output $1140$ neurons in the output layer are divided into $12$ parts, each part with $95$ neurons for each state variable, i.e., $p=95$. In the ``Number of parameters", the second number, in the case of ResNet, indicates the number of parameters in the projection network required in the first layer due to size mismatch. The ``$\sin(\:.\:)$" in KAN is an additional function and is discussed in \Cref{Appedix:Network Architecture}.}
\label{Table:Syngas: network size}
\renewcommand{\arraystretch}{1.25}
\begin{tabular}{l|L{3.5cm}|C{4.15cm}|C{4.15cm}} \hline
\rowcolor{orange!45}  &  & DON & DOK \\ \hline
& Network Type & ResNet & ResNet \\
& Network size & $[12-200\times 10 - 1140]$ & $[12-200\times 10 - 1140]$ \\
& Activation function & $\text{tanh}(\:.\:)$ & $\text{tanh}(\:.\:)$ \\
\multirow{-4}{*}{Branch}& Number of parameters & $593540+2600$ & $593540+2600$\\ \hline
& Network Type & ResNet & KAN 3rd order Jacobi ($\alpha = 1.0$, $\beta = 1.0$) \\
& Network size & $[1-200\times 10 - 1140]$ & $[1-75\times 4 - 1140]$ \\
& Activation function & $\tanh(\:.\:)$ & $\sin(\:.\:)$ (*ref \Cref{Appedix:Network Architecture}) \\
\multirow{-4}{*}{Trunk}& Number of parameters & $591340+400$ & $409800$\\ \hline
\multicolumn{2}{l|}{Total parameters} & 1187880 (1.2 M) & 1005940 (1.0 M) \\ \hline
\end{tabular}
\end{table}

\begin{table}[H]
\centering
\caption{\textbf{Training error in trunk training in the two-step training of DeepONet for syngas problem.} Mean of relative $L_2$ error in trunk training for the two different trunk architectures considered. i.e., ResNet and KAN. The second column shows the type of network architecture, and the third column shows the type of loss function considered. The last column shows the mean of the relative $L_2$ error (\%) in three independent training runs. The convergence of the means of relative $L_2$ error with epoch is shown in \Cref{Figure:Syngas:Covergence 2-step training L2 trunk training} (in \Cref{Appendix:Subsection:Syngas problem}). }
\label{Table:Syngas:2-step Trunk trining error}
\begin{tabular}{L{2.cm}|C{1.75cm}|C{3cm}|C{3.5cm}} \hline
\rowcolor{orange!45} \multicolumn{4}{c}{Two-step training of DeepONet: Trunk training} \\ \hline
\rowcolor{orange!45} Case & Trunk & Loss function &  Relative $L_2$ error (\%) \\ \hline
ResNet-NA & ResNet &Non-Adaptive & 0.1639, 0.1638, 0.1703 \\ \hline
\rowcolor{magenta!25} ResNet-Ad-A & ResNet & Adaptive: Type-A & 0.0541, 0.0542, 0.0545 \\ \hline
\rowcolor{cyan!25} ResNet-Ad-B & ResNet & Adaptive: Type-B & 0.0120, 0.0143, 0.0132 \\ \hline
KAN-NA & KAN & Non-Adaptive & 0.1499, 0.1527, 0.1505 \\ \hline
\rowcolor{magenta!25} KAN-Ad-A & KAN & Adaptive: Type-A & 0.0281, 0.0309, 0.0287 \\ \hline
\rowcolor{cyan!25} KAN-Ad-B & KAN & Adaptive: Type-B & 0.0081, 0.0079, 0.0088 \\ \hline
\end{tabular}
\end{table}

\begin{figure}[H]
    \centering
    \begin{subfigure}[b]{1\textwidth}
    \includegraphics[width=1\textwidth]{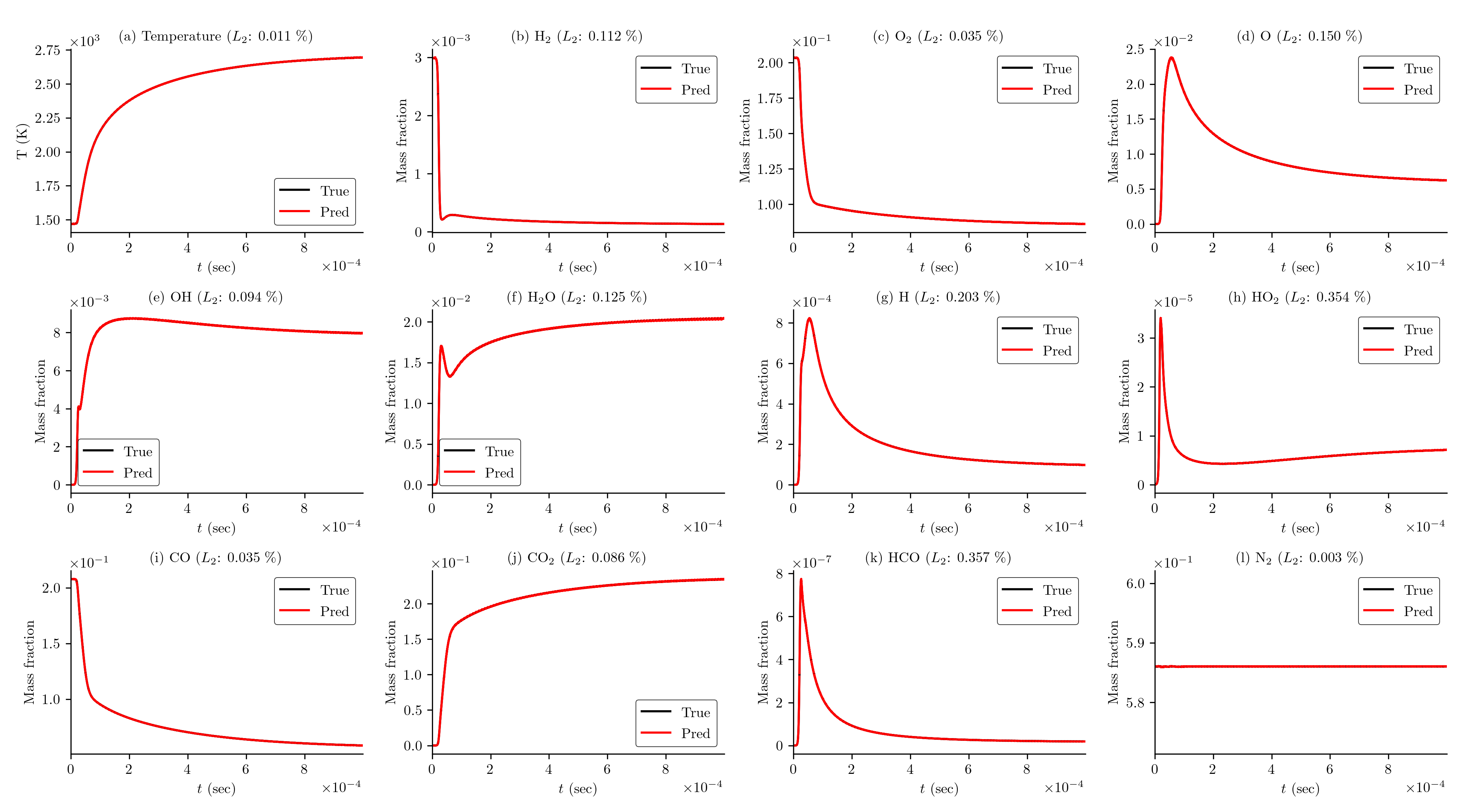}
    \caption{Test Sample 2 (2S-Ad-B)}
    \end{subfigure}
\end{figure}
\begin{figure}[H]\ContinuedFloat
    \begin{subfigure}[b]{1\textwidth}
    \includegraphics[width=1\textwidth]{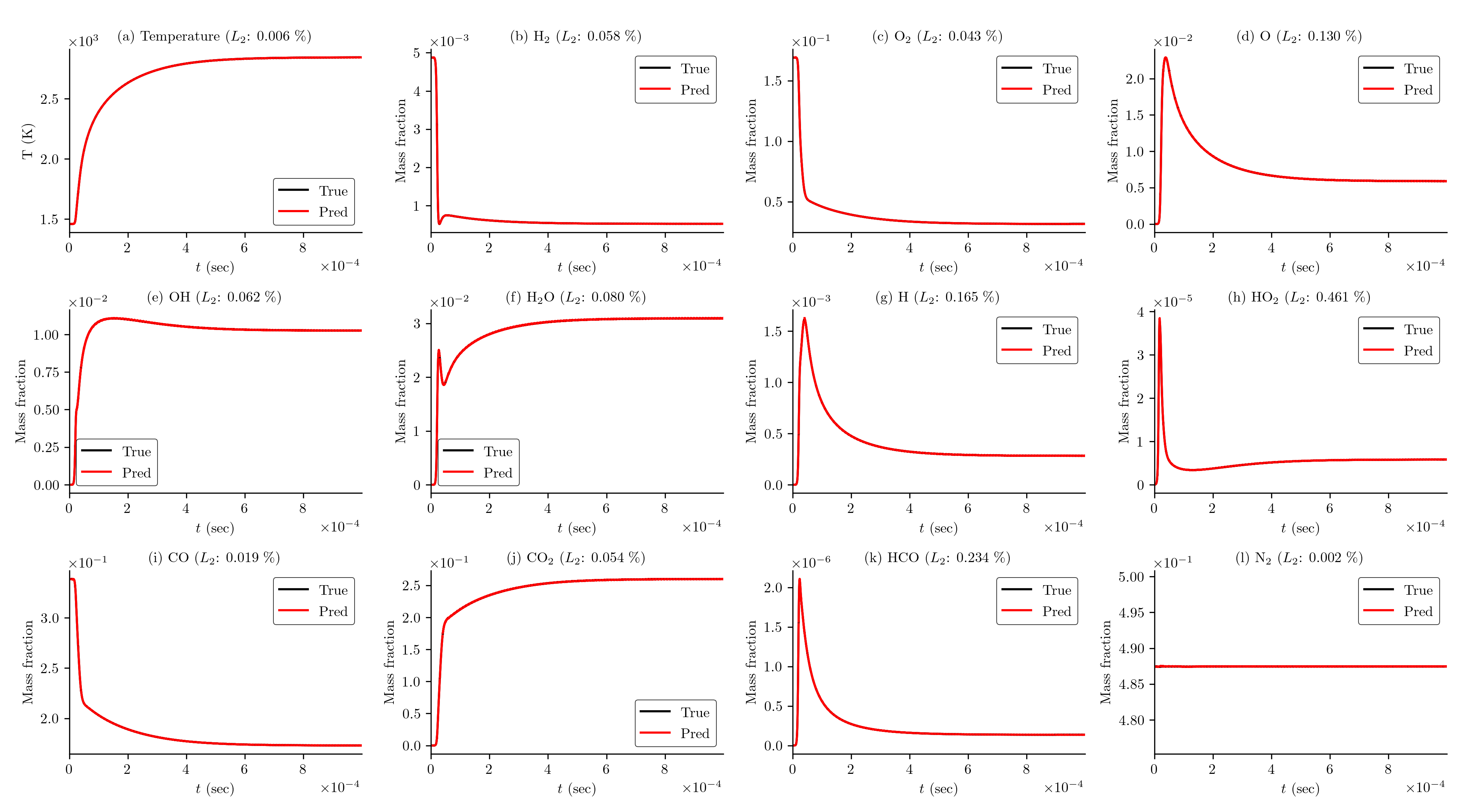}
    \caption{Test Sample 3 (2S-Ad-B)}
    \end{subfigure}
    \caption{\textbf{Syngas problem, additional sample test result, 2S-Ad-B:} Plot showing two representative sample results from the test dataset when predicted 2S-Ad-B, i.e., DeepOKAN trained using the two-step training method and adaptive loss function Type-B. The number in the bracket indicates the relative $L_2$ errors for each species (calculated as discussed in \Cref{Subsection:Numerical:Syngas:Result analysis}). The sample result corresponds to one of the higher errors in prediction. The predicted sample shows good accuracy with the true value.}
    \label{Figure:Syngas:2 step: Sample results appendix}
\end{figure}

\begin{figure}[H]
    \centering
    \begin{subfigure}[b]{1\textwidth}
    \includegraphics[width=1\textwidth]{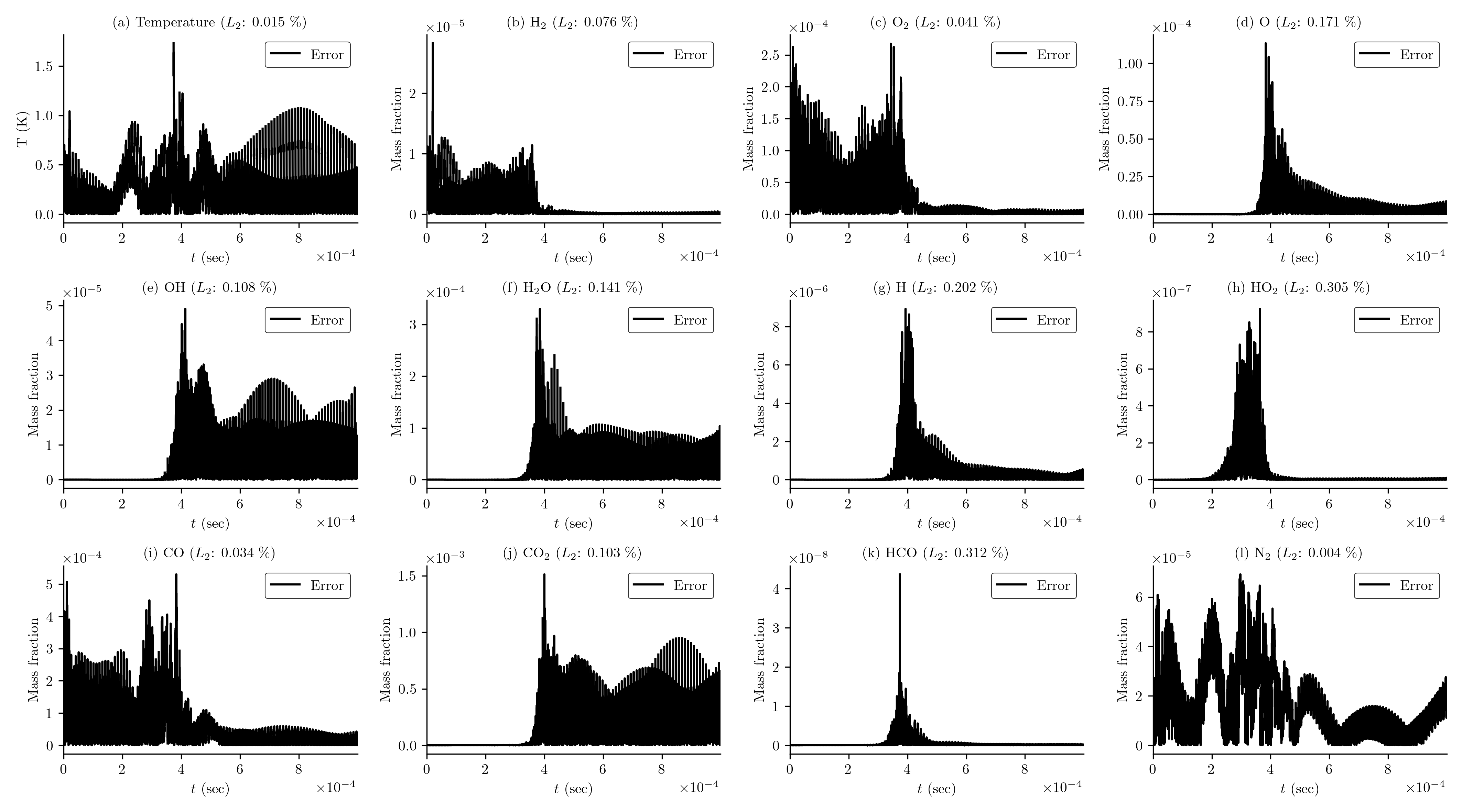}
    \caption{Test Sample 1 (2S-Ad-B)}
    \end{subfigure}
\end{figure}
\begin{figure}[H]\ContinuedFloat
    \centering
    \begin{subfigure}[b]{1\textwidth}
    \includegraphics[width=1\textwidth]{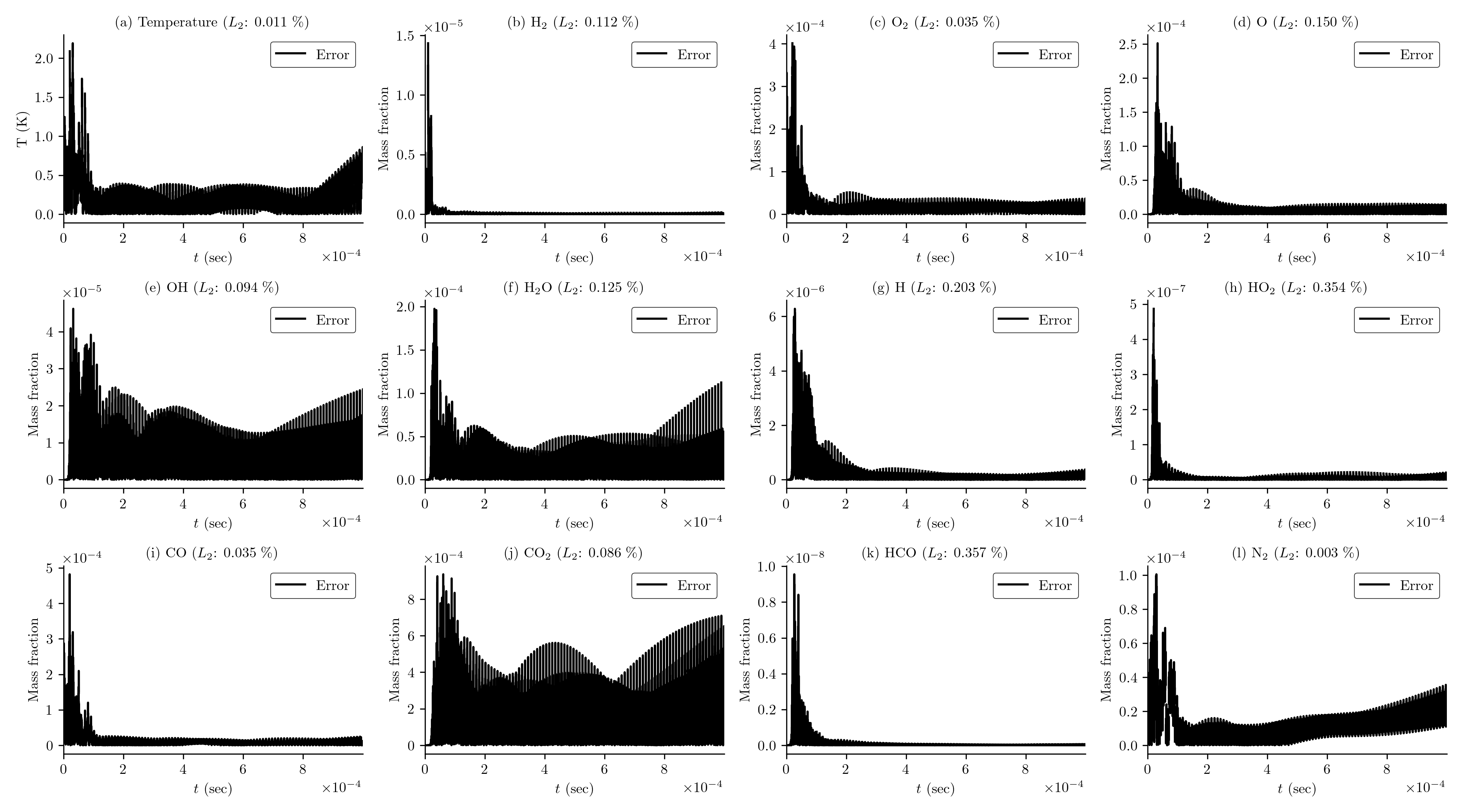}
    \caption{Test Sample 2 (2S-Ad-B)}
    \end{subfigure}
\end{figure}
\begin{figure}[H]\ContinuedFloat
    \centering
    \begin{subfigure}[b]{1\textwidth}
    \includegraphics[width=1\textwidth]{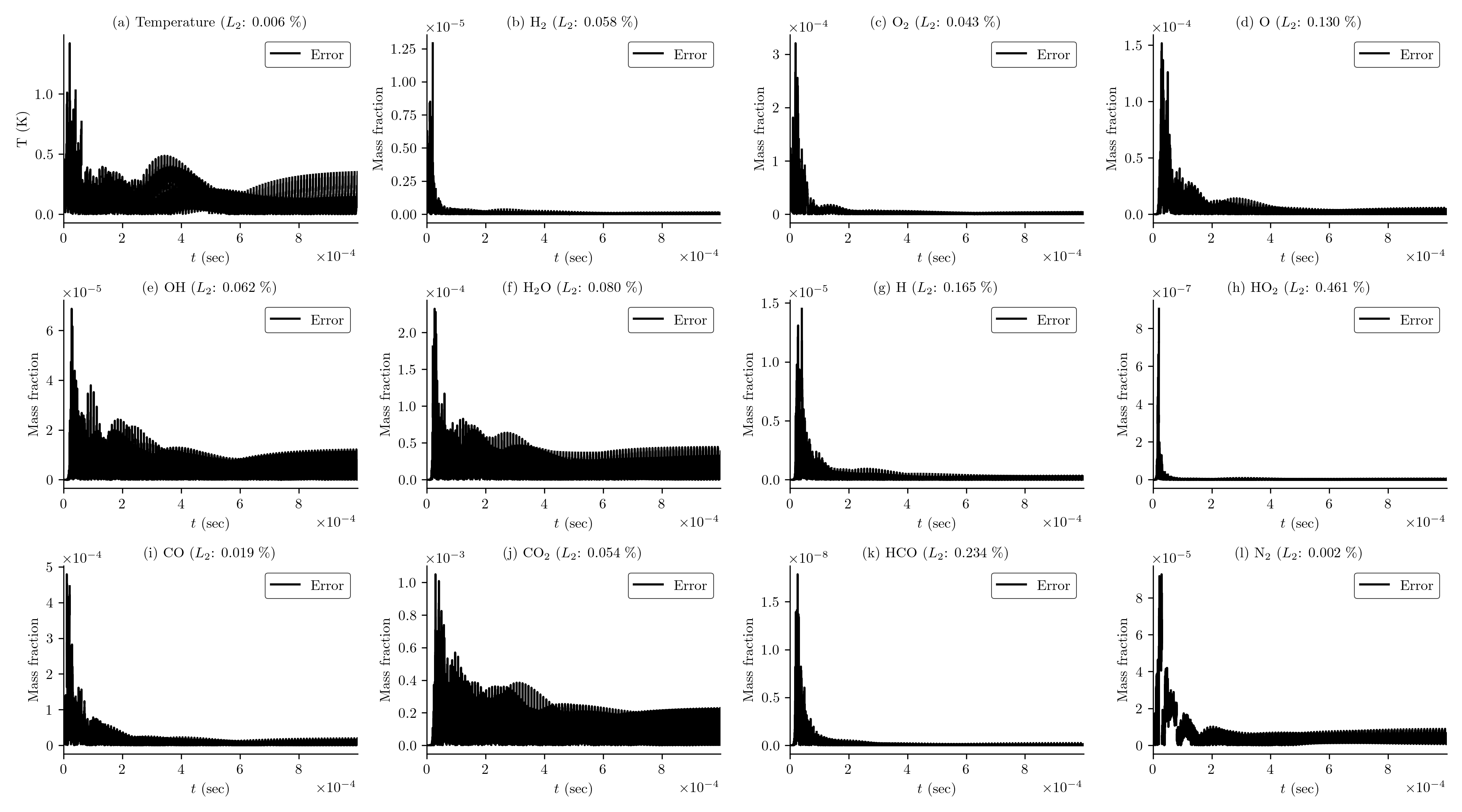}
    \caption{Test Sample 3 (2S-Ad-B)}
    \end{subfigure}
    \caption{\textbf{Syngas problem, 2S-Ad-B:} Plot showing the point-wise error for three representative sample results from the test dataset when predicted using 2S-Ad-B, i.e., DeepOKAN trained using the two-step training method and adaptive loss function Type-B. The number in the bracket indicates the relative $L_2$ errors for each species (calculated as discussed in \Cref{Subsection:Numerical:Syngas:Result analysis}). The sample result corresponds to one of the higher errors in prediction.}
    \label{Figure:Syngas:Pointwise error 2S-Ad-B}
\end{figure}

\begin{table}[H]
\centering
\caption{\textbf{Syngas problem, Mass conserving DeepONet:} Relative $L_2$ error in training and testing for mass conserving DeepONet (\Cref{Subsection:Mass conserving DeepONet}) for syngas problem for three independent runs. For comparison, we have shown the training and testing error correspond to DeepONet trained without considering mass conservation, i.e., 2S-Ad-B. The convergence of the relative $L_2$ error in trunk and branch training for 2S-Ad-B-CoM is shown in \Cref{Figure:Syngas:Convergence mass conserve DOK} (in \Cref{Appendix:Section:Additional Results}).}
\label{Table:Syngas:Mass conserve DeepONet:Relative L2}
\begin{tabular}{L{2.5cm}|C{3.15cm}|C{3.15cm}}
 \hline
\rowcolor{orange!50} &  \multicolumn{2}{c}{Relative $L_2$ error (\%)}\\  \cline{2-3}
\rowcolor{orange!50} \multirow{-2}{*}{Case} & Training & Testing \\ \hline
2S-Ad-B & 0.041, 0.040, 0.041 & 0.041, 0.040, 0.041 \\ \hline
\rowcolor{cyan!25} 2S-Ad-B-CoM & 0.035, 0.037, 0.034 & 0.035, 0.037, 0.034 \\ \hline
\end{tabular}
\end{table}

\begin{table}[H]
\centering
\caption{\textbf{Syngas problem, mass conserving DeepONet, mean and standard deviation of the \% relative $L_2$ error} for the test samples when predicted using mass-conserving DeepONet, i.e., 2S-Ad-B-CoM. For comparison, we have also shown the mean and standard deviation of the \% relative $L_2$ error for the test sample when predicted 2S-Ad-B. The violin plots of the relative $L_2$ error of 2S-Ad-B-CoM and 2S-Ad-B are shown in \Cref{Figure:Syngas:Mass conserving DeepONet:Violin plot}. The relative $L_2$ errors are calculated similarly to those discussed in \Cref{Subsection:Numerical:Syngas:Result analysis}.}
\label{Table:Syngas:Mass conserve DeepONet:Mean and Standard deviation}
\resizebox{\textwidth}{!}{%
\renewcommand{\arraystretch}{1.25}
\begin{tabular}{C{1.7cm}|c|c|c|c|c|c|c|c|c|c|c|c|c} \hline
\rowcolor{orange!45}Case & & T & H$_2$ & O$_2$ & O & OH & H$_2$O & HO$_2$ & H & CO & CO$_2$ & HCO & N$_2$ \\ \hline
& $\mu$ & 0.006  & 0.052 & 0.026 & 0.088 & 0.054 & 0.066 & 0.116 & 0.25 &  0.018 & 0.049 & 0.182 & 0.002 \\ \cline{2-14}
\multirow{-2}{*}{2S-Ad-B} & $\sigma$ & 0.002 & 0.013 & 0.007 & 0.021 & 0.013 & 0.017 & 0.027 & 0.056 & 0.005 & 0.013 & 0.044 & 0.001 \\ \hline

\rowcolor{cyan!25} & $\mu$ & 0.004 & 0.048 & 0.025 & 0.07 & 0.045 & 0.051 & 0.086 & 0.178 & 0.019 & 0.04 &  0.144 & 0.013 \\ \cline{2-14}
\rowcolor{cyan!25}  \multirow{-2}{*}{\begin{tabular}{c}2S-Ad-B\\-CoM\end{tabular}} & $\sigma$ & 0.001 & 0.013 & 0.007 & 0.017 & 0.01 & 0.011 & 0.022 & 0.047 & 0.003 & 0.009 & 0.035 & 0.003 \\ \hline

\end{tabular}}
\end{table}

\begin{figure}[H]
    \centering
    \begin{subfigure}[b]{1\textwidth}
    \includegraphics[width=1\textwidth]{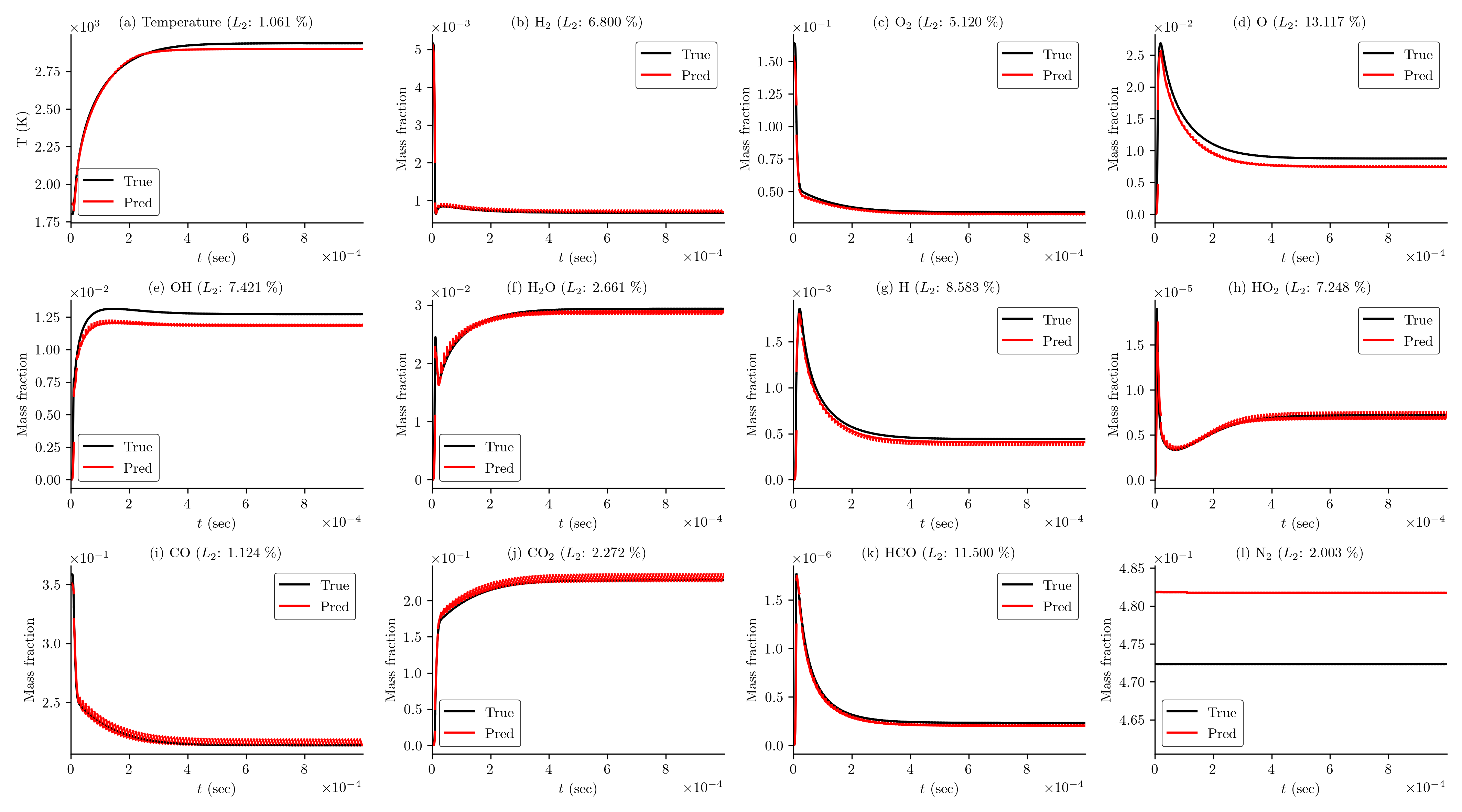}
    \caption{DOK-Ad-B, Extrapolation sample \# 2}
    \end{subfigure}
\end{figure}
\begin{figure}[H]\ContinuedFloat
    \begin{subfigure}[b]{1\textwidth}
    \includegraphics[width=1\textwidth]{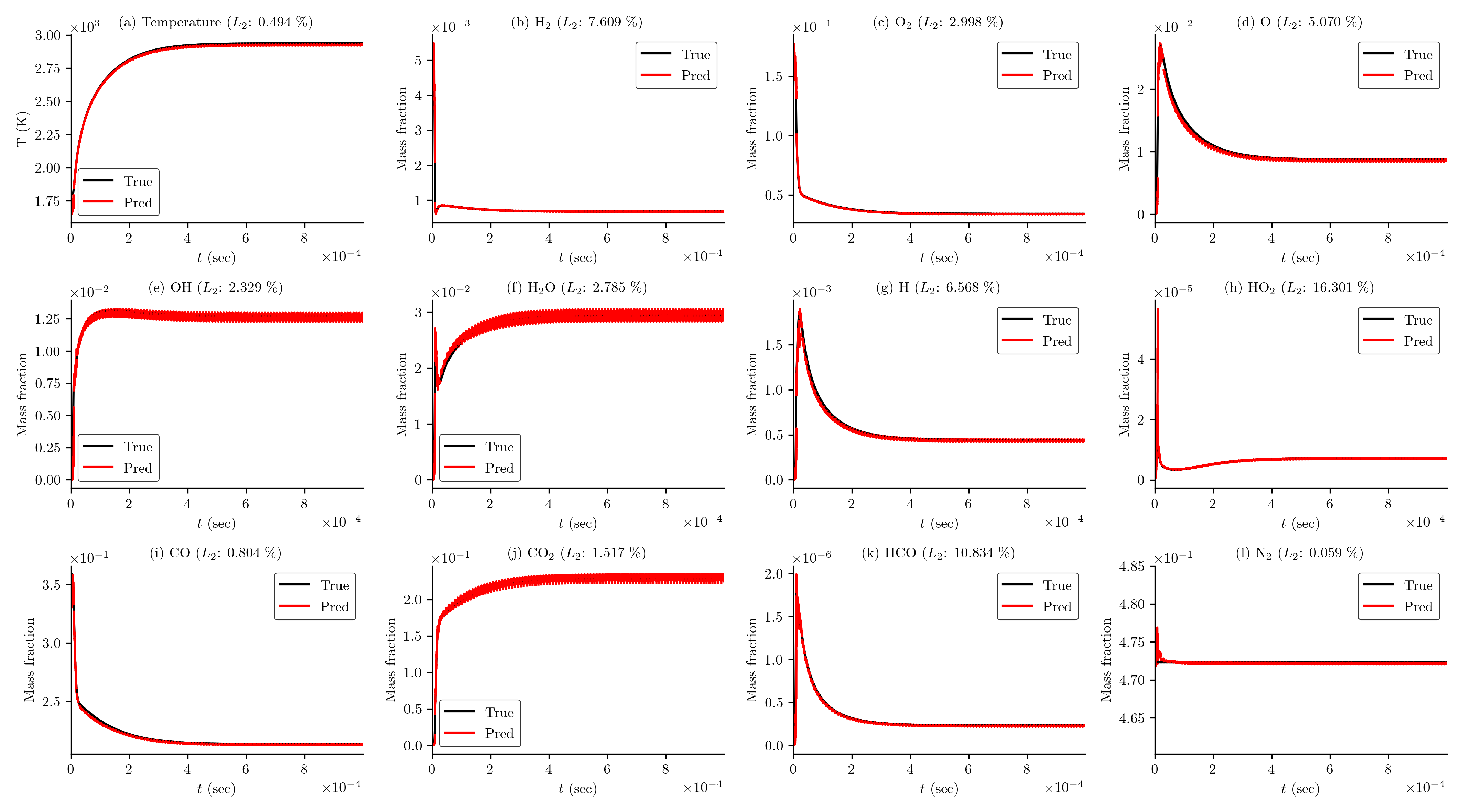}
    \caption{2S-Ad-B, Extrapolation sample \# 2}
    \end{subfigure}
    \caption{\textbf{Syngas problem, DeepONet extrapolation result \# 2.} Plot showing the dynamics of the state variables of the syngas problem for extrapolation dataset \#2 when predicted using (a) DOK-Ad-B and (b) 2S-Ad-B method. The predicted dynamics show good accuracy with the true dynamics.}
    \label{Figure:Syngas:Extrapolation sample 2}
\end{figure}

\begin{figure}[H]
    \centering
    \includegraphics[width=1\textwidth]{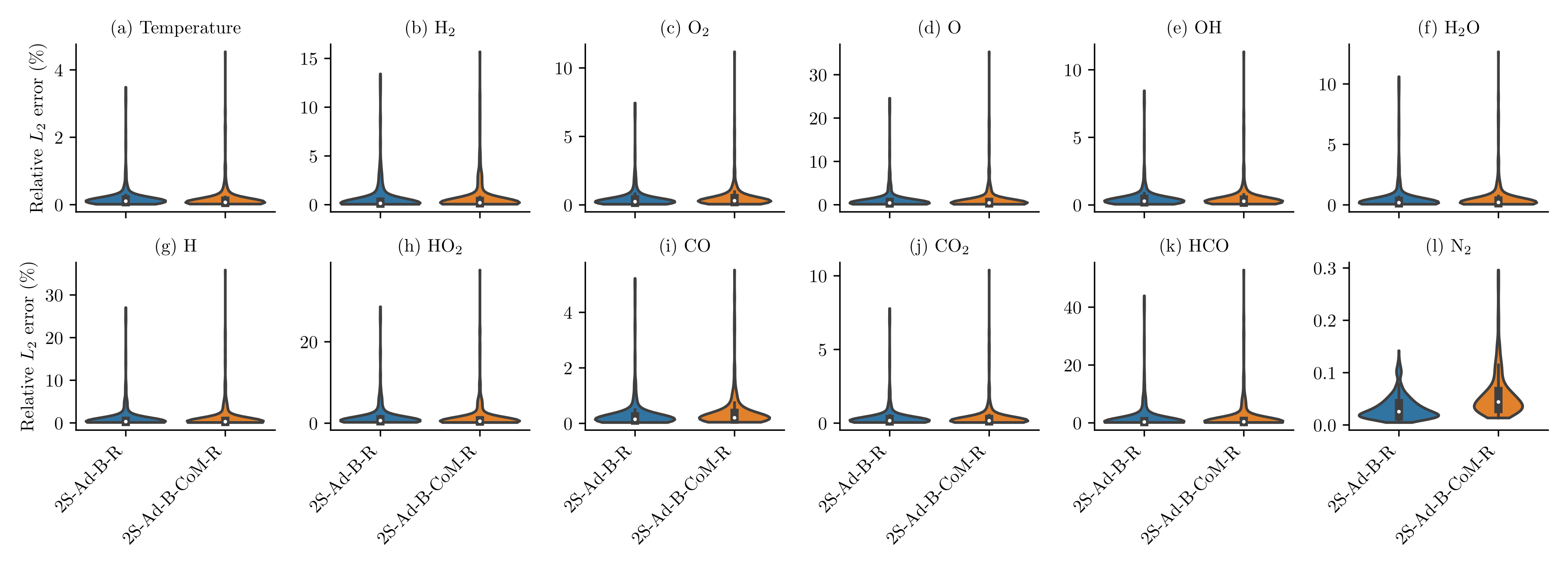}
    \caption{\textbf{Syngas problem, Recursive prediction, Violin plot} for relative $L_2$ error of the reconstructed prediction of the state variables of the test dataset. The outputs are predicted recursively as discussed in \Cref{Subsection:Recursive prediction}. The relative $L_2$ errors are calculated similarly as discussed in \Cref{Subsection:Numerical:Syngas:Result analysis}.}
    \label{Figure:Syngas:DeepONet:Recursive prediction:Violin plot}
\end{figure}

We discussed the recursive predictions in \Cref{Subsubsection:Numerical:Syngas:Recursive prediction}. The relative $L_2$ error (not reconstructed) in recursive prediction over three independent runs for recursive prediction using DeepOKAN trained using Adaptive loss Type-B with two step training method (2S-Ad-B-R) and using mass conserving DeepOKAN trained using Adaptive loss Type-B with two step training method (2S-Ad-B-CoM-R), we obtained relatuve $L_2$ errors listed in \Cref{Table:Syngas:Recursive:Test error}. The relative $L_2$ error for the reconstructed prediction for individual state variables is shown in \Cref{Table:Syngas:DeepONet:Recursive prediction error}, and the violin plots for them are shown in \Cref{Figure:Syngas:DeepONet:Recursive prediction:Violin plot}.

\begin{table}[H]
    \centering
    \caption{\textbf{Syngas problem, testing error for recursive prediction:} Relative $L_2$ error in prediction for test dataset when recursively predicted the output as discussed in \Cref{Subsubsection:Numerical:Syngas:Recursive prediction} with $n_t+1=101$, predicted using DeepOKAN trained using Adaptive loss Type-B with two step training method (2S-Ad-B-R) and using mass conserving DeepOKAN trained using Adaptive loss Type-B with two step training method (2S-Ad-B-CoM-R). The table shows the prediction using the three independently trained models for each DeepONet discussed in the previous sections.}
    \label{Table:Syngas:Recursive:Test error}
    \renewcommand{\arraystretch}{1.25}
    \begin{tabular}{l|C{5cm}} \hline
      \rowcolor{orange!45} Case & Relative $L_2$ error (\%) \\ \hline
      2S-Ad-B-R  & 0.632, 0.509, 0.606\\ \hline
      \rowcolor{cyan!25} 2S-Ad-B-CoM-R & 0.643, 0.565, 0.556 \\ \hline
    \end{tabular}
\end{table}

\begin{figure}[H]
    \centering
    \begin{subfigure}[b]{1\textwidth}
    \includegraphics[width=1\textwidth]{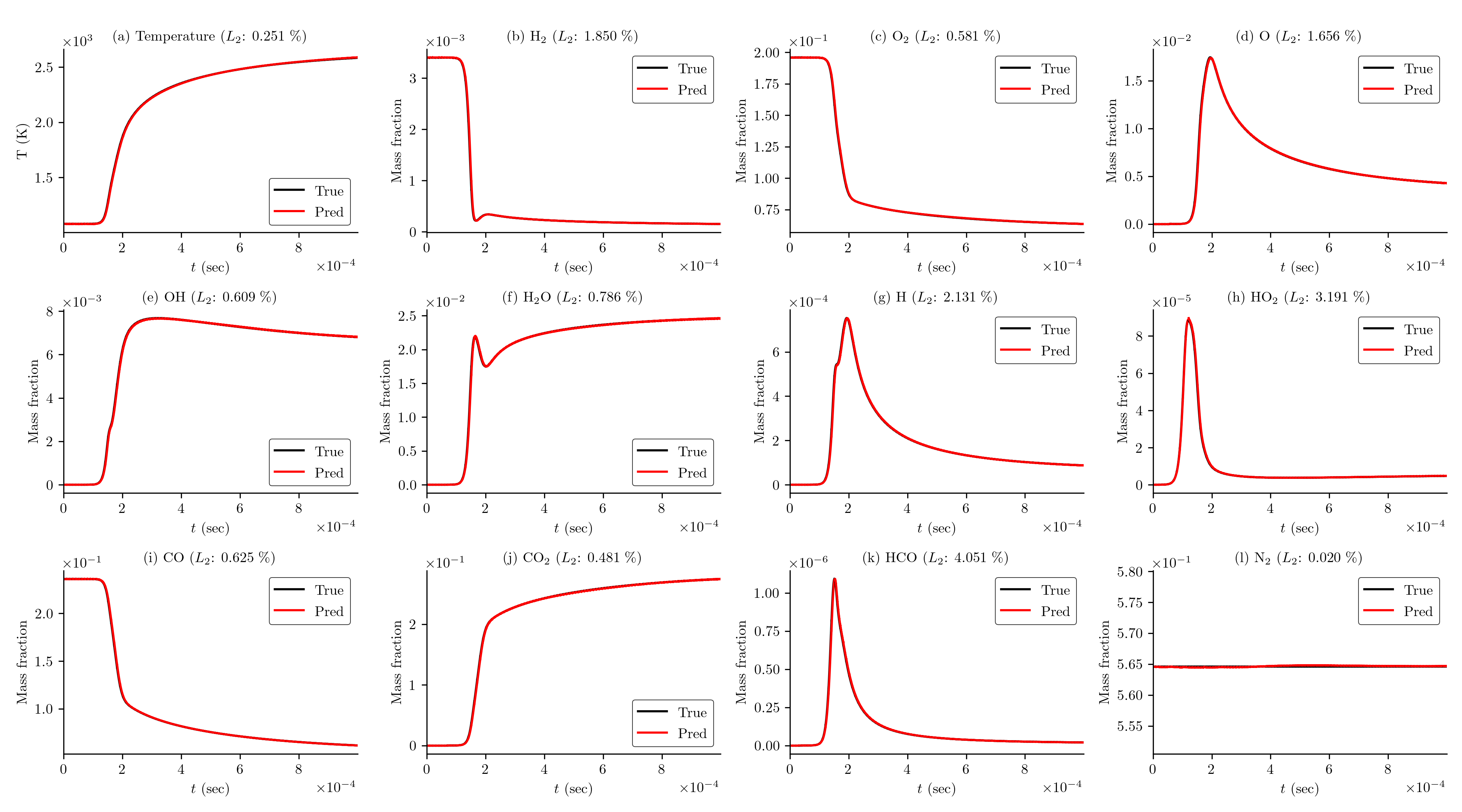}
    \caption{Recursive prediction, Sample result 2}    
    \end{subfigure}
\end{figure}

\begin{figure}[H]\ContinuedFloat
    \begin{subfigure}[b]{1\textwidth}
    \includegraphics[width=1\textwidth]{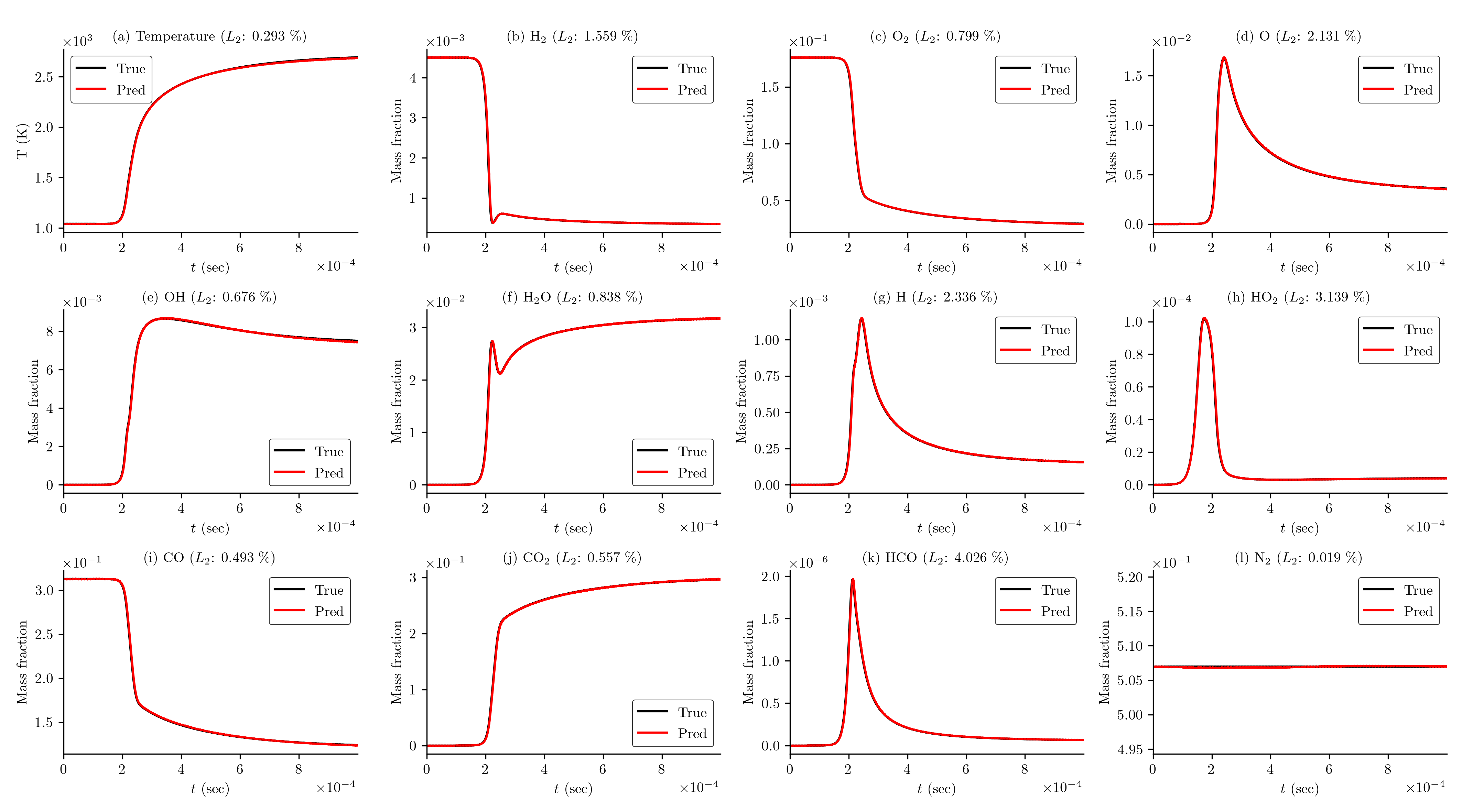}
    \caption{Recursive prediction, Sample result 3}    
    \end{subfigure}
    \caption{\textbf{Syngas problem, Recursive prediction, sample results.} Plot showing the dynamics of the state variables of the syngas problem for test dataset when predicted using recursive prediction. (a) shows the dynamics of the state variables corresponding to 90 percentile error in predicted state variable H$_2$O (0.786\%), (b) shows the dynamics of the state variables corresponding to 90 percentile error in predicted state variable Temperature (0.293\%). The plots show the dynamics of the predicted state variable when predicted recursively using the trained model 2S-Ad-B, which we called 2S-Ad-B-R. Both the plots show good accuracy in prediction.}
    \label{Figure:Syngas:Recursive prediction:Sample results: Appendix}
\end{figure}
\subsubsection{Error Convergence for Syngas problem}
\label{Appendix:Subsubsection:Error Convergence for Syngas problem}
\begin{figure}[H]
    \centering
     \includegraphics[width=1\textwidth]{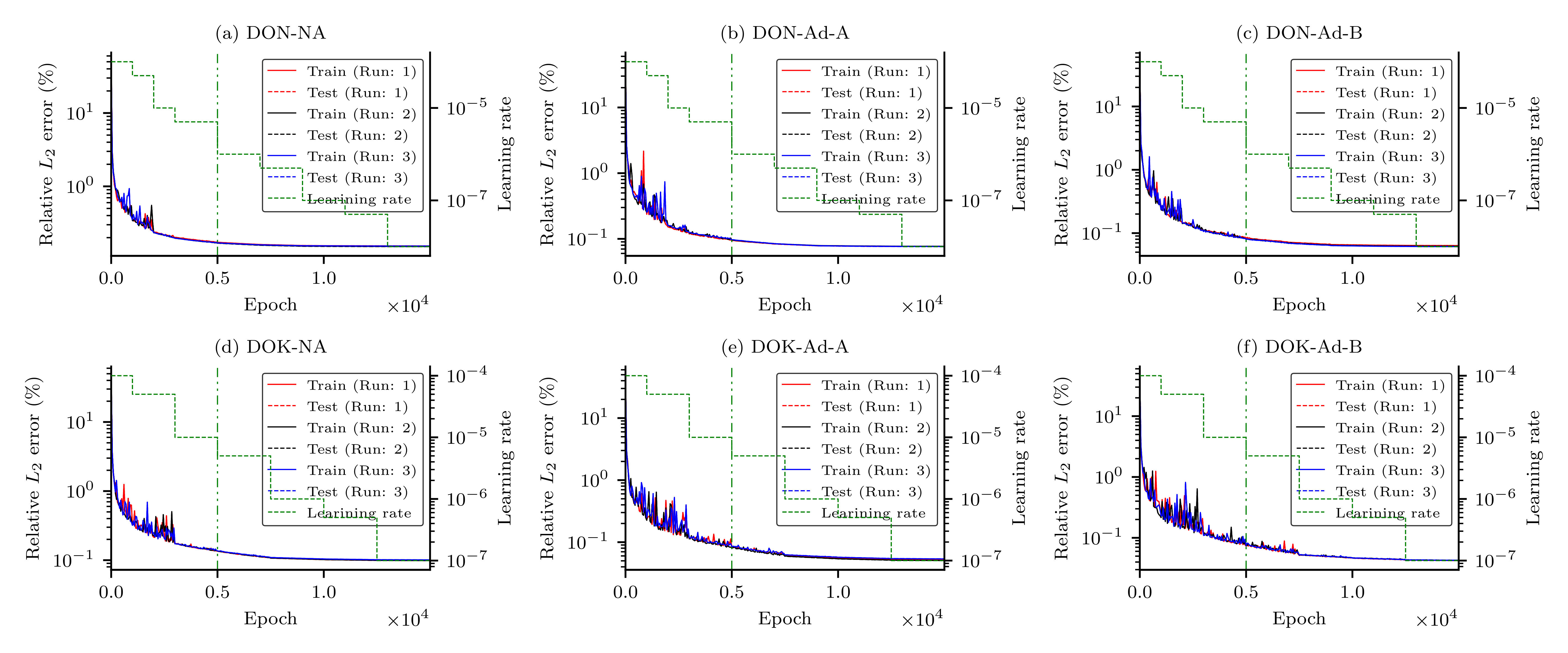}
    \caption{\textbf{Convergence of error for Syngas problem for DeepONet.} Convergence of mean of relative $L_2$ error in training and testing with epoch for different methods of DeepONet and loss functions considered. The plots show the convergence for the three independent runs. The three runs follow a similar convergence path with epochs. We have also shown the learning rate considered for each method. We have considered 60 mini-batch up to 5 k epoch and 120 mini-batch after 5 k epoch and indicated by the vertical line at 5 k epoch.}
    \label{Figure:Syngas:Convergence L2 error one step}
\end{figure}

\begin{figure}[H]
    \centering
    \includegraphics[width=1\textwidth]{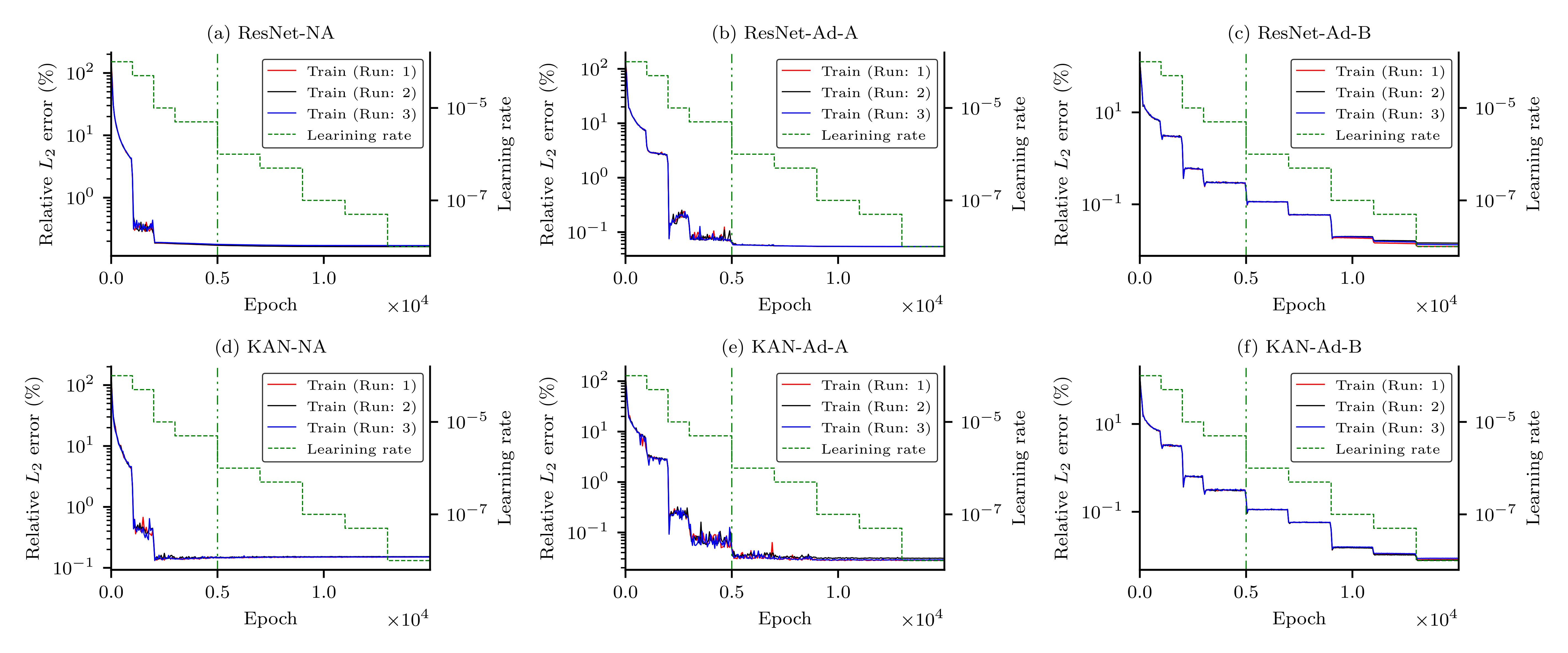}
    \caption{\textbf{Convergence of error for Syngas problem for trunk training when trained using two-step training of DeepONet.} Convergence of mean of relative $L_2$ error in trunk training with epoch for different methods of DeepONet and loss functions considered when trained using two-step training. The plots show the convergence for the three independent runs. The three runs follow a similar convergence path with epochs. We have also shown the learning rate considered for each method. We have considered 60 mini-batch up to 5 k epoch and 120 mini-batch after 5 k epoch and indicated by the vertical line at 5 k epoch.}
    \label{Figure:Syngas:Covergence 2-step training L2 trunk training}
\end{figure}

\begin{figure}[H]
    \centering
    \includegraphics[width=1\textwidth]{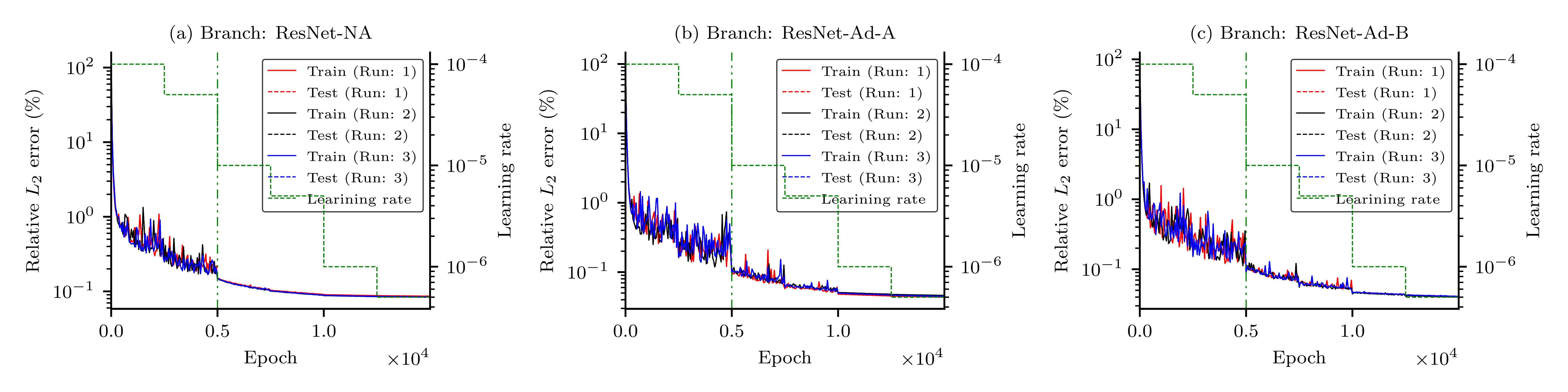}
    \caption{\textbf{Convergence of error for Syngas problem for branch training when trained using two-step training of DeepONet.} Convergence of mean of relative $L_2$ error in branch training for training and testing dataset with epoch for different methods of DeepONet and loss functions considered when trained using two-step training. The plots show the convergence for the three independent runs. The three runs follow a similar convergence path with epochs. We have also shown the learning rate considered for each method. We have considered 60 mini-batches up to 5 k epochs and 120 mini-batches after 5 k epochs, and indicated by the vertical line at 5 k epochs.}
    \label{Figure:Syngas:Convergence:2-step:L2 branch training}
\end{figure}

\begin{figure}[H]
    \centering
    \includegraphics[width=0.975\textwidth]{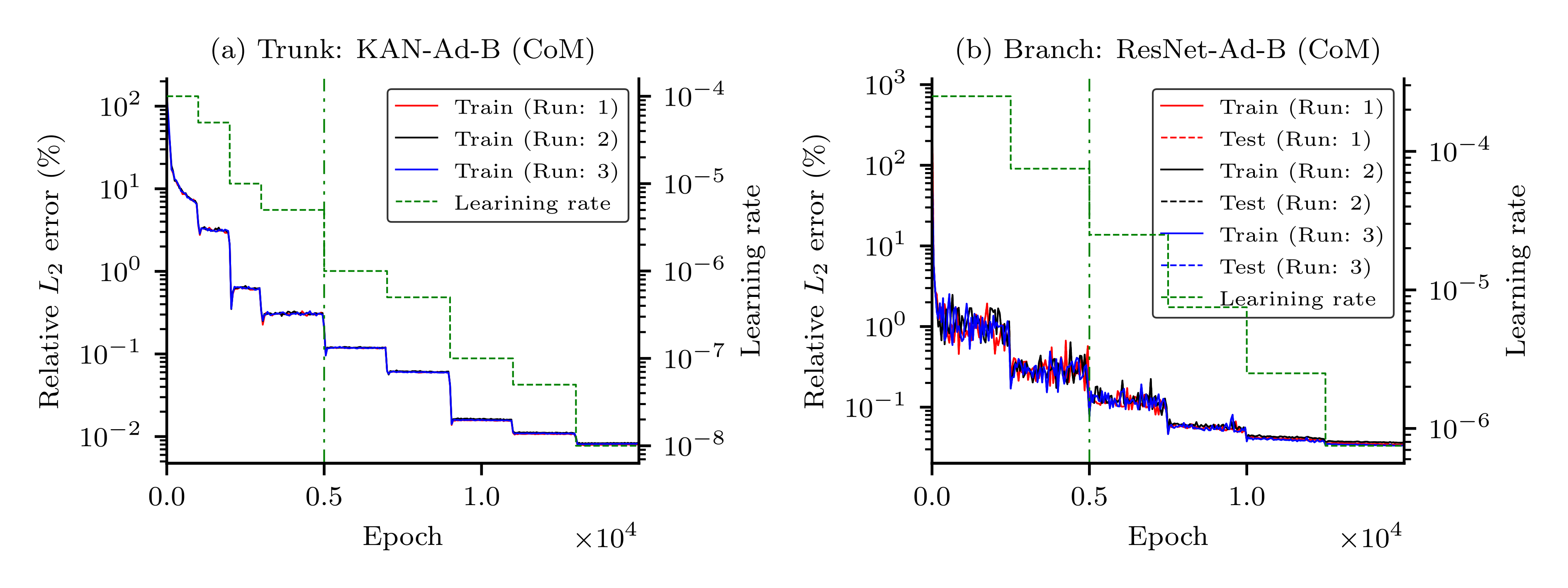}
    \caption{\textbf{Syngas problem, convergence Mass conserving DeepOKAN, 2 step training.} Convergence of error for Syngas problem for trunk training and branch training when trained using two-step training of DeepOKAN with mass conservation (ref \Cref{Subsection:Mass conserving DeepONet}). Convergence of mean of relative $L_2$ error in trunk and branch training with epoch for adaptive loss Type-B. The plots show the convergence for the three independent runs. The three runs follow a similar convergence path with epochs. We have also shown the learning rate considered for each method. We have considered 60 mini-batch up to 5 k epoch and 120 mini-batch after 5 k epoch and indicated by the vertical line at 5 k epoch.}
    \label{Figure:Syngas:Convergence mass conserve DOK}
\end{figure}

In \Cref{Figure:Syngas:Convergence L2 error one step}, we have shown the convergence of relative $L_2$ error (calculated as discussed in \Cref{Subsection:Error matric}) in training and testing for DON and DOK when trained using different loss functions. We have also shown the learning rate considered in the training. Similarly, the convergence of relative $L_2$ error in training for trunk training in the case of two-step training is shown in \Cref{Figure:Syngas:Covergence 2-step training L2 trunk training} and that of training and testing in branch training in the case of two-step training is shown in \Cref{Figure:Syngas:Convergence:2-step:L2 branch training}. These convergence plots show the convergence of relative $L_2$ error (calculated as discussed in \Cref{Subsection:Error matric}) for three independent run for each method. We observed that all the three runs follow similar converge path. Furthermore, the error in the case of adaptive loss functions is smaller than the non adaptive case. However, the non adaptive loss converge marginally faster than the adaptive loss functions. In \Cref{Figure:Syngas:Convergence mass conserve DOK}, we have shown the converge of relative $L_2$ error in the case of two-step method with mass conservation discussed in \Cref{Subsection:Numerical:Syngas:Mass conserving DeepONet}.


\begin{figure}[H]
    \centering
    \includegraphics[width=0.975\textwidth]{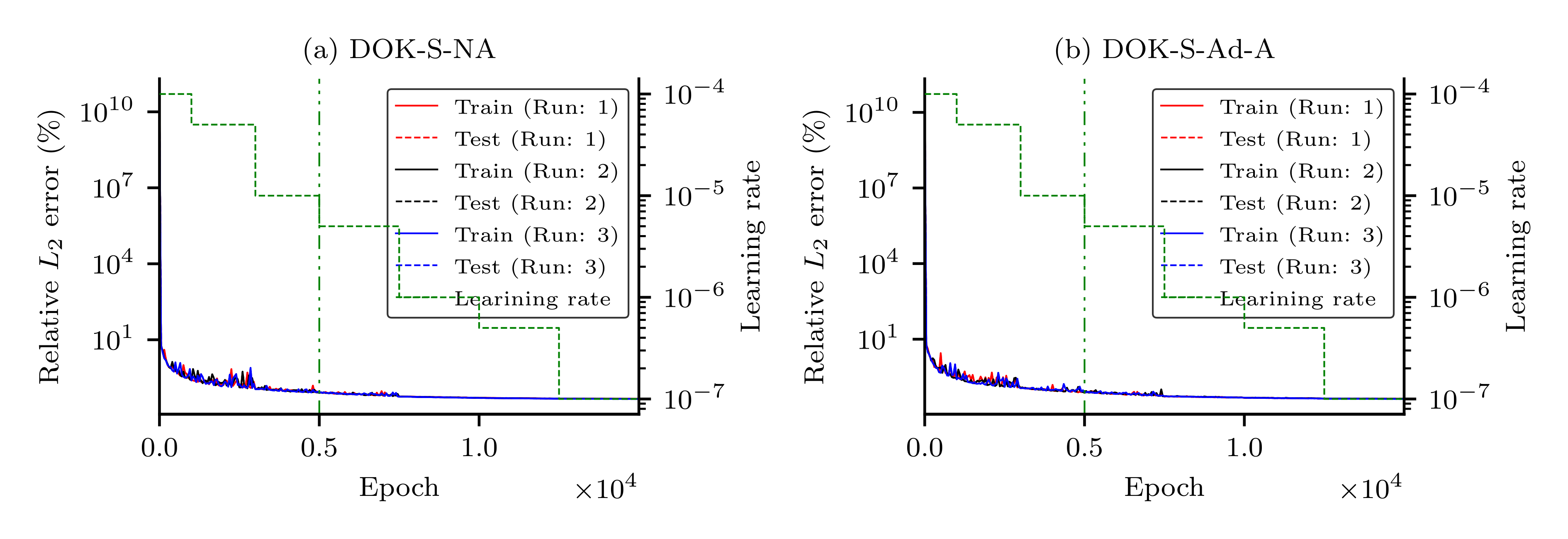}
    \caption{\textbf{Syngas problem, convergence of softmax Mass conserving DeepOKAN.} Convergence of error for Syngas problem when trained using DOK with and without adaptive loss function Type-B with mass conservation using softmax function (ref \Cref{subsection:Numerical:Syngas:Mass conservation using softmax function}). The plots show the convergence of mean relative $L_2$ error for the three independent runs. The three runs follow a similar convergence path with epochs. We have also shown the learning rate considered for each method. We have considered 60 mini-batch up to 5 k epoch and 120 mini-batch after 5 k epoch and indicated by the vertical line at 5 k epoch.}
 \label{Figure:Syngas:Softmax_mass_Conseving_Convergence}
\end{figure}
 
\subsection{GRI-Mech 3.0 problem}
\label{Appendix:Subsection:GRI problems}
\begin{figure}[H]
    \centering
    \includegraphics[width=0.975\textwidth]{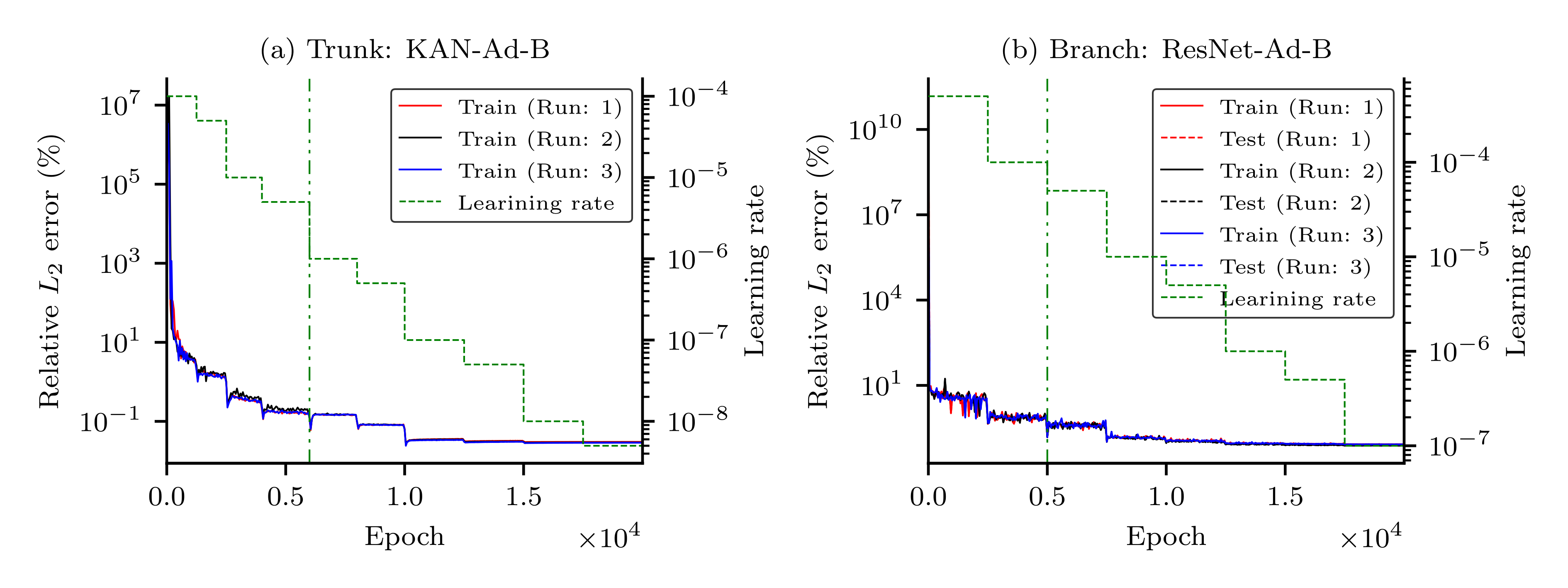}
    \caption{\textbf{GRI problem, convergence of DeepOKAN.} Convergence of error for GRI problem for trunk training and branch training when trained using two-step training of DeepOKAN. Convergence of mean of relative $L_2$ error in trunk and branch training with epoch for adaptive loss Type-B. The plots show the convergence for the three independent runs. The three runs follow a similar convergence path with epochs. We have also shown the learning rate considered for each method. We have considered 20 mini-batch up to 6 k epoch and 40 mini-batch after 6 k epoch during the trunk training and indicated by the vertical line at 6 k epoch. Similarly, We have considered 60 mini-batch up to 5 k epoch and 120 mini-batch after 5 k epoch during the branch training and indicated by the vertical line at 5 k epoch}
    \label{Figure:GRI:Convergence DOK}
\end{figure}

\begin{figure}[H]
    \centering
    \begin{subfigure}[b]{1\textwidth}
    \includegraphics[width=1\textwidth]{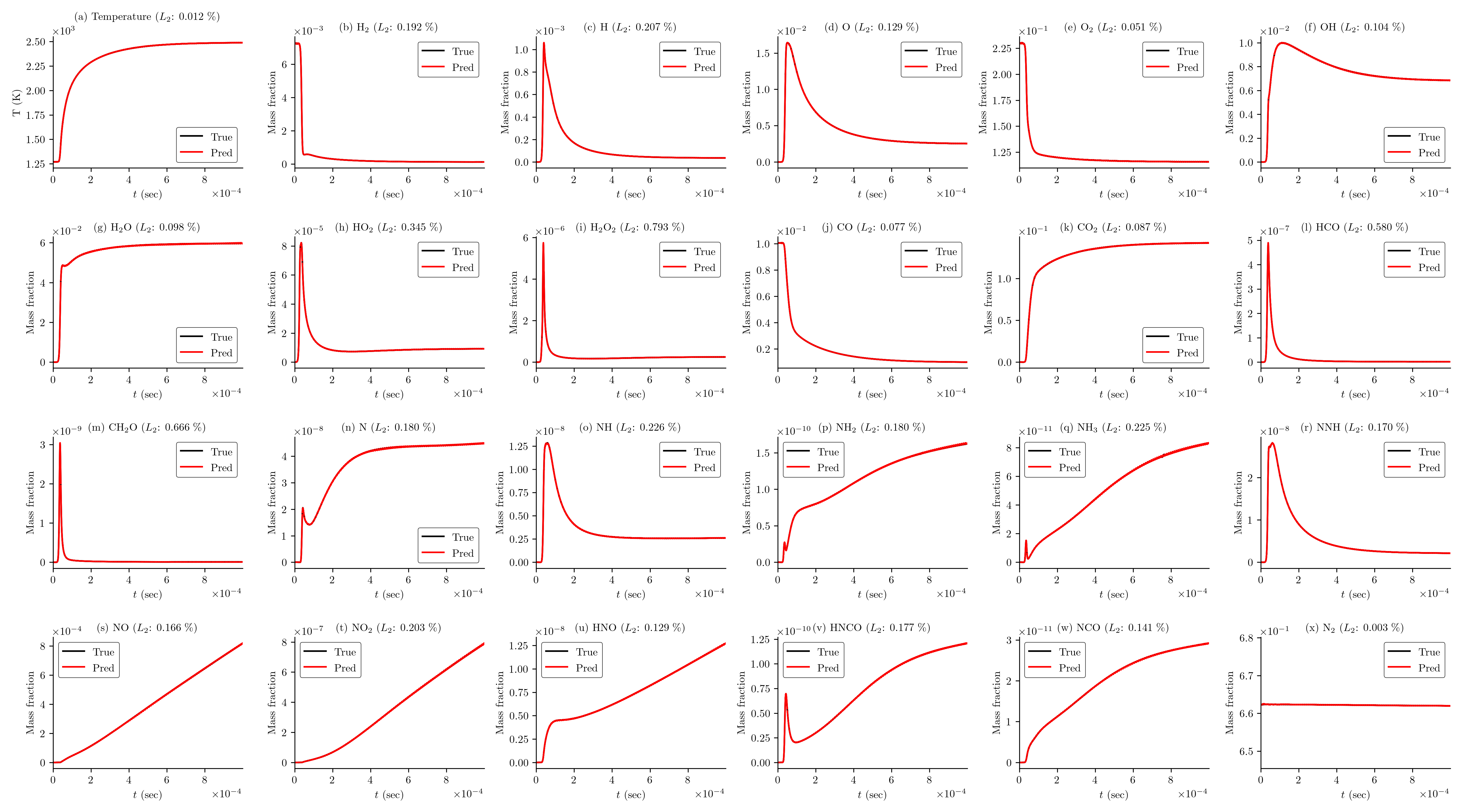}
    \caption{Sample 2}
    \end{subfigure}
    \begin{subfigure}[b]{1\textwidth}
    \includegraphics[width=1\textwidth]{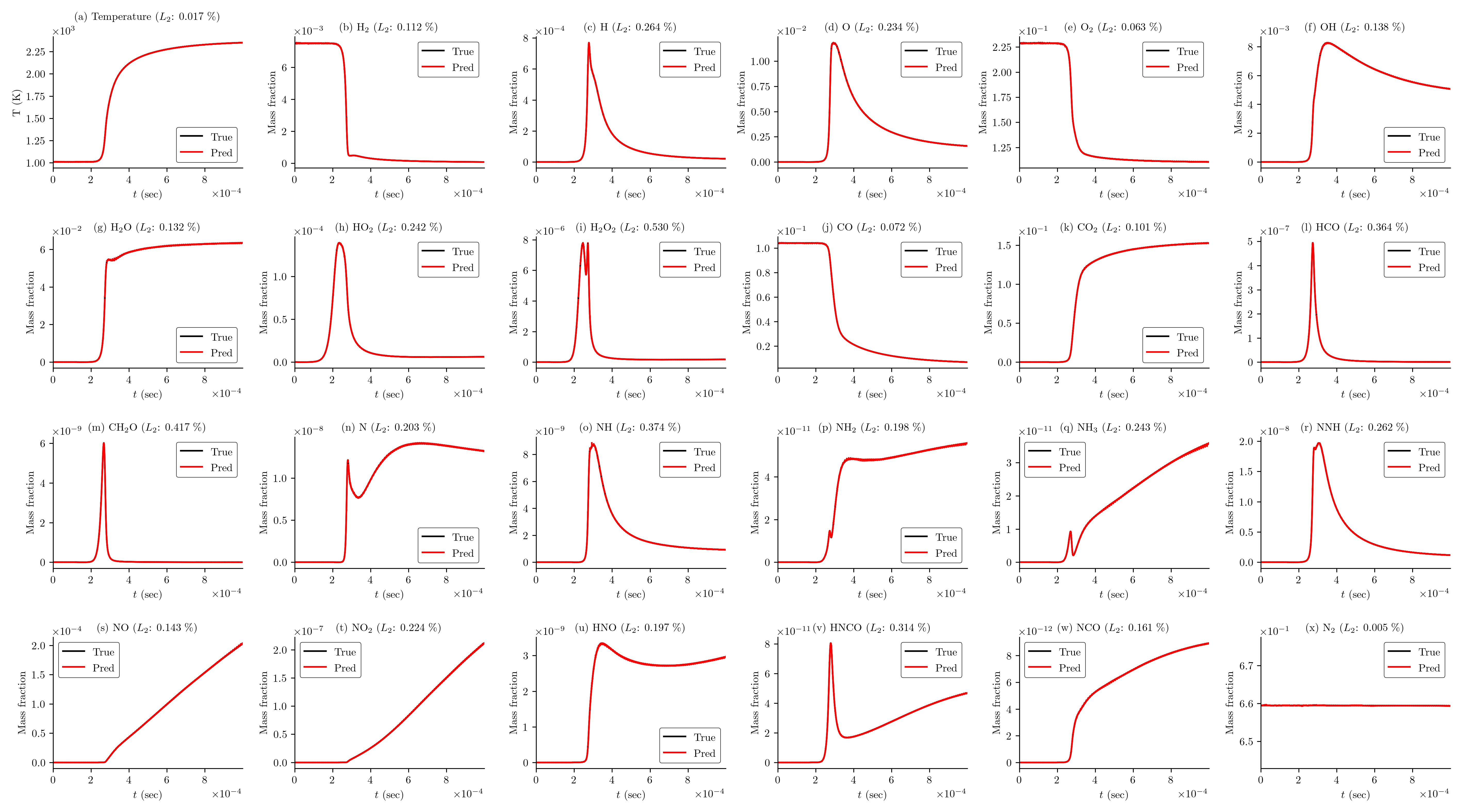}
    \caption{Sample 3}
    \end{subfigure}
    \caption{\textbf{GRI problem, sample result from test dataset.} Plot showing the two sample predicted dynamics of the state variables for GRI problem. The sample result corresponds to one of the higher errors in prediction. The predicted sample shows good agreement with the true value.}
    \label{Figure:GRI:Sample result:Appendix}
\end{figure}

We have discussed the results for the GRI-Mech 3.0 problem in \Cref{Section:Numerical results:GRI Problem}. In this section, we will present additional results for the GRI-Mech 3.0 problem not included in \Cref{Section:Numerical results:GRI Problem}. In \Cref{Table:GRI problem:Min max training-testing data}, we have shown the minimum and maximum values of the training and testing dataset of the state variables of the GRI-Mech 3.0 problem. The sum of the species mass fraction (23 species) at any time instant lies between $1.0000002$ and $0.9999988$ for the training dataset and $1.000000013$ and $0.9999988$ for the testing dataset. In \Cref{Table:GRI: network size}, we have shown the network detail for DOK, including activation function, layers, number of parameters, etc. In \Cref{Figure:GRI:Convergence DOK}, we show the convergence of relative $L_2$ error for trunk and branch training using two-step training of DeepOKAN. Predicted samples for the GRI Mech 3.0 problem show good agreement with the true solution as given in \Cref{Figure:GRI:Sample result:Appendix}.

\begin{table}[H]
    \centering
    \caption{\textbf{GRI Mech 3.0 problem, Minimum and maximum value} of the training and testing dataset for the GRI problem for the state variables (temperature and each of the $23$ species).}
    \label{Table:GRI problem:Min max training-testing data}
    \begin{tabular}{c|c|c|c|c}
    \hline
    \rowcolor{orange!45} & \multicolumn{2}{c|}{Training dataset} & \multicolumn{2}{c}{Testing dataset} \\ \cline{2-5} 
    \rowcolor{orange!45} \multirow{-2}{*}{State variable} & Minimum Value & Maximum Value & Minimum Value & Maximum Value \\\hline
    Temperature & 1.000e+03 & 2.870e+03 & 1.000e+03 & 2.867e+03 \\ \hline
    \rowcolor{cyan!25} H$_2$ & 7.214e-05 & 1.787e-02 & 7.763e-05 & 1.787e-02 \\ \hline
    H & 5.110e-14 & 3.873e-03 & 5.961e-14 & 3.864e-03 \\ \hline
    \rowcolor{cyan!25} O & 1.160e-12 & 2.356e-02 & 1.334e-12 & 2.355e-02 \\ \hline
    O$_2$ & 7.253e-03 & 2.298e-01 & 7.791e-03 & 2.298e-01 \\ \hline
    \rowcolor{cyan!25} OH & 3.626e-14 & 1.921e-02 & 4.513e-14 & 1.921e-02 \\ \hline
    H$_2$O & 1.104e-14 & 1.248e-01 & 1.614e-14 & 1.241e-01 \\ \hline
    \rowcolor{cyan!25} HO$_2$ & 1.681e-12 & 1.428e-04 & 1.943e-12 & 1.428e-04 \\ \hline
    H$_2$O$_2$ & 1.595e-17 & 9.155e-06 & 2.187e-17 & 8.794e-06 \\ \hline
    \rowcolor{cyan!25} CO & 5.820e-03 & 2.482e-01 & 6.318e-03 & 2.482e-01 \\ \hline
    CO$_2$ & 3.257e-12 & 1.867e-01 & 3.767e-12 & 1.856e-01 \\ \hline
    \rowcolor{cyan!25} HCO & 5.465e-17 & 2.748e-06 & 7.557e-17 & 2.716e-06 \\ \hline
    CH$_2$O & 9.630e-21 & 4.200e-08 & 1.439e-20 & 4.112e-08 \\ \hline
    \rowcolor{cyan!25} N & 2.698e-27 & 1.705e-06 & 2.975e-27 & 1.665e-06 \\ \hline
    NH & 1.500e-28 & 2.212e-07 & 1.889e-28 & 2.133e-07 \\ \hline
    \rowcolor{cyan!25} NH$_2$ & 3.043e-31 & 4.038e-08 & 4.532e-31 & 3.959e-08 \\ \hline
    NH$_3$ & 5.108e-33 & 2.286e-08 & 9.024e-33 & 2.260e-08 \\ \hline
    \rowcolor{cyan!25} NNH & 5.047e-19 & 1.079e-07 & 5.642e-19 & 1.074e-07 \\ \hline
    NO & 7.873e-27 & 8.422e-03 & 8.644e-27 & 8.260e-03 \\ \hline
    \rowcolor{cyan!25} NO$_2$ & 5.099e-37 & 3.982e-06 & 5.130e-37 & 3.881e-06 \\ \hline
    HNO & 2.777e-30 & 5.054e-07 & 3.333e-30 & 5.049e-07 \\ \hline
    \rowcolor{cyan!25} HNCO & 2.370e-31 & 2.262e-08 & 3.628e-31 & 2.234e-08 \\ \hline
    NCO & 5.533e-31 & 5.498e-09 & 7.369e-31 & 5.351e-09 \\ \hline
    \rowcolor{cyan!25} N$_2$ & 5.421e-01 & 6.623e-01 & 5.422e-01 & 6.623e-01 \\ \hline
    \end{tabular}
\end{table}
\begin{table}[H]
\centering
\caption{\textbf{DeepONet size considered for GRI problem}. We consider two different types of architectures. In DON (DeepONet), ResNet is considered in both the branch and trunk, while in DOK (DeepOKAN), ResNet is considered in the branch, and KAN is considered in the trunk. The ResNet size $[12-200\times 10-2280]$ indicates that there are $12$ neurons in the input layer, $10$ hidden layers with $200$ neurons in each layer, and the output layer contains $2280$ neurons. The KAN $[1-95\times 5 -2280]$ indicates five layers with 95 edges. We consider the Jacobi polynomial for KAN. We discussed the ResNet and KAN architecture in \Cref{Appedix:Network Architecture}. The output $2280$ neurons are divided into $24$ parts, each part with $95$ neurons for each state variable, i.e., $p=95$. In the ``Number of parameters", the second number, in the case of ResNet's, indicates the number of parameters in the projection network required in the first layer due to size mismatch. The ``sin(\:.\:)" in KAN is an additional function and is discussed in \Cref{Appedix:Network Architecture}.}
\label{Table:GRI: network size}
\begin{tabular}{L{1.5cm}|L{4cm}|C{3.75cm}} \hline
\rowcolor{orange!45}  &  & DOK \\ \hline
& Network Type  & ResNet \\
& Network size & $[24-250\times 10 - 2280]$ \\
& Activation function & $\text{tanh}(\:.\:)$ \\
\multirow{-4}{*}{Branch}& Number of parameters & $1143280+6250$\\ \hline
& Network Type & KAN 3\textsuperscript{rd} order Jacobi ($\alpha = 1.0$, $\beta = 1.0$) \\
& Network size & $[1-95\times 5 - 2280]$ \\
& Activation function &  $\sin(\:.\:)$ (*ref \Cref{Appedix:Network Architecture}) \\
\multirow{-4}{*}{Trunk}& Number of parameters  & $1011180$\\ \hline
\multicolumn{2}{l|}{Total parameters} & 2160710 (2.2 M) \\ \hline
\end{tabular}
\end{table}

\begin{table}[H]
\centering
\caption{\textbf{GRI problem, mean and standard deviation} of the \% relative $L_2$ error for the test samples. The state variable describing the dynamics of H$_2$O$_2$ has the maximum mean relative $L_2$ error. The violin plots for the relative $L_2$ error are shown in \Cref{Figure:GRI:Violin plot}. The relative $L_2$ errors are calculated similarly as discussed in \Cref{Subsection:Numerical:Syngas:Result analysis}.}
\label{Table:GRI:Mean and SD for test dataset}
\renewcommand{\arraystretch}{1.25}
\begin{tabular}{C{1.15cm}|C{1cm}|C{1cm}?C{1.15cm}|C{1cm}|C{1.15cm}?C{1.15cm}|C{1cm}|C{1cm}}
\hline
\rowcolor{orange!45}  States & $\mu$ & $\sigma$ & States & $\mu$ & $\sigma$ & States & $\mu$ & $\sigma$\\ \hline
Temp & 0.008 & 0.002 & H$_2$O$_2$ & 0.362 & 0.095 & NH$_3$ & 0.117 & 0.028 \\ \hline
\rowcolor{cyan!25} H$_2$ & 0.100 & 0.024 & CO & 0.037 & 0.014 & NNH & 0.094 & 0.028 \\ \hline
H & 0.121 & 0.034 & CO$_2$ & 0.059 & 0.012 & NO & 0.095 & 0.023 \\ \hline
\rowcolor{cyan!25} O & 0.096 & 0.022 & HCO & 0.209 & 0.064 & NO$_2$ & 0.127 & 0.029 \\ \hline
O$_2$ & 0.057 & 0.019 & CH$_2$O & 0.284 & 0.104 & HNO & 0.099 & 0.025 \\ \hline
\rowcolor{cyan!25} OH & 0.064 & 0.014 & N & 0.101 & 0.023 & HNCO & 0.100 & 0.034 \\ \hline
H$_2$O & 0.068 & 0.014 & NH & 0.102 & 0.045 & NCO & 0.089 & 0.023 \\ \hline
\rowcolor{cyan!25} HO$_2$ & 0.240 & 0.077 & NH$_2$ & 0.096 & 0.026 & N$_2$ & 0.003 & 0.001 \\ \hline
 \end{tabular}
\end{table}
\subsection{Implementation and additional result for AMORE in FNO}
\label{Appendix:Subsection:AMORE in FNO}
\begin{figure}[H]
    \centering
    \includegraphics[width=0.9\textwidth]{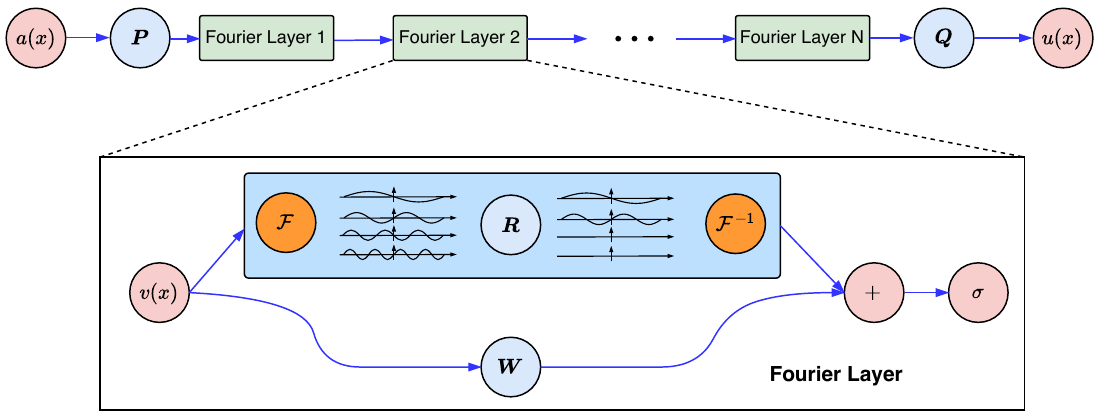}
    \caption{\textbf{Schematic of a FNO.} In the top, we have shown a complete FNO which includes (i) lift to a higher dimensional channel space using a network $\bm{P}$ (e.g., a single layer without an activation function), (ii) a series of Fourier layers and (iii) a projection network $\bm{Q}$ (e.g., a single layer without an activation function). In the bottom, we have shown a detail of a Fourier layer. Inside a Fourier layer, the input function is transomed to frequency domain using FFT, the selective modes are multiply with learnable parameters. The weighted modes are transformed to physical domain using IFFT. This output is added to a weighted (learnable) residual connection. After the residual connection an activation is considered. The trainable parameters are the uplift network $\bm{P}$, the projection network $\bm{Q}$, the weights in the Fourier domain $\bm{R}$ and the residual network $\bm{W}$.}
    \label{Figure:Schematic FNO}
\end{figure}

FNO \citep{Li_2020_FNO} consists of multiple Fourier layers. In \Cref{Figure:Schematic FNO}, we have shown a schematic of a FNO which maps input function to a output function $a(x)\mapsto u(x)$. Each Fourier layers consists of 3 parts: an FFT, which transforms the input function to the frequency domain, and then the selected modes are multiplied by learnable parameters. The second part is an inverse FFT which transforms the weighted modes to physical domain. The third part is a weighted (learnable) residual connection added to the inverse FFT output. After the residual connection an activation is considered. There are also two network at the beginning and end of FNO which uplift and down-lift the input and output respectively.
\par
\begin{figure}[H]
    \centering
    \includegraphics[width=.85\textwidth]{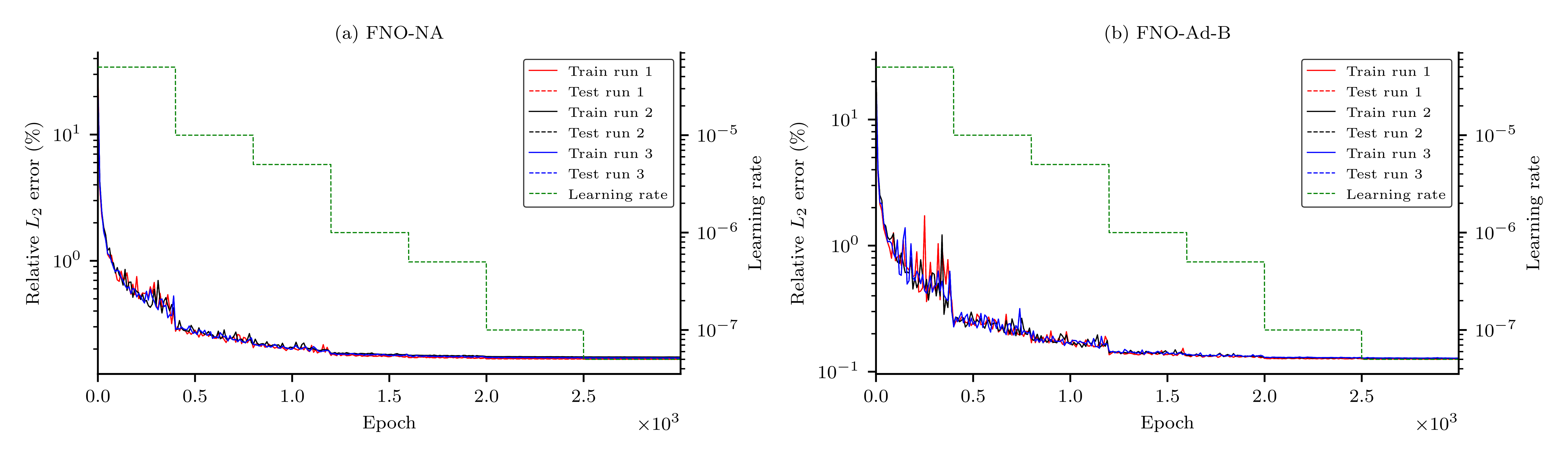}
    \caption{\textbf{Syngas problem, convergence of FNO.} Convergence of mean of relative $L_2$ error in training and testing with epoch for FNO and different loss function (a) FNO-NA (b) FNO-Ad-B. The plots show the convergence over the three independent runs. The three runs follow a similar convergence path with epochs. We have also shown the learning rate for one of the runs. We consider a batch size of 286 (720 mini-batches) for training.}
    \label{Figure:Syngas:FNO:convergence L2 error}
\end{figure}

 Similar to the DeepONet problem, we consider the same time decomposition of ${n_t+1}=101\Delta t$ and map the initial condition of each segment to the time history of 101 time points. Normalization scheme considered for the input initial condition and the output is similar to the one considered for the syngas problem. The time points are normalized to $[-1,1]$ using the minimum and maximum value of $t$ at $t_0$ and $t_{101}$. The time coordinates are concatenated with the feature dimensions thereby making the input size $bs\times {(n_t+1)}\times 13$. The output size $bs\times {(n_t+1)}\times 12$. In the present study, we consider five Fourier layers with 64 hidden dimension and 16 modes retain after the FFT. The gelu activation function is considered for the Fourier layer except for the last Fourier layer. The total number of trainable parameters is $350156$. As discussed in \Cref{Section:AMORE:FNO:Syngas:FNO}, we consider two loss functions, the first one is the non adaptive loss function (FNO-NA) given by \Cref{Eq:Loss:Data_loss} and second one is the adaptive loss function Type-B (FNO-Ad-B) given by \Cref{Eq:Adaptive loss:Type-B}. We also note that in this case, we have not considered the loss due to CoM constraint and the output bound considered in DeepONet given by \Cref{Eq:Output restriction}. We train out FNO model using Adam optimizer in a PyTorch environment with 64-bit data precision. We updated the adaptive weights at every 10 epoch after 10\textsuperscript{th} epoch. The mean relative $L_2$ error in training and testing for three different runs are shown in \Cref{Table:Syngas:FNO:L2 error} and the convergence of error with epoch are shown in \Cref{Figure:Syngas:FNO:convergence L2 error}. Similar to the DeepONet, as discussed in \Cref{Subsection:Numerical:Syngas:Result analysis}, we study the results with the test dataset in more detail. The mean and the standard deviation of the relative $L_2$ error of the reconstructed test dataset for each state variable for both the method for FON-NA and FNO-Ad-B are shown in \Cref{Table:Syngas:FNO:Mean SD}. The violin plots of the relative $L_2$ error for of the state variable for FON-NA and FNO-Ad-B are shown in \Cref{Figure:Syngas:FNO:Violin plot L2 error}. We observed that the 
\begin{table}[H]
\centering
\caption{\textbf{Training and testing error for FNO, Syngas problem.} Relative $L_2$ error in training and testing for FNO when trained with different loss functions. In the second column, w e have indicated the type of loss function considered for each case. The third and fourth  columns show the mean relative $L_2$ error (\%) for training and testing, respectively, for three independent training runs for each case. The convergence of the mean of relative $L_2$ error with epoch is shown in \Cref{Figure:Syngas:FNO:convergence L2 error}. The last column shows the training time for three different runs. We also like to mention that the training time include the calculation of relative $L_2$ error in both norm and physical space at every 10 epoch.}
\label{Table:Syngas:FNO:L2 error}
\begin{tabular}{L{1.75cm}|C{2.75cm}|C{3cm}|C{3cm}|C{3cm}} \hline
\rowcolor{orange!45} & & \multicolumn{2}{c|}{Relative $L_2$ error (\%)} & \\ \cline{3-4}
\rowcolor{orange!45} \multirow{-2}{*}{Case} & \multirow{-2}{*}{Loss function} & Training & Testing & \multirow{-2}{*}{Training time (Hrs)}\\ \hline
FNO-NA & Non-Adaptive & 0.167, 0.173, 0.170 & 0.166, 0.172, 0.169 & 13.72, 13.69, 13.77 \\ \hline
\rowcolor{cyan!25} FNO-Ad-B & Adaptive: Type-B & 0.126, 0.128, 0.128 & 0.125, 0.128, 0.128 & 14.66, 14.8, 14.61 \\ \hline
\end{tabular}
\end{table}

\begin{table}[H]
\centering
\caption{\textbf{Syngas problem, FNO, mean and standard deviation of the \% relative $L_2$ error} for the reconstructed test samples when predicted using FNO, i.e., FNO-NA and FNO-Ad-B. The violin plots of the relative $L_2$ error are shown in \Cref{Figure:Syngas:FNO:Violin plot L2 error}. The relative $L_2$ errors are calculated similarly to those discussed in \Cref{Subsection:Numerical:Syngas:Result analysis}.}
\label{Table:Syngas:FNO:Mean SD}
\resizebox{\textwidth}{!}{%
\renewcommand{\arraystretch}{1.25}
\begin{tabular}{l|c|c|c|c|c|c|c|c|c|c|c|c|c} \hline
\rowcolor{orange!45}Case & & T & H$_2$ & O$_2$ & O & OH & H$_2$O & HO$_2$ & H & CO & CO$_2$ & HCO & N$_2$ \\ \hline
 & $\mu$ & 0.015 & 0.199 & 0.082 & 0.373 & 0.184 & 0.268 & 0.598 & 1.395 & 0.052 & 0.159 & 0.941 & 0.003  \\ \cline{2-14}
\multirow{-2}{*}{FNO-NA} & $\sigma$ & 0.004 & 0.083 & 0.026 & 0.093 & 0.047 & 0.067 & 0.155 & 0.261 & 0.014 & 0.048 & 0.214 & 0.001  \\ \hline
\rowcolor{cyan!25} & $\mu$ & 0.024 & 0.158 & 0.084 & 0.272 & 0.147 & 0.214 & 0.347 & 0.608 & 0.064 & 0.137 & 0.519 & 0.008  \\ \cline{2-14}
\rowcolor{cyan!25} \multirow{-2}{*}{FNO-Ad-B} & $\sigma$ & 0.005 & 0.032 & 0.017 & 0.049 & 0.030 & 0.047 & 0.066 & 0.080 & 0.014 & 0.032 & 0.096 & 0.002  \\ \hline
\end{tabular}}
\end{table}

\setcounter{table}{0}
\setcounter{figure}{0}
\setcounter{equation}{0}
\section{Additional discussion on Computational cost}
\label{Appendix:Section:Numeical:Computational cost}
In the present study, we have considered two different types of DeepONet, i.e., DON, where both branch and trunk are ResNet, and DOK, where the branch is ResNet and the trunk is KAN, and two different paradigms of training methods. In the previous sections, we have discussed the errors for each of the methods for two different problems, i.e., the syngas and the GRI-Mech 3.0. The training time for each of the methods and three independent runs is shown in \Cref{Tabel:Syngas:Training time}. The DeepONets are trained using NVIDIA L40S GPU (\url{https://www.nvidia.com/en-us/data-center/l40s/}) in a network cluster. The variations in computation time are different even for the same method due to network latency and the existing state of the computing and node sharing. The training time for adaptive loss functions is marginally higher. Furthermore, the computational time for the two-step method is higher than the one-step method. In the case of trunk training of the two-step training, we need to train an additional tensor ($\bm{A}$) whose size is too large. Future research may focus on reducing the computational time for the two-step method. We also observed that the training time for the trunk in the case of the GRI problem does not scale with the number of parameters, number of state variables, and number of epochs considered compared to the syngas problem, even though the number of parameters and dataset size are larger. This is because we use a smaller number of mini-batches in the training of the trunk for the GRI problem compared to the syngas problem.
\begin{table}[H]
\centering
\caption{\textbf{Training time for Syngas and GRI problem} for different method considered. We observed that the training time for the adaptive loss functions are marginally higher compared to the non-adaptive loss function. The training time for two-step training is higher than one-step training. We also observed that the training time is marginally different even for the same method. This may be due to the network latency and the existing state of the computing and node sharing. We trained the DeepONets using NVIDIA L40S GPU using Adam \cite{Kingma_2014adam} optimizer in TensorFlow version 2 environment with 64-bit data precision. We also like to mention that we have calculated the adaptive weights in norm and physical space using mini-batch and numpy after every 50 epochs. Similarly, we have calculated the relative $L_2$ after every 50 epochs in norm and physical space using mini-batch and numpy. Since we have calculated $L_2$ error it added approximate 40 minutes time for $L_2$ error calculated.}
\label{Tabel:Syngas:Training time}
\begin{tabular}{L{4.5cm}?L{2.25cm}|C{3.25cm}|C{3.25cm}} \hline
\rowcolor{magenta!90} \multicolumn{4}{c}{\textbf{{\color{white}Syngas problem}}} \\ \hline
\rowcolor{orange!60} \multicolumn{4}{c}{Training time: One-step training} \\ \hline
\rowcolor{orange!30} & Case & Loss function & Time (hrs) \\ \hline
& DON-NA & Non-Adaptive & 10.91, 10.32, 10.86\\ \cline{2-4}
& DON-Ad-A & Adaptive: Type-A & 11.42, 11.37, 11.41 \\ \cline{2-4}
\multirow{-3}{4.25cm}{DeepONet for syngas problem (15 k epochs)} & DON-Ad-B & Adaptive: Type-B & 11.24, 11.05, 11.28 \\ \hline
& DOK-NA & Non-Adaptive & 10.59, 10.56, 10.45 \\ \cline{2-4}
& DOK-Ad-A & Adaptive: Type-A & 11.31, 11.71, 11.80 \\ \cline{2-4}
\multirow{-3}{4.25cm}{DeepOKAN for syngas problem (15 k epochs)} & DOK-Ad-B & Adaptive: Type-B & 11.72, 11.71, 11.55 \\ \hline
\rowcolor{orange!60} \multicolumn{4}{c}{Training time: Two-step training} \\ \hline
\rowcolor{orange!60} \multicolumn{4}{c}{Training time: trunk training} \\ \hline
\rowcolor{orange!30} & Case & Loss function & Time (hrs) \\ \hline
& ResNet-NA  & Non-Adaptive & 19.28, 19.33, 19.36 \\ \cline{2-4}
& ResNet-Ad-A & Adaptive: Type-A & 19.83, 19.88, 20.25 \\ \cline{2-4}
\multirow{-3}{4.25cm}{Trunk (ResNet) training for syngas problem (15 k epochs)}  & ResNet-Ad-B & Adaptive: Type-B &  20.58, 20.70, 20.82 \\ \hline
& KAN-NA & Non-Adaptive & 19.14, 19.54, 19.13 \\ \cline{2-4}
& KAN-Ad-A & Adaptive: Type-A & 19.68, 19.69, 20.49 \\ \cline{2-4}
\multirow{-3}{4.25cm}{Trunk (KAN) training for syngas problem (15 k epochs)} & KAN-Ad-B & Adaptive: Type-B & 19.61, 19.89, 19.71 \\ \hline
\rowcolor{orange!60} \multicolumn{4}{c}{Training time: branch training} \\ \hline
\rowcolor{orange!30} & Case & Loss function & Time (hrs) \\ \hline
& 2S-NA & Non-adaptive & 9.34, 9.84, 9.89 \\ \cline{2-4}
& 2S-Ad-A & Adaptive: Type-A & 9.90, 10.34, 10.38 \\ \cline{2-4}
\multirow{-3}{4.25cm}{Branch (ResNet) training for syngas problem (15 k epochs)} & 2S-Ad-B & Adaptive: Type-B & 10.32, 9.89, 10.30 \\ \hline
\rowcolor{magenta!90} \multicolumn{4}{c}{\textbf{{\color{white}GRI-Mech 3.0 problem}}} \\ \hline
\rowcolor{orange!60} \multicolumn{4}{c}{Training time: Two-step training} \\ \hline
\rowcolor{orange!60} \multicolumn{4}{c}{Training time: trunk training} \\ \hline
\rowcolor{orange!30} & Case & Loss function & Time (hrs) \\ \hline
Trunk (KAN) training for GRI problem (20 k epochs) & KAN-Ad-B & Adaptive: Type-B & 23.19, 23.25, 23.42  \\ \hline
\rowcolor{orange!60} \multicolumn{4}{c}{Training time: branch training} \\ \hline
\rowcolor{orange!45} & Case & Loss function & Time (hrs) \\ \hline
Branch (ResNet) training for syngas problem (20 k epochs) & 2S-Ad-B & Adaptive: Type-B & 25.78, 25.36, 25.47 \\ \hline
\end{tabular}
\end{table}

\end{document}

%% file: Header_setting.tex
\usepackage[utf8]{inputenc} 
\usepackage[T1]{fontenc}
\usepackage[english]{babel}

\usepackage{amsmath,amssymb,amsfonts} 
\usepackage{graphicx}                 
\usepackage{newtxtext,newtxmath}      
\usepackage{bm}
\usepackage[table]{xcolor}            
\usepackage[skip=2pt,font=small]{caption}
\usepackage{subcaption}
\DeclareCaptionFont{Cfont}{\fontsize{10}{12}\selectfont} 
\usepackage[colorlinks]{hyperref}
\hypersetup{linkcolor=blue, citecolor=blue, urlcolor=blue}
\usepackage[capitalise]{cleveref}     
\crefname{figure}{fig.}{figs.} 
\Crefname{figure}{Fig.}{Figs.} 
\Crefname{equation}{Eq.}{Eqs.} 
\makeatletter
\renewcommand{\fnum@figure}{Fig. \thefigure}
\makeatother

\usepackage[numbers,compress,square,sort&compress]{natbib}
\usepackage{booktabs}

\usepackage{enumitem}
\usepackage{authblk}        

\usepackage{wrapfig}
\usepackage{lipsum}
\usepackage{fancyhdr}
\usepackage{chemformula}
\usepackage{afterpage}
\usepackage{lscape}
\usepackage{soul}                      
\usepackage{tikz}
\usetikzlibrary{positioning,shapes.geometric,arrows,arrows.meta}
\usepackage{setspace}
\usepackage{bm}
\usepackage{listings}
\usepackage[parfill]{parskip}
\renewcommand{\headrulewidth}{.4pt}
\usepackage{multirow}
\usepackage{array}
\usepackage{longtable}
\usepackage{tabulary}
\usepackage{makecell}
\usepackage{float}
\newcolumntype{L}[1]{>{\raggedright\let\newline\\\arraybackslash\hspace{0pt}}m{#1}}
\newcolumntype{C}[1]{>{\centering\let\newline\\\arraybackslash\hspace{0pt}}m{#1}}
\newcolumntype{R}[1]{>{\raggedleft\let\newline\\\arraybackslash\hspace{0pt}}m{#1}}
\newcolumntype{?}{!{\vrule width 1.5pt}} 

\usepackage{pdfpages}

%% file: main.bbl
\begin{thebibliography}{58}
\providecommand{\natexlab}[1]{#1}
\providecommand{\url}[1]{\texttt{#1}}
\expandafter\ifx\csname urlstyle\endcsname\relax
  \providecommand{\doi}[1]{doi: #1}\else
  \providecommand{\doi}{doi: \begingroup \urlstyle{rm}\Url}\fi

\bibitem[Maas and Pope(1992)]{MP92}
U.~Maas and S.~B. Pope.
\newblock Simplifying chemical kinetics: Intrinsic low-dimensional manifolds in composition space.
\newblock \emph{Combustion and Flame}, 88\penalty0 (3):\penalty0 239--264, 1992.
\newblock \doi{https://doi.org/10.1016/0010-2180(92)90034-M}.
\newblock URL \url{https://www.sciencedirect.com/science/article/pii/001021809290034M}.

\bibitem[Bykov and Maas(2007)]{Bykov_2007aa}
V.~Bykov and U.~Maas.
\newblock The extension of the ildm concept to reaction--diffusion manifolds.
\newblock \emph{Combustion Theory and Modelling}, 11\penalty0 (6):\penalty0 839--862, 11 2007.
\newblock \doi{10.1080/13647830701242531}.
\newblock URL \url{https://doi.org/10.1080/13647830701242531}.

\bibitem[Lam and Goussis(1994)]{LG94}
S.~H. Lam and D.~A. Goussis.
\newblock The csp method for simplifying kinetics.
\newblock \emph{International Journal of Chemical Kinetics}, 26\penalty0 (4):\penalty0 461--486, 2021/06/10 1994.
\newblock \doi{https://doi.org/10.1002/kin.550260408}.
\newblock URL \url{https://doi.org/10.1002/kin.550260408}.

\bibitem[Jung et~al.(2025)Jung, Lacey, Babaee, and Chen]{JLBC25}
K.~S. Jung, C.~E. Lacey, H.~Babaee, and J.~H. Chen.
\newblock Accelerating high-fidelity simulations of chemically reacting flows using reduced-order modeling with time-dependent bases.
\newblock \emph{Computer Methods in Applied Mechanics and Engineering}, 437:\penalty0 117758, 2025.
\newblock \doi{https://doi.org/10.1016/j.cma.2025.117758}.
\newblock URL \url{https://www.sciencedirect.com/science/article/pii/S0045782525000301}.

\bibitem[Sharma et~al.(2020)Sharma, Johnson, Kessler, and Moses]{Sharma_2020_Deep}
Alisha~J Sharma, Ryan~F Johnson, David~A Kessler, and Adam Moses.
\newblock Deep learning for scalable chemical kinetics.
\newblock In \emph{AIAA scitech 2020 forum}, page 0181, 2020.

\bibitem[Brown et~al.(2021)Brown, Antil, L{\"o}hner, Togashi, and Verma]{Brown_2021_RESDON}
Thomas~S. Brown, Harbir Antil, Rainald L{\"o}hner, Fumiya Togashi, and Deepanshu Verma.
\newblock {Novel DNNs for Stiff ODEs with Applications to Chemically Reacting Flows}.
\newblock In Heike Jagode, Hartwig Anzt, Hatem Ltaief, and Piotr Luszczek, editors, \emph{High Performance Computing}, pages 23--39, Cham, 2021. Springer International Publishing.
\newblock ISBN 978-3-030-90539-2.

\bibitem[Raissi et~al.(2019)Raissi, Perdikaris, and Karniadakis]{Raissi_2019_PINN}
Maziar Raissi, Paris Perdikaris, and George~E Karniadakis.
\newblock Physics-informed neural networks: A deep learning framework for solving forward and inverse problems involving nonlinear partial differential equations.
\newblock \emph{Journal of Computational Physics}, 378:\penalty0 686--707, 2019.

\bibitem[Schiassi et~al.(2021)Schiassi, Furfaro, Leake, De~Florio, Johnston, and Mortari]{Schiassi_2021_Extreme}
Enrico Schiassi, Roberto Furfaro, Carl Leake, Mario De~Florio, Hunter Johnston, and Daniele Mortari.
\newblock Extreme theory of functional connections: A fast physics-informed neural network method for solving ordinary and partial differential equations.
\newblock \emph{Neurocomputing}, 457:\penalty0 334--356, 2021.

\bibitem[De~Florio et~al.(2022)De~Florio, Schiassi, and Furfaro]{Florio_2022_SKPINNXTFC}
Mario De~Florio, Enrico Schiassi, and Roberto Furfaro.
\newblock Physics-informed neural networks and functional interpolation for stiff chemical kinetics.
\newblock \emph{Chaos (Woodbury, N.Y.)}, 32, 06 2022.
\newblock \doi{10.1063/5.0086649}.

\bibitem[Frankel et~al.(2024)Frankel, De~Florio, Schiassi, and Sela]{Frankel_2024_Hybrid}
Matthew Frankel, Mario De~Florio, Enrico Schiassi, and Lina Sela.
\newblock Hybrid chemical and data-driven model for stiff chemical kinetics using a physics-informed neural network.
\newblock \emph{Engineering Proceedings}, 69\penalty0 (1):\penalty0 40, 2024.

\bibitem[Kim et~al.(2021)Kim, Ji, Deng, Ma, and Rackauckas]{Kim_2021_Stiff}
Suyong Kim, Weiqi Ji, Sili Deng, Yingbo Ma, and Christopher Rackauckas.
\newblock Stiff neural ordinary differential equations.
\newblock \emph{Chaos: An Interdisciplinary Journal of Nonlinear Science}, 31\penalty0 (9):\penalty0 093122, 09 2021.
\newblock ISSN 1054-1500.
\newblock \doi{10.1063/5.0060697}.
\newblock URL \url{https://doi.org/10.1063/5.0060697}.

\bibitem[Kumar et~al.(2025)Kumar, Kumar, and Pal]{Kumar_2025_NODE}
Tadbhagya Kumar, Anuj Kumar, and Pinaki Pal.
\newblock A physics-constrained neural ordinary differential equations approach for robust learning of stiff chemical kinetics.
\newblock \emph{Combustion Theory and Modelling}, 29\penalty0 (3):\penalty0 332--347, 2025.
\newblock \doi{10.1080/13647830.2025.2478266}.
\newblock URL \url{https://doi.org/10.1080/13647830.2025.2478266}.

\bibitem[Ji et~al.(2021)Ji, Qiu, Shi, Pan, and Deng]{Ji_2021_Stiff}
Weiqi Ji, Weilun Qiu, Zhiyu Shi, Shaowu Pan, and Sili Deng.
\newblock {Stiff-PINN: Physics-Informed Neural Network for Stiff Chemical Kinetics}.
\newblock \emph{The Journal of Physical Chemistry A}, 125\penalty0 (36):\penalty0 8098--8106, 2021.
\newblock \doi{10.1021/acs.jpca.1c05102}.
\newblock URL \url{https://doi.org/10.1021/acs.jpca.1c05102}.
\newblock PMID: 34463510.

\bibitem[Ayed et~al.(2019)Ayed, de~Bézenac, Pajot, Brajard, and Gallinari]{Ayed_2019_Dynamics}
Ibrahim Ayed, Emmanuel de~Bézenac, Arthur Pajot, Julien Brajard, and Patrick Gallinari.
\newblock {Learning Dynamical Systems from Partial Observations}.
\newblock 2019.
\newblock URL \url{https://arxiv.org/abs/1902.11136}.

\bibitem[Kostic et~al.(2022)Kostic, Novelli, Maurer, Ciliberto, Rosasco, and Pontil]{Kostic_2022_Koopman}
Vladimir Kostic, Pietro Novelli, Andreas Maurer, Carlo Ciliberto, Lorenzo Rosasco, and Massimiliano Pontil.
\newblock {Learning Dynamical Systems via Koopman Operator Regression in Reproducing Kernel Hilbert Spaces}.
\newblock 35:\penalty0 4017--4031, 2022.
\newblock URL \url{https://proceedings.neurips.cc/paper_files/paper/2022/file/196c4e02b7464c554f0f5646af5d502e-Paper-Conference.pdf}.

\bibitem[Lu et~al.(2021)Lu, Jin, Pang, Zhang, and Karniadakis]{Lu_2021_DeepONet}
Lu~Lu, Pengzhan Jin, Guofei Pang, Zhongqiang Zhang, and George~Em Karniadakis.
\newblock {Learning nonlinear operators via DeepONet based on the universal approximation theorem of operators}.
\newblock \emph{Nature Machine Intelligence}, 3\penalty0 (3):\penalty0 218–229, mar 2021.
\newblock ISSN 2522-5839.
\newblock \doi{10.1038/s42256-021-00302-5}.
\newblock URL \url{http://dx.doi.org/10.1038/s42256-021-00302-5}.

\bibitem[Li et~al.(2020)Li, Kovachki, Azizzadenesheli, Liu, Bhattacharya, Stuart, and Anandkumar]{Li_2020_FNO}
Zongyi Li, Nikola Kovachki, Kamyar Azizzadenesheli, Burigede Liu, Kaushik Bhattacharya, Andrew Stuart, and Anima Anandkumar.
\newblock {Fourier neural operator for parametric partial differential equations}.
\newblock \emph{arXiv preprint arXiv:2010.08895}, 2020.

\bibitem[Tripura and Chakraborty(2023)]{Tripura_2023_WNO}
Tapas Tripura and Souvik Chakraborty.
\newblock Wavelet neural operator for solving parametric partial differential equations in computational mechanics problems.
\newblock \emph{Computer Methods in Applied Mechanics and Engineering}, 404:\penalty0 115783, 2023.
\newblock ISSN 0045-7825.
\newblock \doi{https://doi.org/10.1016/j.cma.2022.115783}.
\newblock URL \url{https://www.sciencedirect.com/science/article/pii/S0045782522007393}.

\bibitem[Kovachki et~al.(2023)Kovachki, Li, Liu, Azizzadenesheli, Bhattacharya, Stuart, and Anandkumar]{Kovachki_2023_Neural}
Nikola Kovachki, Zongyi Li, Burigede Liu, Kamyar Azizzadenesheli, Kaushik Bhattacharya, Andrew Stuart, and Anima Anandkumar.
\newblock {Neural operator: Learning maps between function spaces with applications to pdes}.
\newblock \emph{Journal of Machine Learning Research}, 24\penalty0 (89):\penalty0 1--97, 2023.

\bibitem[Centofanti et~al.(2024)Centofanti, Ghiotto, and Pavarino]{centofanti_2024_HHM}
Edoardo Centofanti, Massimiliano Ghiotto, and Luca~F. Pavarino.
\newblock Learning the hodgkin–huxley model with operator learning techniques.
\newblock \emph{Computer Methods in Applied Mechanics and Engineering}, 432:\penalty0 117381, 2024.
\newblock ISSN 0045-7825.
\newblock \doi{https://doi.org/10.1016/j.cma.2024.117381}.
\newblock URL \url{https://www.sciencedirect.com/science/article/pii/S0045782524006364}.

\bibitem[Pellegrini et~al.(2025)Pellegrini, Ghiotto, Centofanti, and Pavarino]{pellegrini_2025_HDIModel}
Luca Pellegrini, Massimiliano Ghiotto, Edoardo Centofanti, and Luca~F Pavarino.
\newblock {Learning High-dimensional Ionic Model Dynamics Using Fourier Neural Operators}.
\newblock \emph{Machine Learning for Computational Science and Engineering}, 1\penalty0 (2):\penalty0 46, 2025.

\bibitem[Venturi and Casey(2023)]{Venturi_2022_SVDPF}
Simone Venturi and Tiernan Casey.
\newblock {SVD perspectives for augmenting DeepONet flexibility and interpretability}.
\newblock \emph{Computer Methods in Applied Mechanics and Engineering}, 403:\penalty0 115718, 2023.
\newblock ISSN 0045-7825.
\newblock \doi{https://doi.org/10.1016/j.cma.2022.115718}.
\newblock URL \url{https://www.sciencedirect.com/science/article/pii/S0045782522006739}.

\bibitem[Goswami et~al.(2024)Goswami, Jagtap, Babaee, Susi, and Karniadakis]{Goswami_2024_Stiff_Kinetics}
Somdatta Goswami, Ameya~D. Jagtap, Hessam Babaee, Bryan~T. Susi, and George~Em Karniadakis.
\newblock Learning stiff chemical kinetics using extended deep neural operators.
\newblock \emph{Computer Methods in Applied Mechanics and Engineering}, 419:\penalty0 116674, 2024.
\newblock ISSN 0045-7825.
\newblock \doi{https://doi.org/10.1016/j.cma.2023.116674}.
\newblock URL \url{https://www.sciencedirect.com/science/article/pii/S0045782523007971}.

\bibitem[Kumar and Echekki(2024)]{Kumar_2024_Combustion}
Anuj Kumar and Tarek Echekki.
\newblock {Combustion chemistry acceleration with DeepONets}.
\newblock \emph{Fuel}, 365:\penalty0 131212, 2024.
\newblock ISSN 0016-2361.
\newblock \doi{https://doi.org/10.1016/j.fuel.2024.131212}.
\newblock URL \url{https://www.sciencedirect.com/science/article/pii/S0016236124003582}.

\bibitem[Weng et~al.(2025)Weng, Li, Zhang, Chen, and Zhou]{Weng_2025_ExFNO}
Yuting Weng, Han Li, Hao Zhang, Zhi~X. Chen, and Dezhi Zhou.
\newblock {Extended Fourier Neural Operators to learn stiff chemical kinetics under unseen conditions}.
\newblock \emph{Combustion and Flame}, 272:\penalty0 113847, 2025.
\newblock ISSN 0010-2180.
\newblock \doi{https://doi.org/10.1016/j.combustflame.2024.113847}.
\newblock URL \url{https://www.sciencedirect.com/science/article/pii/S001021802400556X}.

\bibitem[Box and Cox(1964)]{BoxCox_1964_BCtransf}
G.~E.~P. Box and D.~R. Cox.
\newblock An analysis of transformations.
\newblock \emph{Journal of the Royal Statistical Society: Series B (Methodological)}, 26\penalty0 (2):\penalty0 211--243, 07 1964.
\newblock ISSN 0035-9246.
\newblock \doi{10.1111/j.2517-6161.1964.tb00553.x}.
\newblock URL \url{https://doi.org/10.1111/j.2517-6161.1964.tb00553.x}.

\bibitem[Gu and Dao(2023)]{Gu_2023_mamba}
Albert Gu and Tri Dao.
\newblock Mamba: Linear-time sequence modeling with selective state spaces.
\newblock \emph{arXiv e-prints}, pages arXiv--2312, 2023.

\bibitem[Hu et~al.(2026)Hu, Daryakenari, Shen, Kawaguchi, and Karniadakis]{Hu_2024_state}
Zheyuan Hu, Nazanin~Ahmadi Daryakenari, Qianli Shen, Kenji Kawaguchi, and George~Em Karniadakis.
\newblock State-space models are accurate and efficient neural operators for dynamical systems.
\newblock \emph{Neural Networks}, 197:\penalty0 108496, 2026.
\newblock ISSN 0893-6080.
\newblock \doi{https://doi.org/10.1016/j.neunet.2025.108496}.
\newblock URL \url{https://www.sciencedirect.com/science/article/pii/S0893608025013772}.

\bibitem[Pandey et~al.(2025)Pandey, Wei, Babaee, and Karniadakis]{Pandey_2025_kineticmamba}
Additi Pandey, Liang Wei, Hessam Babaee, and George~Em Karniadakis.
\newblock {Kinetic-Mamba: Mamba-Assisted Predictions of Stiff Chemical Kinetics}.
\newblock \emph{arXiv preprint arXiv:2512.14471}, 2025.
\newblock URL \url{https://arxiv.org/abs/2512.14471}.

\bibitem[Chen and Chen(1995)]{Chen_1995_universal_operator}
Tianping Chen and Hong Chen.
\newblock Universal approximation to nonlinear operators by neural networks with arbitrary activation functions and its application to dynamical systems.
\newblock \emph{IEEE Transactions on Neural Networks}, 6\penalty0 (4):\penalty0 911--917, 1995.

\bibitem[Jin et~al.(2022)Jin, Meng, and Lu]{Pengzhan_2022_MIONet}
Pengzhan Jin, Shuai Meng, and Lu~Lu.
\newblock {MIONet: Learning Multiple-Input Operators via Tensor Product}.
\newblock \emph{SIAM Journal on Scientific Computing}, 44\penalty0 (6):\penalty0 A3490--A3514, 2022.
\newblock \doi{10.1137/22M1477751}.
\newblock URL \url{https://doi.org/10.1137/22M1477751}.

\bibitem[Lu et~al.(2022)Lu, Meng, Cai, Mao, Goswami, Zhang, and Karniadakis]{Lu_2022_Comp_deep_fno}
Lu~Lu, Xuhui Meng, Shengze Cai, Zhiping Mao, Somdatta Goswami, Zhongqiang Zhang, and George~Em Karniadakis.
\newblock {A comprehensive and fair comparison of two neural operators (with practical extensions) based on FAIR data}.
\newblock \emph{Computer Methods in Applied Mechanics and Engineering}, 393:\penalty0 114778, 2022.
\newblock ISSN 0045-7825.
\newblock \doi{https://doi.org/10.1016/j.cma.2022.114778}.
\newblock URL \url{https://www.sciencedirect.com/science/article/pii/S0045782522001207}.

\bibitem[Liu et~al.(2024{\natexlab{a}})Liu, Nath, and Cai]{Liu_2022_Causality}
Lizuo Liu, Kamaljyoti Nath, and Wei Cai.
\newblock {A Causality-DeepONet for Causal Responses of Linear Dynamical Systems}.
\newblock \emph{Communications in Computational Physics}, 35\penalty0 (5):\penalty0 1194--1228, 2024{\natexlab{a}}.
\newblock ISSN 1991-7120.
\newblock \doi{https://doi.org/10.4208/cicp.OA-2023-0078}.
\newblock URL \url{http://global-sci.org/intro/article_detail/cicp/23189.html}.

\bibitem[Zhu et~al.(2023{\natexlab{a}})Zhu, Feng, Lin, and Lu]{Zhu_2023_fourier}
Min Zhu, Shihang Feng, Youzuo Lin, and Lu~Lu.
\newblock {Fourier-DeepONet: Fourier-enhanced deep operator networks for full waveform inversion with improved accuracy, generalizability, and robustness}.
\newblock \emph{Computer Methods in Applied Mechanics and Engineering}, 416:\penalty0 116300, 2023{\natexlab{a}}.
\newblock ISSN 0045-7825.
\newblock \doi{https://doi.org/10.1016/j.cma.2023.116300}.
\newblock URL \url{https://www.sciencedirect.com/science/article/pii/S0045782523004243}.

\bibitem[Kiyani et~al.(2025)Kiyani, Manav, Kadivar, De~Lorenzis, and Karniadakis]{Kiyani_2024_Crack}
Elham Kiyani, Manav Manav, Nikhil Kadivar, Laura De~Lorenzis, and George~Em Karniadakis.
\newblock Predicting crack nucleation and propagation in brittle materials using {{Deep Operator Networks}} with diverse trunk architectures.
\newblock \emph{Computer Methods in Applied Mechanics and Engineering}, 441:\penalty0 117984, 2025.
\newblock ISSN 0045-7825.
\newblock \doi{10.1016/j.cma.2025.117984}.

\bibitem[Rivera-Casillas et~al.(2025)Rivera-Casillas, Dutta, Cai, Loveland, Nath, Shukla, Trahan, Lee, Farthing, and Dawson]{Rivera-Casillas_2025_SWE}
Peter Rivera-Casillas, Sourav Dutta, Shukai Cai, Mark Loveland, Kamaljyoti Nath, Khemraj Shukla, Corey Trahan, Jonghyun Lee, Matthew Farthing, and Clint Dawson.
\newblock {A Neural Operator-Based Emulator for Regional Shallow Water Dynamics}.
\newblock \emph{arXiv pre-print. 2502.14782}, 2025.
\newblock URL \url{https://arxiv.org/abs/2502.14782}.

\bibitem[Nath et~al.(2025)Nath, Shukla, Tsai, bin Waheed, Huber, Alpak, Chen, Lu, and St-Cyr]{Nath_2025_DeepONet_FWI}
Kamaljyoti Nath, Khemraj Shukla, Victor~C. Tsai, Umair bin Waheed, Christian Huber, Omer Alpak, Chuen-Song Chen, Ligang Lu, and Amik St-Cyr.
\newblock {Leveraging Deep Operator Networks ({DeepONet}) for Acoustic Full Waveform Inversion ({FWI})}.
\newblock \emph{arXiv preprint arXiv:2504.10720}, 2025.
\newblock URL \url{https://arxiv.org/abs/2504.10720}.

\bibitem[Higgins(2021)]{Higgins_2021}
Irina Higgins.
\newblock Generalizing universal function approximators.
\newblock \emph{Nature Machine Intelligence}, 3\penalty0 (3):\penalty0 192--193, 2021.

\bibitem[Liu et~al.(2024{\natexlab{b}})Liu, Wang, Vaidya, Ruehle, Halverson, Solja{\v{c}}i{\'c}, Hou, and Tegmark]{Liu_2024_KAN}
Ziming Liu, Yixuan Wang, Sachin Vaidya, Fabian Ruehle, James Halverson, Marin Solja{\v{c}}i{\'c}, Thomas~Y Hou, and Max Tegmark.
\newblock Kan: Kolmogorov-arnold networks.
\newblock \emph{arXiv preprint arXiv:2404.19756}, 2024{\natexlab{b}}.

\bibitem[Abueidda et~al.(2025)Abueidda, Pantidis, and Mobasher]{Abueidda_2025_DeepOKAN}
Diab~W. Abueidda, Panos Pantidis, and Mostafa~E. Mobasher.
\newblock {DeepOKAN: Deep operator network based on Kolmogorov Arnold networks for mechanics problems}.
\newblock \emph{Computer Methods in Applied Mechanics and Engineering}, 436:\penalty0 117699, 2025.
\newblock ISSN 0045-7825.
\newblock \doi{https://doi.org/10.1016/j.cma.2024.117699}.
\newblock URL \url{https://www.sciencedirect.com/science/article/pii/S0045782524009538}.

\bibitem[Shukla et~al.(2024)Shukla, Toscano, Wang, Zou, and Karniadakis]{Shukla_2024_DeepOKAN}
Khemraj Shukla, Juan~Diego Toscano, Zhicheng Wang, Zongren Zou, and George~Em Karniadakis.
\newblock {A comprehensive and FAIR comparison between MLP and KAN representations for differential equations and operator networks}.
\newblock \emph{Computer Methods in Applied Mechanics and Engineering}, 431:\penalty0 117290, 2024.
\newblock ISSN 0045-7825.
\newblock \doi{https://doi.org/10.1016/j.cma.2024.117290}.
\newblock URL \url{https://www.sciencedirect.com/science/article/pii/S0045782524005462}.

\bibitem[Lee and Shin(2024)]{Sanghyun_2024_two_step}
Sanghyun Lee and Yeonjong Shin.
\newblock {On the Training and Generalization of Deep Operator Networks}.
\newblock \emph{SIAM Journal on Scientific Computing}, 46\penalty0 (4):\penalty0 C273--C296, 2024.
\newblock \doi{10.1137/23M1598751}.
\newblock URL \url{https://doi.org/10.1137/23M1598751}.

\bibitem[Kingma and Ba(2014)]{Kingma_2014adam}
Diederik~P Kingma and Jimmy Ba.
\newblock {Adam: A method for stochastic optimization}.
\newblock \emph{arXiv preprint arXiv:1412.6980}, 2014.

\bibitem[Babu{\v{s}}ka and Melenk(1997)]{Babuvska_1997_PoU}
Ivo Babu{\v{s}}ka and Jens~M Melenk.
\newblock The partition of unity method.
\newblock \emph{International Journal for Numerical Methods in Engineering}, 40\penalty0 (4):\penalty0 727--758, 1997.

\bibitem[Lee et~al.(2021)Lee, Trask, Patel, Gulian, and Cyr]{Lee_2021_PoU}
Kookjin Lee, Nathaniel~A. Trask, Ravi~G. Patel, Mamikon~A. Gulian, and Eric~C. Cyr.
\newblock {Partition of unity networks: deep hp-approximation}.
\newblock \emph{CoRR}, abs/2101.11256, 2021.
\newblock URL \url{https://arxiv.org/abs/2101.11256}.

\bibitem[Trask et~al.(2022)Trask, Henriksen, Martinez, and Cyr]{N_trask_2022_PoU}
Nathaniel Trask, Amelia Henriksen, Carianne Martinez, and Eric Cyr.
\newblock {Hierarchical partition of unity networks: fast multilevel training}.
\newblock In \emph{Mathematical and Scientific Machine Learning}, pages 271--286. PMLR, 2022.

\bibitem[Sharma and Shankar(2025)]{Sharma_2025_Ensemble}
Ramansh Sharma and Varun Shankar.
\newblock Ensemble and {{Mixture-of-Experts DeepONets For Operator Learning}}.
\newblock \emph{arXiv preprint arXiv:2405.11907}, 2025.

\bibitem[McClenny and Braga-Neto(2023)]{Mcclenny_2023_self_adaptive}
Levi~D McClenny and Ulisses~M Braga-Neto.
\newblock Self-adaptive physics-informed neural networks.
\newblock \emph{Journal of Computational Physics}, 474:\penalty0 111722, 2023.

\bibitem[Anagnostopoulos et~al.(2024)Anagnostopoulos, Toscano, Stergiopulos, and Karniadakis]{Anagnostopoulos_2024_RBA}
Sokratis~J Anagnostopoulos, Juan~Diego Toscano, Nikolaos Stergiopulos, and George~Em Karniadakis.
\newblock Residual-based attention in physics-informed neural networks.
\newblock \emph{Computer Methods in Applied Mechanics and Engineering}, 421:\penalty0 116805, 2024.

\bibitem[Goswami et~al.(2023)Goswami, Bora, Yu, and Karniadakis]{Goswami_2023_Physics}
Somdatta Goswami, Aniruddha Bora, Yue Yu, and George~Em Karniadakis.
\newblock Physics-informed deep neural operator networks.
\newblock In Timon Rabczuk and Klaus-J{\"u}rgen Bathe, editors, \emph{Machine Learning in Modeling and Simulation: Methods and Applications}, pages 219--254. Springer International Publishing, Cham, 2023.
\newblock ISBN 978-3-031-36644-4.
\newblock \doi{10.1007/978-3-031-36644-4_6}.
\newblock URL \url{https://doi.org/10.1007/978-3-031-36644-4_6}.

\bibitem[Kontolati et~al.(2023)Kontolati, Goswami, Shields, and Karniadakis]{Kontolati_2023}
Katiana Kontolati, Somdatta Goswami, Michael~D. Shields, and George~Em Karniadakis.
\newblock On the influence of over-parameterization in manifold based surrogates and deep neural operators.
\newblock \emph{Journal of Computational Physics}, 479:\penalty0 112008, April 2023.
\newblock ISSN 0021-9991.
\newblock \doi{10.1016/j.jcp.2023.112008}.
\newblock URL \url{http://dx.doi.org/10.1016/j.jcp.2023.112008}.

\bibitem[Peyvan et~al.(2024)Peyvan, Oommen, Jagtap, and Karniadakis]{Ahmad_2024_RiemannONets}
Ahmad Peyvan, Vivek Oommen, Ameya~D. Jagtap, and George~Em Karniadakis.
\newblock {RiemannONets: Interpretable neural operators for Riemann problems}.
\newblock \emph{Computer Methods in Applied Mechanics and Engineering}, 426:\penalty0 116996, 2024.
\newblock ISSN 0045-7825.
\newblock \doi{https://doi.org/10.1016/j.cma.2024.116996}.
\newblock URL \url{https://www.sciencedirect.com/science/article/pii/S0045782524002524}.

\bibitem[Jin et~al.(2025)Jin, Zhang, Cao, Zhang, Bora, Krishnaswamy, Karniadakis, and Espinosa]{Jin_2025_Characterization}
Hanxun Jin, Boyu Zhang, Qianying Cao, Enrui Zhang, Aniruddha Bora, Sridhar Krishnaswamy, George~Em Karniadakis, and Horacio~D. Espinosa.
\newblock Characterization and {{Inverse Design}} of {{Stochastic Mechanical Metamaterials Using Neural Operators}}.
\newblock \emph{Advanced Materials}, page 2420063, 2025.
\newblock ISSN 0935-9648, 1521-4095.
\newblock \doi{10.1002/adma.202420063}.

\bibitem[Karniadakis and Sherwin(2005)]{Karniadakis_2005_Spectral_book}
George Karniadakis and Spencer~J Sherwin.
\newblock \emph{Spectral/hp element methods for computational fluid dynamics}.
\newblock Oxford University Press, 2005.

\bibitem[Paykani et~al.(2022)Paykani, Chehrmonavari, Tsolakis, Alger, Northrop, and Reitz]{Paykani_2022_Synthesis}
Amin Paykani, Hamed Chehrmonavari, Athanasios Tsolakis, Terry Alger, William~F. Northrop, and Rolf~D. Reitz.
\newblock Synthesis gas as a fuel for internal combustion engines in transportation.
\newblock \emph{Progress in Energy and Combustion Science}, 90:\penalty0 100995, 2022.
\newblock ISSN 0360-1285.
\newblock \doi{https://doi.org/10.1016/j.pecs.2022.100995}.
\newblock URL \url{https://www.sciencedirect.com/science/article/pii/S0360128522000041}.

\bibitem[Goodwin et~al.(2017)Goodwin, Moffat, and Speth]{Goodwin_2017_Cantera}
David~G Goodwin, Harry~K Moffat, and Raymond~L Speth.
\newblock Cantera: {{An Object-oriented Software Toolkit}} for {{Chemical Kinetics}}, {{Thermodynamics}}, and {{Transport Processes}}. {{Version}} 2.3.0.
\newblock \emph{Zenodo}, 2017.
\newblock \doi{10.5281/zenodo.170284}.

\bibitem[Zhu et~al.(2023{\natexlab{b}})Zhu, Zhang, Jiao, Karniadakis, and Lu]{Zhu_2023_Reliable_extrapolation}
Min Zhu, Handi Zhang, Anran Jiao, George~Em Karniadakis, and Lu~Lu.
\newblock Reliable extrapolation of deep neural operators informed by physics or sparse observations.
\newblock \emph{Computer Methods in Applied Mechanics and Engineering}, 412:\penalty0 116064, 2023{\natexlab{b}}.
\newblock ISSN 0045-7825.
\newblock \doi{https://doi.org/10.1016/j.cma.2023.116064}.
\newblock URL \url{https://www.sciencedirect.com/science/article/pii/S0045782523001883}.

\bibitem[Nath et~al.(2026)Nath, Kumar, Smith, and Karniadakis]{Nath_2024_Digital}
Kamaljyoti Nath, Varun Kumar, Daniel~J. Smith, and George~Em Karniadakis.
\newblock A digital twin for diesel engines: Operator-infused physics-informed neural networks with transfer learning for engine health monitoring.
\newblock \emph{Engineering Applications of Artificial Intelligence}, 170:\penalty0 114052, 2026.
\newblock ISSN 0952-1976.
\newblock \doi{https://doi.org/10.1016/j.engappai.2026.114052}.
\newblock URL \url{https://www.sciencedirect.com/science/article/pii/S0952197626003337}.

\end{thebibliography}
